\documentclass[10pt,twocolumn,letterpaper]{article}

\usepackage[pagenumbers]{cvpr} 

\usepackage{array}
\usepackage{colortbl}
\usepackage{multirow}
\usepackage{makecell}

\definecolor{cvprblue}{rgb}{0.21,0.49,0.74}
\usepackage[pagebackref,breaklinks,colorlinks,allcolors=cvprblue]{hyperref}
\providecommand{\methodname}{RoboRAM}
\providecommand{\rmbshift}{RMBench-$\Delta$}
\providecommand{\rmashift}{RoboMemArena-$\Delta$}
\usepackage{graphicx,booktabs,multirow,colortbl,array,amssymb}
\usepackage[section]{placeins}
\providecommand{\methodname}{RoboRAM}
\definecolor{shiftbwd}{HTML}{9A5E2C}\definecolor{shiftfwd}{HTML}{C0675A}
\definecolor{bwdbg}{HTML}{FBF3E8}\definecolor{fwdbg}{HTML}{FCEFEC}\definecolor{grpbg}{HTML}{F2F2F2}
\definecolor{bwdband}{HTML}{F1E8E1}\definecolor{fwdband}{HTML}{F7EDEB}
\providecommand{\bwdtag}{\leavevmode\mbox{\color{shiftbwd}$\blacktriangleleft\!\!\blacktriangleleft$\,\textbf{Backward}}}
\providecommand{\fwdtag}{\leavevmode\mbox{\color{shiftfwd}$\blacktriangleright\!\!\blacktriangleright$\,\textbf{Forward}}}
\providecommand{\thn}{\,$\rightarrow$\,}
\providecommand{\shiftband}[2]{\par\vspace{1.5pt}\noindent\colorbox{#1}{\parbox{\dimexpr\linewidth-2\fboxsep\relax}{#2}}\par\vspace{1.5pt}}
\definecolor{rampink}{HTML}{C97B6E}
\definecolor{cvprblue}{rgb}{0.21,0.49,0.74}
\usepackage[pagebackref,breaklinks,colorlinks,allcolors=cvprblue]{hyperref}

\def\confName{CVPR}
\def\confYear{2027}

\title{RoboRAM: Random-Access Memory for Robots under Task-Progress Shifts}

\author{
\makebox[9.5em]{Kuancheng Wang\textsuperscript{1}}%
\makebox[9.5em]{Seungho Yeom\textsuperscript{1}}%
\makebox[9.5em]{Yuheng Zhi\textsuperscript{1}} \\
\makebox[9.5em]{Nikhil Shinde\textsuperscript{1}}%
\makebox[9.5em]{Sainan Liu\textsuperscript{2\dag}}%
\makebox[9.5em]{Michael Yip\textsuperscript{1\dag}} \\[4pt]
\textsuperscript{1}UC San Diego \qquad \textsuperscript{2}Intel \qquad \textsuperscript{\dag}Equal advising \\
{\tt\small kuw014@ucsd.edu} \\
{\small \href{https://kuanchengwang.github.io/RoboRAM/}{\textcolor{rampink}{\textbf{\nolinkurl{https://kuanchengwang.github.io/RoboRAM/}}}}}
}

\begin{document}
\providecommand{\methodname}{RoboRAM}
\twocolumn[{%
\renewcommand\twocolumn[1][]{#1}%
\maketitle
\begin{center}
\vspace{-1.5em}
\includegraphics[width=\textwidth]{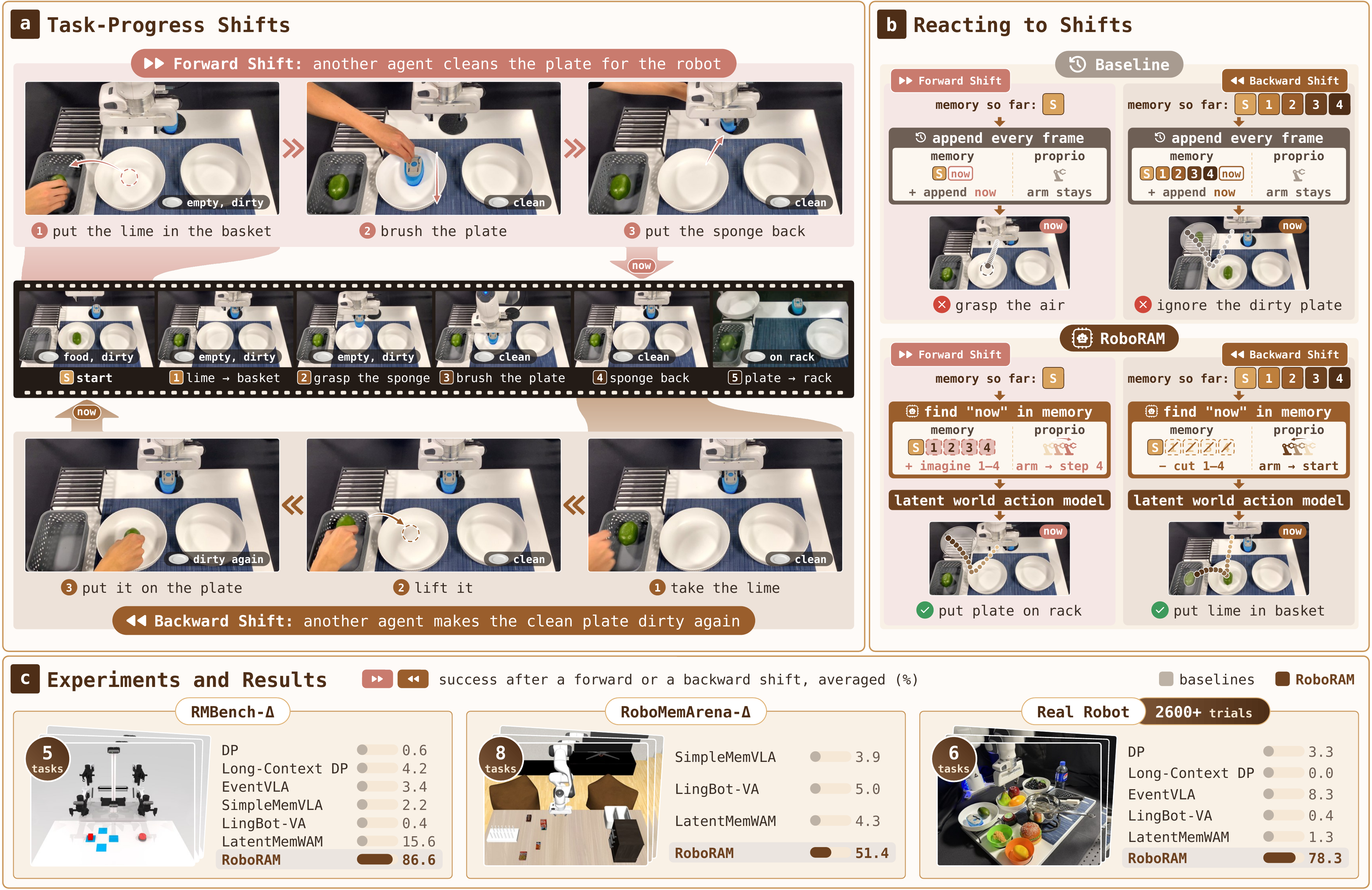}
\captionof{figure}{\textbf{Reacting to task-progress shifts.}
\textbf{(a)}~During execution, another agent shifts the task progress forward or backward.
\textbf{(b)}~A baseline memory policy appends every observation to its memory. After a shift, this memory is inconsistent with the world, and
the policy fails. \methodname{} instead locates the current observation within its memory and revises the memory, extending it with predicted
entries after a forward shift and truncating it after a backward shift.
\textbf{(c)}~Under task-progress shifts, \methodname{} substantially outperforms all baselines on two simulated benchmarks and in real-robot experiments.}
\label{fig:teaser}
\end{center}
}]

\providecommand{\methodname}{RoboRAM}
\begin{abstract}
Robots in shared workspaces must handle other agents, who may undo completed steps of a task or complete upcoming ones.
Memory policies learned by imitation degrade sharply under such task-progress shifts, and demonstrating them is labor-intensive, as shifts in either direction at any timestep form a combinatorial space.
Our key insight is that such shifts need not be demonstrated, since each moves the scene to a point of the episode that the robot has already observed or can predict, so the robot can recover by moving its memory there.
We introduce Robot Random-Access Memory (RoboRAM), a two-stage framework that lets a memory policy jump to any point of its own timeline instead of advancing step by step.
A latent world action model extends the observed history with predicted future steps into a pseudo-memory timeline, and a self-supervised memory localizer moves the memory's pointer to the step of this timeline that matches the current observation, rewinding or fast-forwarding it.
Trained only on standard demonstrations, RoboRAM is robust to shifts in both directions and competitive with state-of-the-art memory policies on unperturbed episodes.
We further introduce \mbox{RMBench-$\Delta$} and \mbox{RoboMemArena-$\Delta$}, task-progress-shifted extensions of two robot memory benchmarks, on which RoboRAM succeeds in 86.6\% and 51.4\% of shifted episodes, whereas no baseline, including state-of-the-art memory vision-language-action and world action models, exceeds 15.6\% and 5.0\%.
RoboRAM also works as a plug-in for existing world action models, raising LingBot-VA's success on shifted episodes of two tasks from 0\% to 51.5\% without changing its weights.
In 2,640 real-robot trials, RoboRAM succeeds in 78.3\% of one-step and 61.3\% of multi-step shifted trials, against at most 8.3\% for any baseline, and without a shift it matches its base world action model and outperforms all other baselines.
\end{abstract}
\section{Introduction}
\label{sec:intro}
Memory lets a robot policy act on what it can no longer observe, such as where an object came from or which steps of a task are done.
Memory-augmented visuomotor policies~\cite{wang2026remember,torne2025learning},
vision-language-action (VLA) models~\cite{yin2026simplememvla,li2026bridgevla++,mark2026bpp}
and world action models (WAMs)~\cite{li2026causal,miao2026rm} learn this by imitation learning,
predicting demonstrated actions from the observed history. Their demonstrations implicitly assume that the robot acts alone, keeping memory and the world in lockstep. The assumption fails once robots share their workspace
with people or other robots. Another agent may undo a completed step, which we call a \emph{backward shift}, or complete an upcoming step for the
robot, a \emph{forward shift}, as shown in~\cref{fig:teaser} (a). After such a \emph{task-progress shift}, the scene and the skills it calls for are still demonstrated, yet the
memory describes a moment that has passed or has not yet arrived. Since such shifts can occur in any shared multi-stage task, a memory policy must
detect them and relocate its memory accordingly.

Existing memory policies are not built for such shifts, since their memory is \emph{sequential-access} and only advances, one observation at a time, illustrated in~\cref{fig:teaser} (b). Long-context policies~\cite{torne2025learning} and memory-augmented VLAs, which keep frames of past
steps~\cite{yang2026eventvla,yin2026simplememvla,li2026bridgevla++},
follow their outdated history, so an undone step still appears done. Fixed-size memories~\cite{wang2026remember} compress
the history and keep no earlier moment to return to, and WAMs~\cite{li2026causal} predict the future but always extend their
memory from the latest step. Imitation learning gives no reason to do otherwise, since in demonstrations reading task progress from the memory
fits the data as well as reading it from the scene, a form of causal confusion~\cite{de2019causal}.
Demonstrations with shifts~\cite{jiang2026robottt} would supply one, but would have to cover every stage of every task in both
directions, which can be extremely labor-intensive (\cref{sec:supp-problem}).

Our key observation is that a task-progress shift moves the scene to an earlier or a later step of the episode's own timeline. The memory is still an accurate record, only misplaced relative to the scene, so recovery reduces to locating the scene on the timeline and resuming there.
This calls for \emph{random access} to any step of the timeline, not only the latest.
We introduce \textbf{Robo}t \textbf{R}andom-\textbf{A}ccess \textbf{M}emory (\textbf{\methodname}), which gives a memory policy this ability
by building on a latent WAM whose memory holds the observed steps followed by the predicted ones, forming a \emph{pseudo-memory timeline}, shown in~\cref{fig:teaser} (b). A
lightweight memory localizer reads this timeline as a progress bar and sets the memory's pointer to the step where the current observation
belongs, rewinding after a backward shift, fast-forwarding after a forward shift and pausing while a change is under way. The localizer is
self-supervised. Each example pairs the memory at one step of a standard demonstration with the scene at another, labeled with the step whose
prediction by the frozen WAM matches that scene. \methodname{} thus learns offline to recover from shifts without labor-intensive shift data,
manual labels or rewards. Our contributions are:
\begin{itemize}
\item We formalize task-progress shifts and argue that imitation leaves a memory policy's behavior under them unconstrained, which we confirm
empirically.
\item We propose \methodname{}, a random-access memory over observed and predicted steps with a learned localizer.
\item We introduce \rmbshift{} and \rmashift{}, which extend the established benchmarks RMBench~\cite{chen2026rmbench}
and RoboMemArena~\cite{lei2026robomemarena} with task-progress shifts.
\item On these benchmarks, \methodname{} succeeds in 86.6\% and 51.4\% of shifted episodes, while no baseline, including the
state-of-the-art memory VLA and WAM, exceeds 15.6\% and 5.0\%, and it stays competitive without shifts. As a plug-in, it raises the existing WAM
LingBot-VA from 0\% to 51.5\% on two tasks without changing its weights. In 2,640 real-robot trials, it succeeds in 78.3\% of trials with a
one-step shift, the best baseline in 8.3\%.
\end{itemize}
   
\section{Related Work}
\label{sec:related}
\textbf{Benchmarks for robot memory.}
Memory-dependent manipulation benchmarks~\cite{chen2026rmbench,lei2026robomemarena,cherepanov2026memory,wang2026remember,yang2026eventvla}
test whether a policy recalls what it can no longer observe, but the robot always acts alone, so the scene changes only through its own actions
and a policy can succeed by following its memory alone. Robustness benchmarks perturb appearance, object poses or
distractors~\cite{zhou2025libero, fei2025libero} and move objects while the policy
acts~\cite{zhang2026libero}.
\rmbshift{} and \rmashift{} instead add backward and forward task-progress shifts to established memory tasks, testing whether a memory policy
recovers from them rather than acting on its outdated memory.

\textbf{Memory policies and memory revision.}
Policies keep long observation histories~\cite{torne2025learning},
keyframes~\cite{yang2026eventvla,fang2025sam2act,li2026bridgevla++},
videos~\cite{yin2026simplememvla}, memory-token banks~\cite{shi2026memoryvla},
compressed or recurrent states~\cite{wang2026remember,zhou2026chronos} or a world action model's key--value
cache~\cite{li2026causal,yang2026memorywam,miao2026rm}, all sequential-access, and world action models use predictions only to act.
Policies learn recovery from changes in training data~\cite{gao2026gated}; 
RoboTTT~\cite{jiang2026robottt} learns it from 30 minutes of a person
removing installed parts, a backward shift, but task-progress shifts need such data at every step in both directions. Others learn how
much memory to trust~\cite{miao2026rm} or revise a predicted plan~\cite{liu2026revision}. \methodname{} learns random access to a latent memory from standard demonstrations alone. 
A detailed comparison with concurrent work is given in \cref{sec:supp-concurrent}.

\section{Method}
\label{sec:method}
\begin{figure*}[t]
\centering
\includegraphics[width=\textwidth]{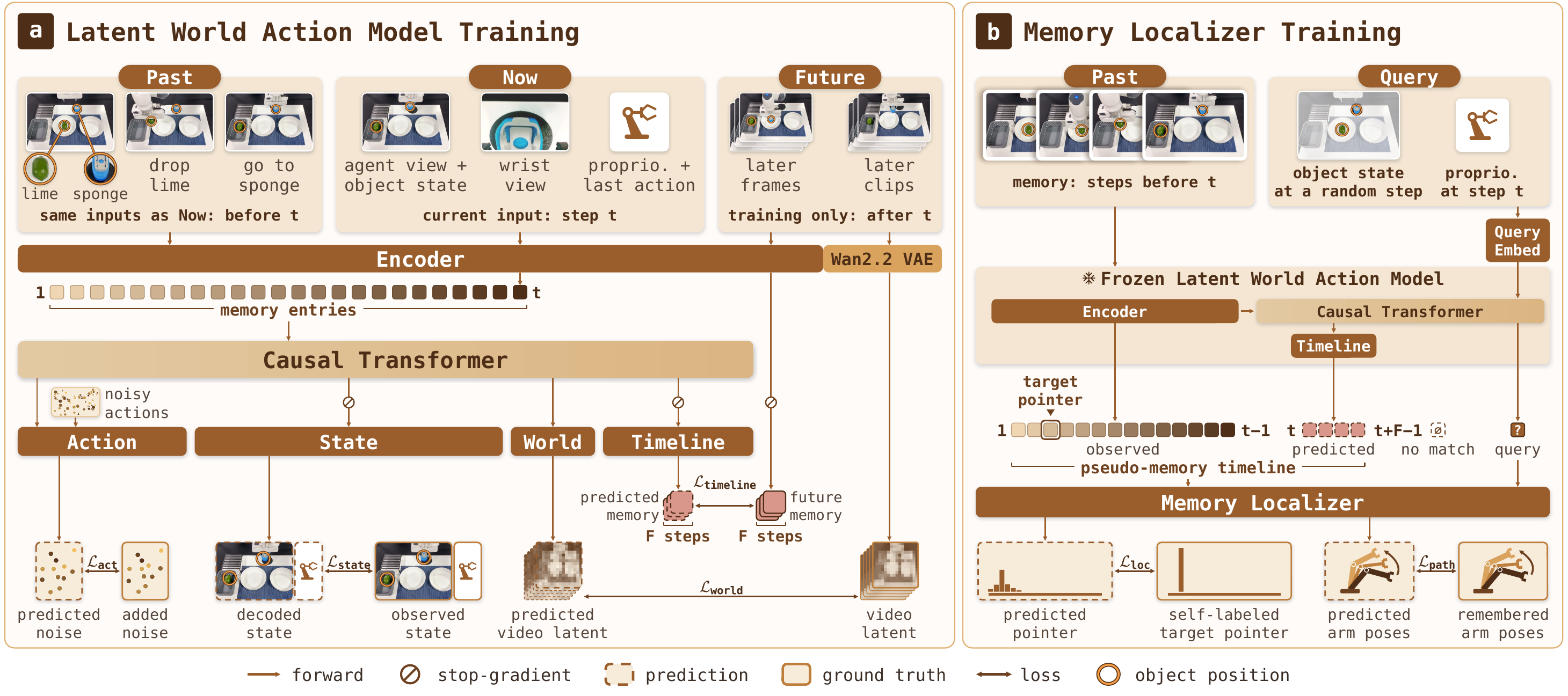}
\caption{\textbf{Overview of \methodname{}.} \textbf{(a)} Training the latent world action model. Its memory holds one entry per past step, and its heads predict the next actions, the current state, future video latents and the next $F$ entries, so that observed and predicted entries form one pseudo-memory timeline. \textbf{(b)} Training the memory localizer on the frozen model, without manual labels. It learns to place a query mimicking a task-progress shift on this timeline and to predict the arm poses back there.}
\label{fig:method}
\end{figure*}

When another agent shifts a task's progress, the robot's memory remains an accurate record of the episode, but of a moment that the scene has
already left or not yet reached. \methodname{} revises such a memory by \emph{random access}. It extends the memory with predicted entries into a
pseudo-memory timeline, locates the current observation on it and moves the memory's pointer there in a single jump.
Sec.~\ref{sec:problem} reduces revision to estimating this pointer, stage~1 learns the timeline (Sec.~\ref{sec:model}), stage~2 learns the
memory localizer that estimates it (Sec.~\ref{sec:training}), and Sec.~\ref{sec:revision} uses both at test time. \methodname{} is learned from standard demonstrations
alone, without demonstrations of shifts or manual labels. Details of the method are given in \cref{sec:supp-method}.

\subsection{Preliminaries}
\label{sec:problem}

\textbf{Memory policies.} We model a manipulation task as a Partially Observable Markov Decision Process (POMDP) with state $z_t=(s_t,q_t,f_t)$, composed of the object \emph{scene} $s_t$,
the robot's \emph{proprioception} $q_t$ and \emph{latent facts} $f_t$ that the images do not show, so that acting requires memory. From
observations $o_t=(I_t,q_t)$, a \emph{memory policy} acts with an action rule $\pi$ and a memory update $u$,
\begin{equation}
\label{eq:update}
a_t\sim\pi(\cdot\mid m_t,o_t),\qquad m_{t+1}=u(m_t,o_t,a_t),
\end{equation}
from an empty memory $m_1$. It is learned by behavior cloning on demonstrations $\xi=(o_1,a_1,\dots,o_T,a_T)\in\mathcal D$ with the loss
$L_0=-\sum_{\xi\in\mathcal D}\sum_t\log\pi(a_t\mid m_t,o_t)$.

\textbf{Task-progress shifts.} In a demonstration only the robot acts, so state and memory advance together along one \emph{course}
$\bigl((z_1,m_1),\dots,(z_T,m_T)\bigr)$, and $(z,m)\in\mathcal D$ denotes a pair on some course. At test time another agent may move the scene at
step $t$ to its state $s_\ell$ at another step $\ell$ of the course, leaving $f_t$, $q_t$ and $m_t$ unchanged. These \emph{task-progress shifts},
\begin{equation}
\label{eq:scenario}
\mathcal S=\bigl\{\bigl((s_\ell,q_t,f_t),m_t\bigr)\notin\mathcal D\ \big|\ s_\ell\neq s_t\bigr\},
\end{equation}
are \emph{backward} if $\ell<t$ and \emph{forward} if $\ell>t$. We restrict $s_\ell$ to the memory's own course, since a scene on no course,
such as a step left half done, would test generalization rather than memory. Each component occurs in $\mathcal D$ but never this combination, so $L_0$
leaves the policy's behavior on $\mathcal S$ unconstrained. We seek an update $\tilde u$, learned from $\mathcal D$ alone, that preserves the
behavior on $\mathcal D$ and completes the task from every shift in $\mathcal S$.

\textbf{Revision as random access.}\label{sec:filter} Because $s_\ell$ lies on the memory's own course, replacing $m_t$ with $m_{\ell+1}$ and
moving the robot to its pose $q_\ell$ restores, up to one step, a memory--observation pair of $\mathcal D$, the only kind the policy was trained on. We call the step to which the current observation belongs the memory's
\emph{pointer}. The update $u$ advances the pointer one step at a time, whereas revision moves it from $t$ to $\ell$, backward or forward, and
thus reduces to estimating $\ell$. This is no lookup by scene, because a scene can recur at several steps of a course, as when a block
returns to where it started, and only the memory tells them apart. After a
backward shift, $m_{\ell+1}$ is contained in $m_t$ as its first $\ell$ entries, whereas after a forward shift it does not exist yet and must be
predicted (Sec.~\ref{sec:model}). In both directions, $\ell$ must be estimated without a demonstrated shift (Sec.~\ref{sec:training}).

\subsection{Latent world action model}
\label{sec:model}

Stage~1 trains the memory policy as a latent world action model, which acts and predicts its next steps in its own latent space, illustrated in \cref{fig:method} (a). Its memory
supplies the past of the timeline and its predictions supply the future.

\textbf{Memory.} At each step, an encoder $E_\omega$ maps the observation, the previous action and the object state $x_t$ to a token that a causal transformer~\cite{chen2021decision,kazemnejad2023impact} processes given the memory,
\begin{equation}
\label{eq:token}
\tau_t=E_\omega(o_t,a_{t-1},x_t),\qquad h_t=\operatorname{Tr}_\omega(\tau_t;m_t).
\end{equation}
The object state $x_t=(p_t,v_t)$ holds the image position $p^k_t$ and the visibility $v^k_t$ of each task object $k$, obtained with an
open-vocabulary detector and a video segmentation model~\cite{liu2024grounding,ravi2025sam}. The memory $m_t$ is
the transformer's key--value cache, with one \emph{observed entry} per past step. Because attention is causal, the first $\ell-1$ entries of $m_t$
are exactly $m_\ell$, so truncating the cache restores an earlier memory exactly.

\textbf{Heads.} Acting needs only the first of four heads. An \emph{action head}, a
conditional diffusion model~\cite{chi2025diffusion}, generates the next $N$ actions from the current
observation and a code $\eta_t$ drawn from a variational information bottleneck
$\mathcal B_\omega(\eta_t\mid h_t)$~\cite{alemi2016deep}. With $\rho_t=(x_t,q_t)$ the \emph{measurement} of step
$t$, a \emph{timeline head} predicts, for each horizon $\delta\le F$, the token $\hat\tau_{t+\delta\mid t}$ and the measurement
$\hat\rho_{t+\delta\mid t}$ of step $t+\delta$, with a mean $\mu_{t+\delta\mid t}$ and a standard deviation $\sigma_{t+\delta\mid t}$ for each
object position. A \emph{state head} reads the measurement $\tilde\rho_t$ back from $h_t$, and a \emph{world head}, used only in training to
ground the representation in the camera images, predicts a density $g_\omega$ over the video latent~\cite{wan2025wan}
$c_{t+1}$ of the next camera clip.

\textbf{Objective.} With $d(\hat\rho,\rho)=\|\hat p-p\|^2+\|\hat q-q\|^2+\operatorname{BCE}(\hat v,v)$ the error between measurements, the
model minimizes
\begin{align}
&L_{\mathrm{model}}=L_{\mathrm{act}}+\beta L_{\mathrm{KL}}+\gamma L_{\mathrm{world}}+L_{\mathrm{timeline}}+L_{\mathrm{state}},\label{eq:model}\\
&L_{\mathrm{act}}=\textstyle\sum_t\mathbb E_{j,\epsilon}\bigl\|\epsilon-\hat\epsilon_\omega\bigl(\tilde A_t^{j},\,j,\,o_t,\eta_t\bigr)\bigr\|^2,\label{eq:act}\\
&L_{\mathrm{KL}}=\textstyle\sum_t\mathrm{KL}\bigl(\mathcal B_\omega(\eta_t\mid h_t)\,\|\,\mathcal N(0,I)\bigr),\nonumber\\
&L_{\mathrm{world}}=-\textstyle\sum_t\log g_\omega(c_{t+1}\mid o_t,\eta_t),\nonumber\\
&L_{\mathrm{timeline}}=\textstyle\sum_{t,\delta}\bigl[\|\hat\tau_{t+\delta\mid t}-\operatorname{sg}(\tau_{t+\delta})\|^2+d(\hat\rho_{t+\delta\mid t},\rho_{t+\delta})\nonumber\\
&\qquad\qquad-\log\mathcal N\bigl(p_{t+\delta};\operatorname{sg}(\mu_{t+\delta\mid t}),\sigma^2_{t+\delta\mid t}\bigr)\bigr],\label{eq:timeline}\\
&L_{\mathrm{state}}=\textstyle\sum_t d(\tilde\rho_t,\rho_t),\nonumber
\end{align}
summed over the steps of every $\xi\in\mathcal D$. $L_{\mathrm{act}}$ is the imitation loss $L_0$ in diffusion form, with $\tilde A_t^{j}$ the
chunk of the next $N$ actions noised to level $j$ by Gaussian noise $\epsilon$, $\operatorname{BCE}$ is the binary cross-entropy and
$\operatorname{sg}$ the stop-gradient. The timeline and state heads read $\operatorname{sg}(h_t)$, so that
forecasting does not compete with imitation for the representation from which the robot acts, and $\sigma$, fitted to the error of $\mu$,
quantifies the policy's own predictive uncertainty. After training, the model is frozen.

\subsection{Self-supervised memory localizer}
\label{sec:training}

Stage~2 trains a \emph{memory localizer} on the episode's \emph{pseudo-memory timeline}, illustrated in \cref{fig:method} (b). We write $n$ for the number of observed entries in
the memory, $t-1$ at step $t$ until a revision changes it. Passed through the frozen transformer, the predicted tokens $\hat\tau_{n+\delta\mid n}$
append $F$ \emph{predicted entries} to them, and these $n+F$ entries form a timeline that runs from the first step through the present into the steps the model expects next. It is a pseudo-memory because its last $F$
entries are predicted rather than observed, yet a step that another agent completes ahead of the robot lies on it like an observed step. The
localizer estimates the pointer, the step on this timeline to which the current observation belongs. Since demonstrations
contain no shifts, both its inputs and its targets are built from the demonstrations and the frozen model.

\textbf{Localizer.} A small attention network with weights $\psi$ reads the pseudo-memory timeline, the proprioception $q$ and the object state
$x$, but not the images. It outputs a distribution $P_\psi(\cdot\mid m,q,x)$ over $\Theta=\{1,\dots,n+F\}\cup\{\varnothing\}$, the entries of the
timeline and $\varnothing$ for an observation that matches no entry, for example while another agent is still moving an object. The
distribution acts as a progress bar over the episode's own memory. Its mass stays on the expected entry $n+1$ while memory and scene agree, and
moves backward, forward or to $\varnothing$ when they do not.

\textbf{Synthetic shifts.} Because the localizer sees the scene only through the object state, a shift is synthesized by exchanging it,
without rendering an image. An example combines the memory and proprioception at step $n+1$ of a demonstration $\xi$ with the
object state at step $\ell$ of a demonstration $\xi'$,
\begin{equation}
\label{eq:example}
(m,q,x)=\bigl(m^{\xi}_{n+1},\,q^{\xi}_{n+1},\,x^{\xi'}_{\ell}\bigr).
\end{equation}
For $\xi'=\xi$, this is the input after a shift to step $\ell$, or the unchanged input if $\ell=n+1$, because a fixed camera sees objects at
rest independently of the robot's pose. For $\xi'\neq\xi$, it may match no step of the course.

\textbf{Self-supervised targets.} The frozen model labels every example itself. Entry $\ell$ is \emph{consistent} with an example if each
visible object $k\in\mathcal V$ lies within a bound $\kappa$ of the position $\mu^k_\ell$ that the timeline head predicts for that entry, in
units of the predicted standard deviation $\sigma^k_\ell$,
\begin{equation}
\label{eq:consistent}
\mathcal E=\bigl\{\ell\le n+F \;\big|\; \textstyle\max_{k\in\mathcal V}\bigl\|(p^k-\mu^k_{\ell})/\sigma^k_{\ell}\bigr\|^2\le\kappa\bigr\}.
\end{equation}
The bound $\kappa$ is measured rather than chosen, as the largest such error of the one-step prediction on the unperturbed demonstrations. The
target is
\begin{align}
&\ell^\star=\begin{cases}
n+1 & \text{if } n+1\in\mathcal E,\\
\operatorname*{arg\,min}\limits_{\ell\in\mathcal E}\, r_\ell & \text{else if } \mathcal E\neq\emptyset,\\
\varnothing & \text{otherwise,}
\end{cases}\label{eq:map}\\
&r_\ell=\bigl(\mathbf 1[\ell>n],\,\mathbf 1[\mathrm{closed}_\ell],\,\|q_\ell-q\|\bigr),\nonumber
\end{align}
with a lexicographic minimum. It prefers the expected entry, then an observed entry to a predicted one, which would credit the robot with
actions it has not executed, then one with open grippers, and then the one whose pose $q_\ell$ is closest to the robot's. The bound and the standard deviations
only build the targets, and the localizer never sees them.

\textbf{Objective.} The localizer minimizes
\begin{equation}
\label{eq:loc}
L_{\mathrm{loc}}=-\log P_\psi(\ell^\star\mid m,q,x)+L_{\mathrm{path}},
\end{equation}
where $L_{\mathrm{path}}$ trains a second output, the \emph{path}, to bring the robot back to any entry through the poses that the entries hold.

\subsection{Random access at test time}
\label{sec:revision}

After every $H$ executed actions, the robot makes a \emph{decision} in which it notices a shift, locates the scene
and moves the memory's pointer there. The action head reads the memory $m$, with $n$ entries, and a memory set aside during a pause is kept as $\bar m$.

\textbf{Notice.} Most decisions see no shift, and revising a correct memory only does harm. A revision is therefore allowed only when the
scene departs from the model's own forecast,
\begin{equation}
\label{eq:gate}
e_t=\max_{k\in\mathcal V_t}\bigl\|p_t^k-\mu^k_{t\mid t'}\bigr\|>b,
\end{equation}
where $\mathcal V_t$ holds the visible objects, $\mu^k_{t\mid t'}$ is the position predicted at the previous decision $t'$, and the bound $b$ is
again measured, as the largest such error on the demonstrations. The gate only permits a revision, and the localizer decides whether
and where the pointer moves.

\textbf{Jump.}\label{sec:locate} The localizer's estimate $\hat\ell$ of the pointer is $\varnothing$ if
$P_\psi(\varnothing\mid m,q_t,x_t)>\frac12$ and the most probable entry otherwise. After a rewind or fast-forward, the arm moves straight to the
pose of the revised memory's last entry before acting.
\begin{itemize}\setlength{\itemsep}{1pt}\setlength{\parskip}{0pt}\setlength{\topsep}{2pt}
\item \textbf{Keep} ($\hat\ell\in\{n,n+1\}$). The frame is written and the robot acts.
\item \textbf{Rewind} ($\hat\ell<n$). The first $\hat\ell$ entries are kept.
\item \textbf{Fast-forward} ($\hat\ell>n+1$). The predicted entries up to $\hat\ell$, and on to the open-gripper entry nearest the arm, are
appended, recording the other agent's steps as the demonstrations do, so they are not repeated.
\item \textbf{Pause} ($\hat\ell=\varnothing$). The memory is set aside as $\bar m$, the path returns the arm to its initial pose, and the robot
acts from an empty memory. Once the localizer places the scene at the end of $\bar m$ or beyond, $\bar m$ is restored, so nothing remembered is
lost.
\end{itemize}

\textbf{Write.} A frame observed while another agent acts belongs to no step of the course, so it is written only as far as the localizer
places it at the expected entry,
\begin{equation}
\label{eq:write}
\tau^{\mathrm w}_t=\lambda\,\tau_t+(1-\lambda)\,\hat\tau_{n+1},
\end{equation}
where $\lambda=\sum_{|\ell-n-1|<H}P_\psi(\ell\mid m,q_t,x_t)$ is the probability that the observation lies within one decision of
entry $n+1$.
\section{Experiments}
\label{sec:experiments}
We build two benchmarks of task-progress shifts on established robot memory benchmarks, \mbox{RMBench-$\Delta$} on
RMBench~\cite{chen2026rmbench} and \mbox{RoboMemArena-$\Delta$} on
RoboMemArena~\cite{lei2026robomemarena}, by adding another agent that undoes or completes steps of the task. We evaluate RoboRAM extensively on both simulation benchmarks and a real robot. We ask whether
\methodname{} recovers from shifts in both directions where state-of-the-art memory policies, VLAs and WAMs do not, what these baselines do
instead and whether task success reveals it, and what revising the memory and each component of \methodname{} contribute. We then test whether
\methodname{} makes an existing WAM reactive without changing its weights, keeps its performance when nothing shifts, carries over to a real
robot and to multi-step shifts, and runs fast enough to deploy.
Details of experiments and results are further provided in \cref{sec:supp-sim,sec:supp-real}.
\subsection{Simulation Experiments}
\subsubsection{Benchmarks and Baselines}
\label{sec:exp-setup}
\textbf{Benchmarks.}
\textbf{\mbox{RMBench-$\Delta$}} uses five memory-dependent dual-arm tasks of RMBench, Put Back Block, Rearrange Blocks, Swap Blocks, Cover Blocks and Blocks Ranking Try, with 50 standard demonstrations provided by RMBench~\cite{chen2026rmbench} without shifts. \textbf{\mbox{RoboMemArena-$\Delta$}} uses eight single-arm tasks of RoboMemArena
with 100 standard demonstrations each provided by RoboMemArena~\cite{lei2026robomemarena}, in which items put into a drawer or a microwave leave the camera's view. In both, a rendered hand makes one task-progress shift per episode.

\textbf{Metrics.}
We report the task success rate (TSR), judged by each base benchmark's success check, with and without a shift, and the success rates after backward and forward shifts (BSR, FSR). After a shift, a policy that follows its outdated memory can still pass this check after a
\emph{ghost action}, \eg, closing its gripper on nothing at a place that a shifted object has left. We therefore count an episode as a success
after a shift only if it completes the task without a ghost action, and the tables also give the task success after a shift with every success
counted. On the real robot, ghost actions are judged in the same way.

\textbf{Baselines.} We compare with Diffusion Policy (DP)~\cite{chi2025diffusion} without memory,
Long-Context DP~\cite{torne2025learning}, EventVLA~\cite{yang2026eventvla},
SimpleMemVLA~\cite{yin2026simplememvla}, LingBot-VA~\cite{li2026causal} and LatentMemWAM. Since the code of
MemoryWAM~\cite{yang2026memorywam} is not released, LatentMemWAM is our own model inspired by its design, with our tracker and
hyperparameters, and is \methodname{}'s stage~1 without the memory localizer. EventVLA and SimpleMemVLA use the checkpoints their authors
released, EventVLA's for RMBench only, so on \mbox{RoboMemArena-$\Delta$}  we compare with SimpleMemVLA, LingBot-VA and LatentMemWAM, and leave out DP and Long-Context DP because of their low success after a shift on \mbox{RMBench-$\Delta$}.

\subsubsection{Main Results}
\textbf{State-of-the-art memory policies, VLAs, and WAMs rarely recover from a task-progress shift, whereas \methodname{} recovers in both directions.}
Without a shift, the strongest baselines succeed about as often as \methodname{} (\cref{tab:main,tab:main-rma}). SimpleMemVLA and LingBot-VA complete 99.6 and 95.6\% of RMBench episodes against 96.0\%, and LingBot-VA completes 78.8\% of \mbox{RoboMemArena-$\Delta$} episodes
against 79.8\%. After a backward shift, no baseline exceeds 12.4 and 7.5\% BSR, whereas \methodname{} reaches 90.0 and 32.5\%. After a forward shift, \methodname{} reaches 83.2 and 70.3\% FSR, against at most 18.8 and 8.3\% for any baseline. Some baselines still complete most forward-shifted episodes, up to 94.0\% for SimpleMemVLA, but not by reacting. In a demonstration only the robot advances the task, so a memory of its own motion, or even the elapsed time, works as a clock that sets the next step without reading the scene. Following this clock, a policy repeats a step that another agent has done, closing its gripper where the object no longer is, yet passes the success check, and after a backward shift it skips the undone step and rarely succeeds. \methodname{} instead resumes from the step that the scene shows, so its successes after a forward shift are almost all free of ghost actions, with 84.8\% task success against 83.2\% FSR on \mbox{RMBench-$\Delta$}, and a failed revision costs it the episode rather than hiding behind a ghost action. LatentMemWAM, the same world action model with the same object positions but without revision, reaches only 12.4\% BSR and 18.8\% FSR on \mbox{RMBench-$\Delta$} and 0.3 and 8.3\% on \mbox{RoboMemArena-$\Delta$}, so the recovery comes from revising the memory.

\begin{table}[t]
\centering\footnotesize
\setlength{\tabcolsep}{0pt}
\newcolumntype{S}{>{\centering\arraybackslash}p{30pt}}
\definecolor{bestbg}{HTML}{EBCB9A}\definecolor{bestbg}{HTML}{EBCB9A}\definecolor{secondbg}{HTML}{DDE7F1}
\caption{\textbf{\mbox{RMBench-$\Delta$}:} Avg.\ success (\%) $\pm$ standard error over five tasks; 50 rollouts/task, except 100 without a shift for RoboRAM, LatentMemWAM and published results. {\setlength{\fboxsep}{1pt}\colorbox{bestbg}{\textbf{Best}} and \colorbox{secondbg}{second best}} per column. $^{\dagger}$Multi-task model.}
\label{tab:main}
\begin{tabular}{@{}l@{\hspace{2pt}}S@{\hspace{2pt}}SS@{\hspace{2pt}}SS@{}}
\toprule
 & {\footnotesize No shift} & \multicolumn{2}{c}{\footnotesize Backward shift} & \multicolumn{2}{c}{\footnotesize Forward shift} \\
\cmidrule(l{2pt}r{2pt}){2-2}\cmidrule(l{2pt}r{2pt}){3-4}\cmidrule(l{2pt}r{0pt}){5-6}
\textbf{Method} & {\footnotesize TSR} & {\footnotesize TSR} & {\footnotesize BSR} & {\footnotesize TSR} & {\footnotesize FSR} \\
\midrule
DP & 4.2{\tiny$\pm$0.9} & 1.2{\tiny$\pm$0.7} & 1.2{\tiny$\pm$0.7} & 0.0{\tiny$\pm$0.0} & 0.0{\tiny$\pm$0.0} \\
Long-Context DP & 81.2{\tiny$\pm$1.9} & \cellcolor{secondbg}13.6{\tiny$\pm$1.8} & 4.4{\tiny$\pm$1.2} & 57.2{\tiny$\pm$2.6} & 4.0{\tiny$\pm$1.2} \\
EventVLA & 93.0{\tiny$\pm$1.1} & 2.4{\tiny$\pm$0.9} & 0.8{\tiny$\pm$0.6} & 30.8{\tiny$\pm$2.5} & 6.0{\tiny$\pm$1.3} \\
SimpleMemVLA$^{\dagger}$ & \cellcolor{bestbg}\textbf{99.6}{\tiny$\pm$0.3} & 8.4{\tiny$\pm$1.4} & 0.0{\tiny$\pm$0.0} & \cellcolor{bestbg}\textbf{94.0}{\tiny$\pm$1.5} & 4.4{\tiny$\pm$1.2} \\
LingBot-VA & 95.6{\tiny$\pm$0.8} & 2.8{\tiny$\pm$1.0} & 0.4{\tiny$\pm$0.4} & 34.0{\tiny$\pm$2.7} & 0.4{\tiny$\pm$0.4} \\
LatentMemWAM (Ours) & 92.4{\tiny$\pm$1.1} & 12.4{\tiny$\pm$1.4} & \cellcolor{secondbg}12.4{\tiny$\pm$1.4} & 47.6{\tiny$\pm$2.4} & \cellcolor{secondbg}18.8{\tiny$\pm$2.1} \\
\textbf{RoboRAM (Ours)} & \cellcolor{secondbg}96.0{\tiny$\pm$0.9} & \cellcolor{bestbg}\textbf{90.0}{\tiny$\pm$1.8} & \cellcolor{bestbg}\textbf{90.0}{\tiny$\pm$1.8} & \cellcolor{secondbg}84.8{\tiny$\pm$1.9} & \cellcolor{bestbg}\textbf{83.2}{\tiny$\pm$2.1} \\
\bottomrule
\end{tabular}
\vspace{6pt}
\caption{\textbf{\mbox{RoboMemArena-$\Delta$}:} Avg.\ success (\%) $\pm$ standard error over eight tasks; 50 rollouts/task in every setting. Shading as in \cref{tab:main}. $^{\dagger}$Multi-task model.}
\label{tab:main-rma}
\begin{tabular}{@{}l@{\hspace{2pt}}S@{\hspace{2pt}}SS@{\hspace{2pt}}SS@{}}
\toprule
 & {\footnotesize No shift} & \multicolumn{2}{c}{\footnotesize Backward shift} & \multicolumn{2}{c}{\footnotesize Forward shift} \\
\cmidrule(l{2pt}r{2pt}){2-2}\cmidrule(l{2pt}r{2pt}){3-4}\cmidrule(l{2pt}r{0pt}){5-6}
\textbf{Method} & {\footnotesize TSR} & {\footnotesize TSR} & {\footnotesize BSR} & {\footnotesize TSR} & {\footnotesize FSR} \\
\midrule
SimpleMemVLA$^{\dagger}$ & 58.3{\tiny$\pm$1.6} & 3.3{\tiny$\pm$0.9} & 2.0{\tiny$\pm$0.7} & 47.3{\tiny$\pm$1.9} & 5.8{\tiny$\pm$1.1} \\
LingBot-VA$^{\dagger}$ & \cellcolor{secondbg}78.8{\tiny$\pm$1.8} & \cellcolor{secondbg}11.5{\tiny$\pm$1.4} & \cellcolor{secondbg}7.5{\tiny$\pm$1.2} & \cellcolor{bestbg}\textbf{77.0}{\tiny$\pm$1.8} & 2.5{\tiny$\pm$0.7} \\
LatentMemWAM (Ours) & 77.0{\tiny$\pm$1.9} & 0.3{\tiny$\pm$0.2} & 0.3{\tiny$\pm$0.2} & 46.8{\tiny$\pm$1.3} & \cellcolor{secondbg}8.3{\tiny$\pm$0.8} \\
\textbf{RoboRAM (Ours)} & \cellcolor{bestbg}\textbf{79.8}{\tiny$\pm$1.9} & \cellcolor{bestbg}\textbf{38.0}{\tiny$\pm$2.3} & \cellcolor{bestbg}\textbf{32.5}{\tiny$\pm$2.2} & \cellcolor{secondbg}75.8{\tiny$\pm$2.0} & \cellcolor{bestbg}\textbf{70.3}{\tiny$\pm$2.0} \\
\bottomrule
\end{tabular}
\end{table}

\begin{figure}[t]
  \centering
  \includegraphics[width=\columnwidth]{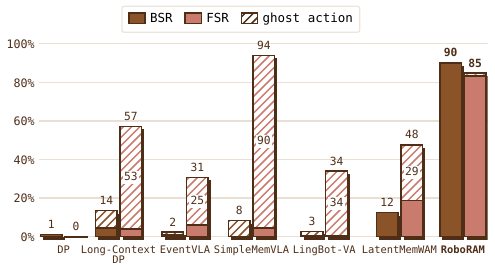}
  \caption{\textbf{Task success hides ghost actions.} Task success rate (\%) after a backward (brown) and a forward shift (light red), averaged over
  the five tasks of \mbox{RMBench-$\Delta$} (50 rollouts per cell), with its value on top. The solid part counts successes without a ghost action
  (BSR and FSR), and the hatched part successes with one.}
  \label{fig:ghost}
\end{figure}

\textbf{After a shift, memory baselines follow their clock instead of the scene, and task success alone hides it.} We pair each shifted
rollout on RoboMemArena tasks 5, 12, 13 and 22 with the unshifted rollout of the same seed. A policy that reads the scene skips the step that
another agent has done and finishes earlier, whereas one that follows a clock finishes as if nothing had shifted. In the 141 pairs in which
LingBot-VA succeeds with and without a forward shift, the two rollouts end within 10 steps of each other in 125 and on the same step in 79, and
SimpleMemVLA ends a median of 2 steps later after the shift. \methodname{} ends a median of 204 steps earlier, 27\% of an unshifted episode.
The arm paths show the same in both directions. Over the 150 steps after a shift, the end effector stays within a median of 0.5\,cm of its
unshifted path for LingBot-VA and 1.8 to 3.2\,cm for SimpleMemVLA, although the shifted object has moved by 58 to 81\,cm, whereas that of
\methodname{} departs by a median of 45\,cm toward the step that the scene now requires. Following the clock still passes the task success check
after a forward shift, since the repeated step is already done (\cref{fig:ghost}). On \mbox{RMBench-$\Delta$}, 61 to 99\% of the baselines'
forward successes include a ghost action, so by task success alone SimpleMemVLA would rank above \methodname{} (94.0 against 84.8\%). LingBot-VA
would match it on \mbox{RoboMemArena-$\Delta$} (77.0 against 75.8\%). After a backward shift, \methodname{} leads every baseline under
either count, by at least 76.4 points on \mbox{RMBench-$\Delta$} and 25.0 on \mbox{RoboMemArena-$\Delta$}.

\begin{figure}[t]
  \centering
  \includegraphics[width=\columnwidth]{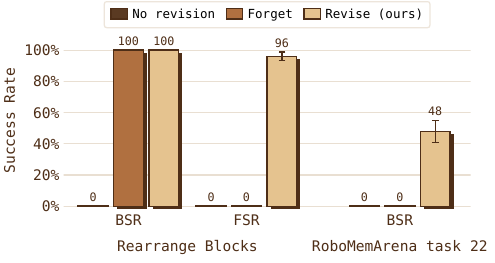}
  \caption{\textbf{Revising vs.\ forgetting the memory.} BSR and FSR (\%), 50 rollouts per bar, $\pm$ s.e.}
  \label{fig:forget}
\end{figure}

\textbf{Revising the memory, rather than keeping or clearing it, is what makes \methodname{} recover.} We keep \methodname{}'s policy, object
positions and shift detection, and change only what happens to the memory after a shift (\cref{fig:forget}). \emph{No revision} keeps the memory
as it is, \emph{Forget} clears it, as a detect-and-reset baseline would, and \emph{Revise} is \methodname{}. No revision reaches 0\% in all three settings of \cref{fig:forget}, so seeing where the objects are now is not enough without revising the memory. Forget helps only when the shift returns the scene to the start, where an
empty memory is correct. It thus matches Revise at 100\% BSR on Rearrange Blocks, but reaches 0\% FSR there against 96\% for Revise. On
RoboMemArena task 22, the robot pours twice and then puts cookies into a microwave, and the backward shift takes the cookies out again after
both pours. Only the memory records the pours, so Forget redoes the whole task and runs out of time in 48 of 50 episodes, with 0\% BSR,
whereas Revise rewinds only to the cookie step and reaches 48\%. The revisions are also precise. Across the five tasks of \cref{tab:main},
98.6\% of the first rewinds and fast-forwards land on an entry that shows the shifted scene.

\textbf{Each component of \methodname{} handles a distinct situation, and the memory is rarely revised when nothing has shifted.}
Without a shift, \methodname{} revises its memory in none of the rollouts on Put Back Block and Rearrange Blocks, where the variant without the gate does not revise it either in 20 rollouts per task. Revision itself makes the policy recover (\cref{tab:ablation}), since without it BSR is 0\% although the object positions are still observed. Without fast-forward into predicted entries, the memory stays behind a forward-shifted scene, so the robot redoes the other agent's step with a ghost action or later decisions rewind or pause by mistake, and FSR falls to 24 and 0\% (\cref{tab:ablation}). Without the gate, the localizer still detects the shift but revises again at later decisions. The learned localizer also outperforms the rule that provides its targets, applied at test time. On Put Back Block, where the block back on its mat looks the same at the start and at the end and only the memory knows that the button has not been pressed, the rule first jumps forward in 24 of 50 backward-shifted rollouts and recovers in 22 only after a later revision, about 90 steps later, whereas the learned localizer always rewinds at once.
Raising the gate bound up to twofold
changes no result by more than two rollouts.

\begin{figure*}[t]
  \centering
  \includegraphics[width=\textwidth]{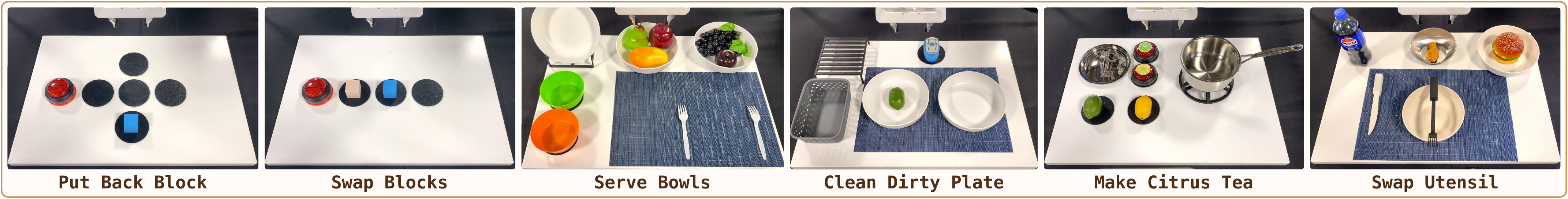}
  \caption{\textbf{Real-robot tasks.} The initial state of each of the six real-robot tasks on a  Franka Emika Panda robot arm.}
  \label{fig:real_tasks}
\end{figure*}

\begin{table}[t]
\centering\footnotesize
\renewcommand{\arraystretch}{0.95}
\setlength{\tabcolsep}{0pt}
\newcolumntype{S}{>{\centering\arraybackslash}p{28.0pt}}
\caption{\textbf{Ablation.} BSR and FSR in \% over 50 rollouts per task and shift, with one component of \methodname{} removed or changed at test time and the weights fixed.}
\begin{tabular}{@{}l@{\hspace{6pt}}SS@{\hspace{6pt}}SS@{}}
\toprule
 & \multicolumn{2}{c}{\makebox[56.0pt][c]{\scriptsize Put Back Block}} & \multicolumn{2}{c}{\makebox[56.0pt][c]{\scriptsize Rearrange Blocks}} \\
\cmidrule(l{1pt}r{\dimexpr1pt+6pt\relax}){2-3}\cmidrule(l{1pt}r{1pt}){4-5}
 & {\scriptsize BSR} & {\scriptsize FSR} & {\scriptsize BSR} & {\scriptsize FSR} \\
\midrule
\methodname{} & \textbf{100} & \textbf{100} & \textbf{100} & 96 \\
\midrule
w/o revision & 0 & 56 & 0 & 0 \\
detect and reset & \textbf{100} & 8 & \textbf{100} & 0 \\
w/o fast-forward & \textbf{100} & 24 & 98 & 0 \\
w/o gate & \textbf{100} & 86 & 98 & 70 \\
training-free localizer & 96 & 90 & 64 & 48 \\
\midrule
gate bound $\times1.5$ & \textbf{100} & \textbf{100} & \textbf{100} & \textbf{100} \\
gate bound $\times2$ & \textbf{100} & \textbf{100} & \textbf{100} & 98 \\
\bottomrule
\end{tabular}
\label{tab:ablation}
\end{table}

\textbf{\methodname{} makes an existing world action model reactive without changing its weights.} LingBot-VA~\cite{li2026causal} has what \methodname{} needs, a key--value cache as memory and predicted video latents that extend it into a pseudo-memory timeline. We freeze it after post-training and train only two small modules on its latents, a head reading object positions with their uncertainty, and the memory localizer, whose targets come from
LingBot-VA's own predictions. On Put Back Block and Rearrange Blocks, LingBot-VA alone reaches 0\% BSR and FSR, and all its successes after a
forward shift involve a ghost action. With \methodname{} it reaches 36\% BSR and 50\% FSR on Put Back Block and 96\% BSR and 24\% FSR on
Rearrange Blocks, 51.5\% on average. Since the policy that acts is unchanged, these gains come from revising its memory alone.

\textbf{\methodname{} does not cost the memory policy its performance when task progress does not shift.} \methodname{} adds
memory revision to LatentMemWAM, our memory policy inspired by MemoryWAM~\cite{yang2026memorywam}, and revision does not lower its success. On the same 100 unperturbed
RMBench episodes per task, \methodname{} completes 96.0\% against 92.4\% for LatentMemWAM, and at least as many on each of the five tasks. \methodname{} is also on par with the published results of SimpleMemVLA (99.6\%) and LingBot-VA (95.6\%)
(\cref{tab:main}).
On \mbox{RoboMemArena-$\Delta$}, it completes 79.8\% against 77.0\% for LatentMemWAM, 78.8\% for LingBot-VA~\cite{li2026causal}, and 58.3\% for SimpleMemVLA~\cite{yin2026simplememvla} (\cref{tab:main-rma}).

\subsection{Real-Robot Experiments}
\subsubsection{Tasks and Baselines}
\begin{table}[t]
\centering\footnotesize
\renewcommand{\arraystretch}{1.05}
\setlength{\tabcolsep}{0pt}
\definecolor{bestbg}{HTML}{EBCB9A}\definecolor{secondbg}{HTML}{DDE7F1}
\caption{\textbf{Real-robot results.} Avg. success rate (\%) $\pm$ standard error across six tasks, with 20 trials per task for each model under each test condition. Shading as in \cref{tab:main}.}
\label{tab:real-averages}
\newcolumntype{R}{>{\centering\arraybackslash}p{33pt}}
\begin{tabular*}{\columnwidth}{@{\extracolsep{\fill}}lRRRRR@{}}
\toprule
 & \multicolumn{1}{c}{No shift} & \multicolumn{2}{c}{Backward shift} & \multicolumn{2}{c}{Forward shift} \\
\cmidrule(lr){2-2}\cmidrule(lr){3-4}\cmidrule(lr){5-6}
Method & TSR & TSR & BSR & TSR & FSR \\
\midrule
DP & 20.8{\tiny$\pm$1.6} & \cellcolor{secondbg}2.5{\tiny$\pm$1.3} & \cellcolor{secondbg}2.5{\tiny$\pm$1.3} & 4.2{\tiny$\pm$1.6} & 4.2{\tiny$\pm$1.6} \\
Long-Context DP & 55.8{\tiny$\pm$3.5} & 0.0{\tiny$\pm$0.0} & 0.0{\tiny$\pm$0.0} & 2.5{\tiny$\pm$1.4} & 0.0{\tiny$\pm$0.0} \\
EventVLA & 76.7{\tiny$\pm$3.7} & 1.7{\tiny$\pm$1.1} & 1.7{\tiny$\pm$1.1} & \cellcolor{secondbg}31.7{\tiny$\pm$3.5} & \cellcolor{secondbg}15.0{\tiny$\pm$1.1} \\
LingBot-VA & 74.2{\tiny$\pm$3.7} & 0.8{\tiny$\pm$0.8} & 0.8{\tiny$\pm$0.8} & 17.5{\tiny$\pm$3.5} & 0.0{\tiny$\pm$0.0} \\
LatentMemWAM & \cellcolor{bestbg}\textbf{98.3}{\tiny$\pm$1.1} & 0.0{\tiny$\pm$0.0} & 0.0{\tiny$\pm$0.0} & 12.5{\tiny$\pm$3.0} & 2.5{\tiny$\pm$1.4} \\
\textbf{RoboRAM (Ours)} & \cellcolor{secondbg}97.5{\tiny$\pm$1.3} & \cellcolor{bestbg}\textbf{85.0}{\tiny$\pm$2.5} & \cellcolor{bestbg}\textbf{85.0}{\tiny$\pm$2.5} & \cellcolor{bestbg}\textbf{75.8}{\tiny$\pm$3.5} & \cellcolor{bestbg}\textbf{71.7}{\tiny$\pm$3.3} \\
\bottomrule
\end{tabular*}
\end{table}
We evaluate the six tasks of \cref{fig:real_tasks} against DP, Long-Context DP, EventVLA, LingBot-VA and LatentMemWAM. Every method is trained per task on the
same demonstrations, and every setting counts 20 trials, in which a person makes the shift by hand. SimpleMemVLA, whose released checkpoints we
use in simulation, is not trained here, because its training requires every demonstration frame to be labeled with its current sub-task in
language, an annotation that standard demonstrations lack and that no other method needs.

\textbf{\methodname{} carries over to a real robot and to multi-step shifts.} After a one-step shift (\cref{tab:real-averages}),
\methodname{} reaches 85.0\% BSR and 71.7\% FSR across the six tasks, 78.3\% on average, against at most 8.3\% for any baseline, and EventVLA's
15.0\% FSR comes from a single task, Serve Bowls. In everyday settings, agents can change several steps at once. On Clean
Dirty Plate, where the robot puts a lime into a basket, brushes the plate with a sponge and moves the plate to a rack, another agent may clean the plate for the robot, or make the clean plate dirty again by putting the lime back on it (\cref{fig:teaser}). On Make Citrus Tea, another agent may put both the lime and the lemon into the pot, or take both out. We test such \emph{multi-step shifts}, which undo or complete two consecutive steps at once, on these two tasks. They move the scene farther along the timeline, so the memory must rewind past several steps or fast-forward deep into its predicted entries, and a fast-forward that reaches the last predicted entry continues with at most three further predictions. Every baseline, LatentMemWAM included, fails in every multi-step shifted trial on both tasks, whereas \methodname{} reaches 35\% BSR and 30\% FSR on Clean Dirty Plate and 90\% BSR and FSR on Make Citrus Tea, 61.3\% on average. Without a shift, it matches LatentMemWAM, the world action model it builds on (97.5 against 98.3\%), and outperforms every other
baseline, the best of which, EventVLA, reaches 76.7\% (\cref{tab:real-averages}).

\textbf{\methodname{} is fast enough to deploy on a real robot.} On the robot's RTX 3090, reading and revising the memory, generating the next 8 actions and writing the executed frames take \methodname{} 25 to 26\,ms per action, as much as DP and EventVLA at 23 to 25\,ms. Object tracking, run in parallel, raises this to 37 to 92\,ms, still below LingBot-VA's 121 to 125\,ms.

\section{Conclusion and Limitations}
\label{sec:conclusion}
We studied task-progress shifts, in which another agent undoes or completes steps of a task while a robot works. \methodname{} treats the memory as a pseudo-memory timeline of observed and predicted steps and learns from standard demonstrations alone to rewind or fast-forward it to the step that the scene shows. In simulation and on a real robot, it recovers from shifts in both directions where state-of-the-art memory policies, VLAs and WAMs fail, keeps its success when nothing shifts, and makes an existing WAM reactive without changing its weights.
\textbf{Limitations.} \methodname{} relies on object positions from an off-the-shelf tracker, so a tracking error can delay a revision or move the memory to the wrong step. Learning object states inside the world action model itself would remove this dependence.

\clearpage
{
    \small
    \bibliographystyle{ieeenat_fullname}
    \bibliography{main}
}
\appendix
\twocolumn[{%
\centering
{\Large\textbf{\thetitle}\par\vspace{0.25em}Supplementary Material\par}\vspace{0.6em}%
\includegraphics[width=0.85\textwidth]{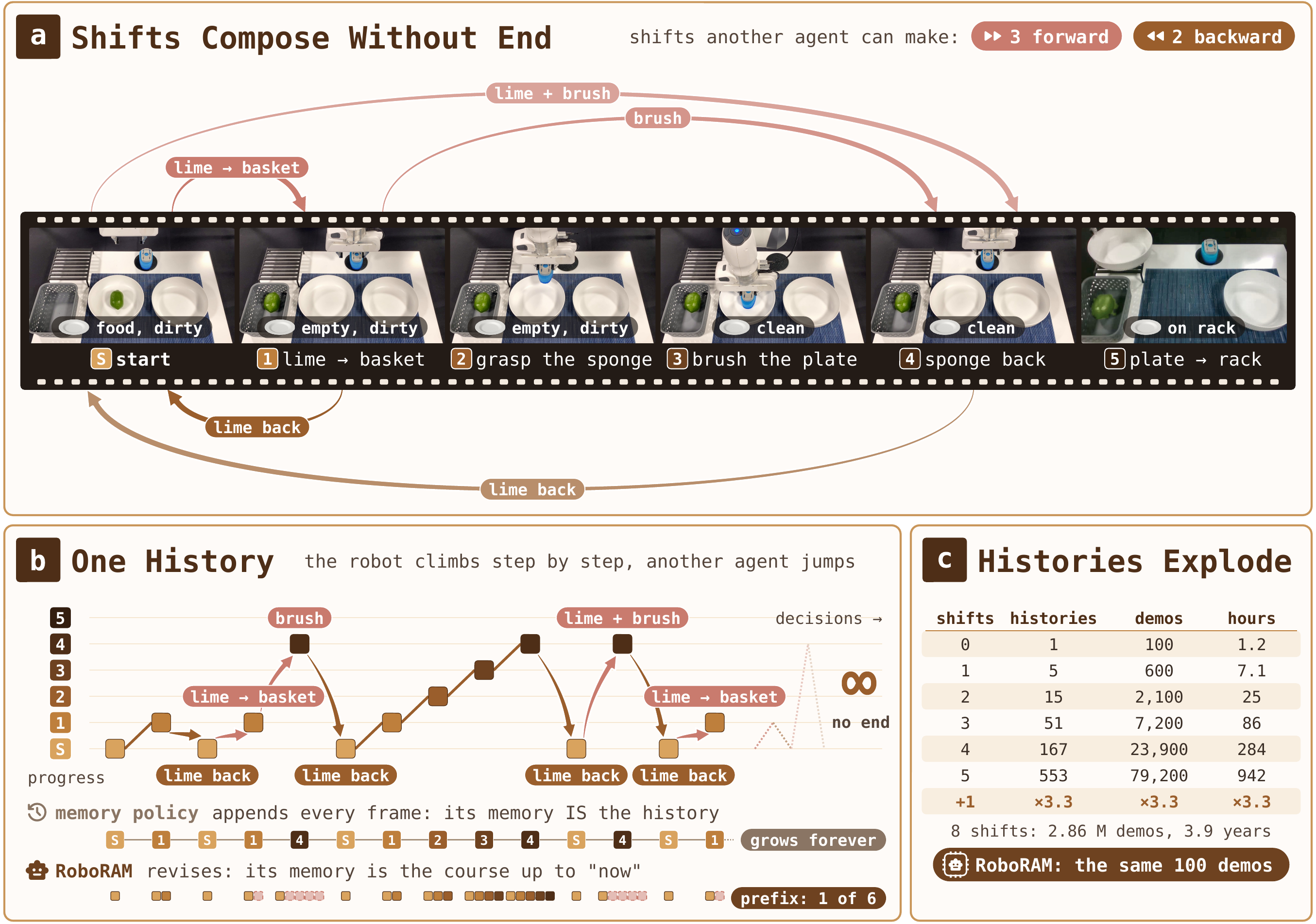}
\captionof{figure}{\textbf{Composed shifts on Clean Dirty Plate.} (a)~The five shifts another agent can make, forward above and backward below.
(b)~One history of eight shifts and the memories of a memory policy and of \methodname{} (solid entries observed, rose dashed entries predicted).
(c)~Histories, and the demonstrations needed to cover them, grow by about 3.3 per additional shift.}
\label{fig:combinatorial}
\vspace{0.6em}}]

\section{Problem: task-progress shifts}
\label{sec:supp-problem}
\label{sec:details}

\subsection{Demonstrations cannot cover every shift}
\label{sec:supp-composed}

Collecting demonstrations of task-progress shifts does not solve the problem, because shifts compose. Another agent may shift task progress
again after the robot has responded, and repeatedly thereafter.

\paragraph{Histories.}
We define a \emph{history} as a finite sequence of task-progress shifts in which the robot takes at least one decision between consecutive shifts
(\cref{fig:combinatorial}b). Since the robot may be at any of several steps when the next shift occurs, and several shifts may be admissible
at each, the number of histories grows exponentially with their length.

\paragraph{Number of histories.}
In Clean Dirty Plate, another agent can make the five shifts of \cref{fig:combinatorial}a, each of which leaves the scene at the start, after the
lime step or after the brushing. There are 5 histories with one shift, 15 with two, 51 with three, 167 with four and 553 with five, and each
additional shift eventually multiplies their number by about 3.3 (\cref{fig:combinatorial}c). These counts are a lower bound,
since they ignore when within a step each shift occurs.

\paragraph{Demonstrations required.}
A memory policy that only appends retains the entire history in its memory, and imitation constrains its actions only after histories that its
demonstrations contain (\cref{sec:problem}). Covering every history of at most a given number of shifts as densely as the unshifted
task therefore multiplies the demonstrations by the number of such histories. Starting from the 100 standard demonstrations recorded
for Clean Dirty Plate, histories of at most one, two, three and four shifts require 600, 2{,}100, 7{,}200 and 23{,}900 demonstrations, and histories
of at most eight shifts about 2.86 million, which at the mean demonstration length of 42.8\,s amount to 7, 25, 86, 284 and about 33{,}950 hours of
teleoperation, the last nearly four years. No finite set of demonstrations covers histories of every length. \methodname{}
instead learns from the 100 demonstrations without a shift and, after each shift, revises its memory back onto the course of the task
(\cref{fig:combinatorial}b), so its demonstrations do not grow with the number of shifts.
\section{Extended related work}
\label{sec:supp-concurrent}
\begin{figure*}[t]
\centering
\includegraphics[width=\linewidth]{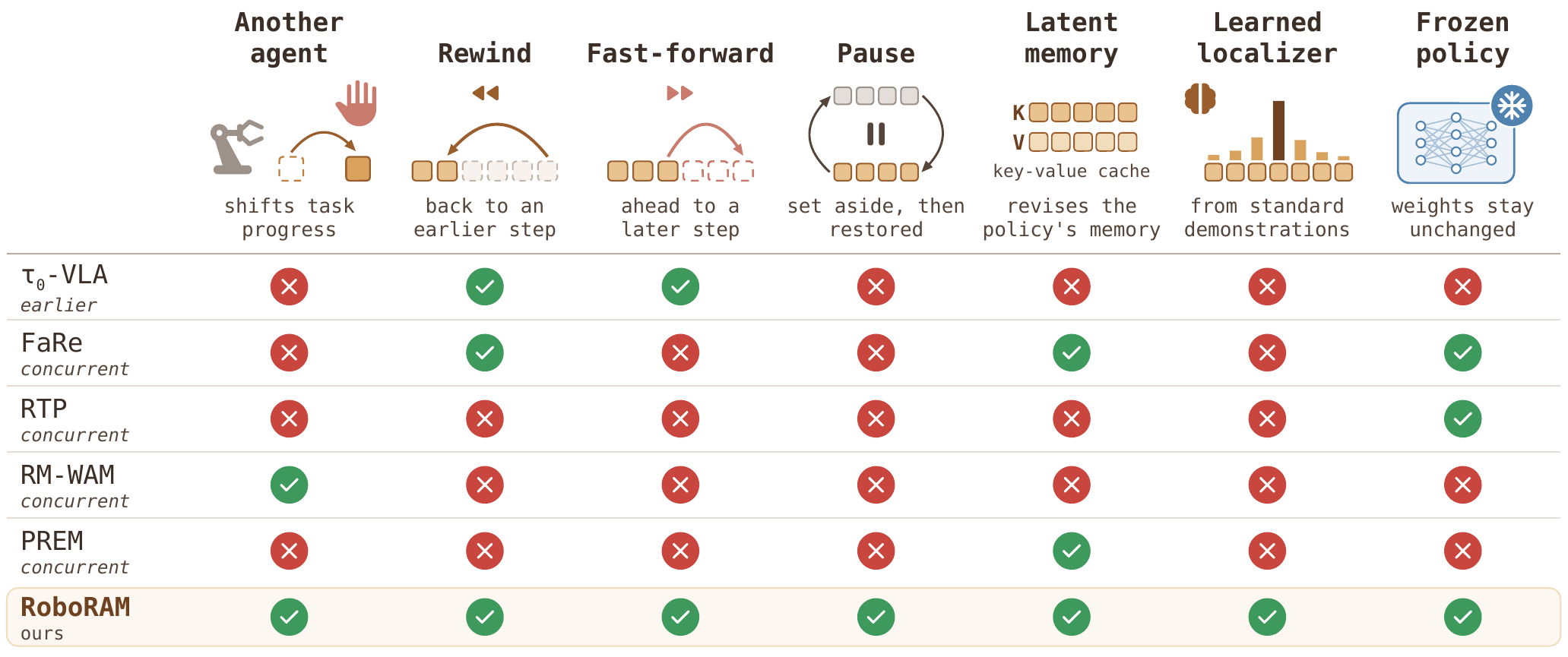}
\caption{\textbf{What \methodname{} can do that the closest methods cannot.} \methodname{} has all seven capabilities, and no other method in it
has more than three. The rows are the closest earlier method, $\tau_0$-VLA~\cite{cai2026tau_0}, and the concurrent methods closest to \methodname{} in
mechanism (FaRe~\cite{li2026fare}, RTP~\cite{liu2026revision}), training data (RM-WAM~\cite{miao2026rm}) and use of prediction error (PREM~\cite{huang2026prem}). The columns say, from left to right,
whether a method handles task progress shifted by another agent (property~(i) in~\cref{sec:supp-concurrent}), rewinds and fast-forwards the memory (ii, iv), pauses by
setting the memory aside and restoring it later, revises the policy's latent memory (iii), decides where the memory moves with a localizer
learned from standard demonstrations alone (v), and keeps the policy frozen (vi).}
\label{fig:capabilities}
\end{figure*}

\begin{figure*}[t]
\centering
\includegraphics[width=\linewidth]{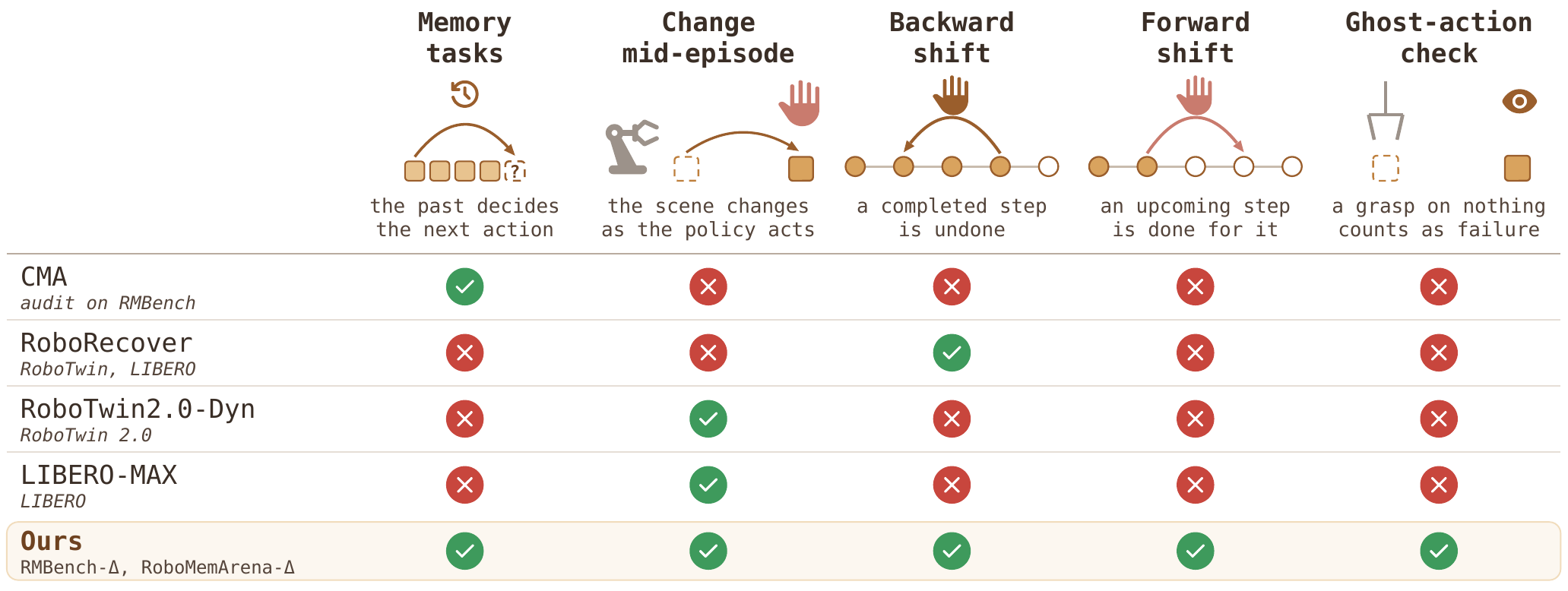}
\caption{\textbf{How our benchmarks differ from concurrent ones.} The rows are recent simulation benchmarks and audits that change the scene or
the memory of a robot policy. The Counterfactual Memory Audit (CMA)~\cite{zhang2026memory} and our benchmarks use memory tasks,
and RoboRecover~\cite{li2026roborecover}, RoboTwin2.0-Dyn~\cite{park2026dynamic} and LIBERO-MAX~\cite{zhang2026libero} use tasks
that need no memory. A check marks \emph{memory tasks}, in which the past decides the next action, a \emph{change mid-episode} while the
policy acts, a \emph{backward shift} that undoes a completed step, a \emph{forward shift} that completes an upcoming step for the robot, and
a \emph{ghost-action check} that counts a grasp on nothing where a shifted object was as a failure.}
\label{fig:benchmarks}
\end{figure*}

To our knowledge, \methodname{} is the first method that learns from standard demonstrations alone to move a robot policy's own latent memory
to the step where the scene now is, backward or forward, after another agent shifts task progress. \Cref{fig:capabilities} makes the gap
concrete: \methodname{} has all seven capabilities it lists, and no other method in it has more than three. \Cref{fig:benchmarks} shows the same for
the benchmarks.

\paragraph{Six properties.}
Six properties together define \methodname{}. (i)~It handles task progress changed by another agent, not by the policy. (ii)~It handles both
directions, a completed step undone and an upcoming step completed for the robot. (iii)~It revises the latent memory the policy acts from, not a
plan, a symbolic or textual task state, or the weight given to the memory. (iv)~It moves the end of that memory back to an observed step or ahead
to a step the policy predicted. (v)~A learned module decides where the memory ends. It learns from standard demonstrations alone, with targets
from the frozen model's own predictions. (vi)~This localizer is trained on the frozen policy and never changes its weights. Among the methods
of \cref{fig:capabilities}, FaRe~\cite{li2026fare} and PREM~\cite{huang2026prem} revise a latent memory (iii), but only FaRe~\cite{li2026fare} moves it to another step (iv), and FaRe~\cite{li2026fare} only cuts it back or
clears it. None of them revises a latent memory in both directions (ii), and none learns to which step of an episode a latent memory should move.

\paragraph{How the closest methods differ.}
\begin{itemize}\setlength{\itemsep}{2pt}
\item \textbf{$\tau_0$-VLA}~\cite{cai2026tau_0}. Its high-level planner keeps a textual memory and learns to correct it when it lags behind,
runs ahead of or misrepresents the robot's progress. It corrects text, not the memory the action policy acts from, it is
not designed for changes made by another agent, and it needs demonstrations annotated with tasks, stages and subtasks. \methodname{} corrects the policy's own latent memory after another
agent's change, needs no annotated demonstrations, and fills the memory ahead with steps the model predicts.
\item \textbf{FaRe}~\cite{li2026fare}, the closest in mechanism. It needs an external 8B progress model and hand-set thresholds: when that model reports a
persistent lack of progress, it keeps the key--value cache of a frozen world action model, cuts it back to a step, before its last measured
progress, whose images are closest to the present, or clears it. It is designed for the policy's own errors, and none of its options extends the cache beyond the executed
history, so after a forward shift it can only keep or cut back a memory that lags behind the scene, or clear it and lose what it held. \methodname{}
learns where the memory belongs, with no external progress model and no hand-set threshold on image distance, and also moves it ahead. 
\item \textbf{RTP}~\cite{liu2026revision}. It revises the predicted visual plan of a fine-tuned LingBot-VA~\cite{li2026causal} when execution disagrees with it, for example after
holds, delays or a 2\,cm shift of the target, with a revision module and a mode estimator learned from the policy's own successful and failed
rollouts. Its observed history is only appended, and it presents its plan revision as a direction toward a ``persistent, revisable,
feedback-driven WAM memory''. \methodname{} revises the observed history itself, after another agent changes the task's progress, and needs no rollouts to learn.
\item \textbf{RM-WAM}~\cite{miao2026rm}, the closest in training data. It also pairs the present with histories from other positions or episodes of
standard demonstrations, but its learned gates only scale how strongly the memory enters the prediction, with one fixed target for each way of
sampling the history. No operation moves the memory, and the gates are trained jointly with the world action model. It is tested with another
agent in a single real-robot task, in which a person intervenes, for example by removing an object already placed. \methodname{} learns to which step the memory belongs and moves it
there, backward or forward, without changing the policy.
\item \textbf{PREM}~\cite{huang2026prem}. It compares the present with the prediction made at the previous decision, as the gate of \methodname{} does, but
at test time uses the error only to set how much of the present it writes into its recurrent visual memory. It never moves its memory to another step, names no other agent,
and is fine-tuned jointly with its policy. In \methodname{}, the forecast error only permits a revision, and a learned localizer decides where
the memory moves.
\end{itemize}

\paragraph{Benchmarks.}
Among the benchmarks of \cref{fig:benchmarks}, only \mbox{RMBench-$\Delta$} and \mbox{RoboMemArena-$\Delta$} shift task progress in both directions
while a memory policy acts, and only they check for ghost actions. CMA~\cite{zhang2026memory}, the only other one with memory
tasks, replaces a policy's memory on RMBench tasks but keeps the scene, so it tests whether the policy follows its memory, not whether it notices a
changed scene. RoboRecover~\cite{li2026roborecover}, RoboTwin2.0-Dyn~\cite{park2026dynamic} and LIBERO-MAX~\cite{zhang2026libero}
use tasks that need no memory. RoboRecover starts the policy at replayed deviation states, and the other two change the scene during execution,
for instance by moving the target, without changing the task's progress.
\clearpage
\makeatletter\global\@colroom\textheight\global\vsize\textheight\makeatother

\twocolumn[{%
\centering\small
\renewcommand{\arraystretch}{1.38}\setlength{\aboverulesep}{0pt}\setlength{\belowrulesep}{0pt}
\setlength{\tabcolsep}{6.3pt}
\definecolor{archgrp}{HTML}{ECECEC}
\begin{tabular}{llll}
\toprule
\textbf{Module} & \textbf{Input $\rightarrow$ output} & \textbf{Layers} & \textbf{Param.}\\
\midrule
\rowcolor{archgrp}\multicolumn{3}{l}{\textbf{Latent world action model} \quad (stage~1, then frozen)} & \textbf{62.3\,M}\\
Image encoders $E_\omega$ & head image, wrist-image mosaic $\rightarrow$ $2\times64$ features & 2$\times$ResNet-18~\cite{he2016deep}, spatial softmax & 22.4\,M\\
Token maps $E_\omega$ & features and $q_t$, $x_t$ and $p_t-p_{t-1}$, $a_{t-1}$ $\rightarrow$ $\tau_t$ (256) & linear & 0.04\,M\\
Transformer $\operatorname{Tr}_\omega$ & $\tau_t$, memory $m_t$ $\rightarrow$ $h_t$ (256) & 6 causal layers, 4 heads, NoPE~\cite{kazemnejad2023impact} & 4.7\,M\\
Bottleneck $\mathcal B_\omega$ & $h_t$ $\rightarrow$ code $\eta_t$ (64) & linear & 0.03\,M\\
Action head $\hat\epsilon_\omega$ & noised chunk of $N=16$ actions, $\tau^o_t$, $\eta_t$ $\rightarrow$ noise & 1-D U-Net~\cite{ronneberger2015u}, 128/256/512 channels & 18.6\,M\\
Timeline head & $\operatorname{sg}(h_t)$ $\rightarrow$ $\hat\tau$, $\hat\rho$ and $\log\sigma^2$ at 25 horizons & two MLPs, hidden 1{,}024 and 512 & 8.8\,M\\
State head & $\operatorname{sg}(h_t)$ $\rightarrow$ $\tilde\rho_t$ (20) & MLP, hidden 512 & 0.4\,M\\
World head $g_\omega$ & $\tau^o_t$, $\eta_t$ $\rightarrow$ one Wan2.2 latent frame ($6{,}912$) & MLP, hidden 512 & 7.3\,M\\
\midrule
\rowcolor{archgrp}\multicolumn{3}{l}{\textbf{Memory localizer} \quad (stage~2)} & \textbf{1.9\,M}\\
Query & $x_t$, $p_t-p_{t-1}$, $q_t$ $\rightarrow$ query token & MLP, hidden 256 & 0.07\,M\\
Scores $P_\psi$ & timeline entries, $h^{\mathrm q}_t$ $\rightarrow$ $P_\psi$ over $\Theta$, path weights $\alpha_i$ & 2 self-attention layers, maps $\operatorname{key}$, $\operatorname{query}$ & 1.8\,M\\
\bottomrule
\end{tabular}
\captionof{table}{\textbf{Architecture of \methodname{}}, for Put Back Block, which tracks two objects, so a measurement $\rho_t=(x_t,q_t)$ has 20
numbers. Numbers in parentheses are output sizes. Images are $256{\times}256$, cropped to $216{\times}216$. The world head is used only in
training. Layers that read or predict the object state grow with the number of objects, and $\operatorname{sg}$ is the stop-gradient.}
\label{tab:supp-arch}
\vspace{1.2em}}]

\section{Method details}
\label{sec:supp-method}
\subsection{Latent world action model}
\label{sec:supp-model}

\Cref{tab:supp-arch} lists every module with its input, output, layers and number of parameters.

\paragraph{Token.}
The token of \cref{eq:token} is $\tau_t=\tau^o_t+W_x\Gamma(p_t,v_t,p_t-p_{t-1})$, where $\Gamma$ stacks the position, visibility
and displacement since the previous step of every object. The \emph{observation token} $\tau^o_t=W_o\operatorname{Enc}(o_t)+W_a a_{t-1}$,
from the images, $q_t$ and the previous action, is the part of the token that does not use the tracker. The $W$ are the linear token maps of
\cref{tab:supp-arch}.

\paragraph{Heads.}
The action head of \cref{eq:act} and the world head read the observation through its token $\tau^o_t$ and the memory through
the code $\eta_t$, so the object state and the memory reach the actions only through $\eta_t$. The noised chunk of the next $N$ actions $A_t$ is
$\tilde A_t^{j}=\sqrt{\bar\alpha_j}A_t+\sqrt{1-\bar\alpha_j}\,\epsilon$ with signal fraction $\bar\alpha_j$. The timeline head predicts at the
horizons $\delta\in\{1,8,16,\dots,192\}$, so $F=192$, and interpolates linearly between them. The world head predicts the Wan2.2-TI2V-5B~\cite{wan2025wan}
latent of the 17-frame head-camera clip that starts at step $t+1$, five frames of $48{\times}12{\times}12$ values at 192 pixels, each predicted
separately, and its loss reaches the memory through the bottleneck.

\paragraph{Localizer.}
The query token embeds $x_t$, $p_t-p_{t-1}$ and $q_t$, with no image and no previous action, and the frozen transformer reads it after
the whole timeline, the memory followed by its $F$ predicted entries, giving $h^{\mathrm q}_t$. The feature $\upsilon_\ell$ of entry $\ell$
comes from the entry's state, the measurement that the state head reads from it, and a flag that tells observed from predicted entries.

\paragraph{Tracker.}
Grounding DINO~\cite{liu2024grounding} finds each object in the first frame from a text prompt, with a box threshold of 0.25 and a text threshold of 0.2.
SAM~2.1 (Hiera-L)~\cite{ravi2025sam}, in bfloat16, then segments the object and propagates its mask frame by frame. An object is visible ($v_t^k=1$) when
SAM~2.1's object score is positive and its mask is not empty. Objects with the same prompt are ordered by their position in the first frame.

\subsection{Memory localizer}
\label{sec:supp-loc}

\paragraph{Scores.}
The localizer scores every entry $\ell$ of the timeline and the answer $\varnothing$,
\begin{align}
&P_\psi(\ell\mid m,q,x)\propto\exp\bigl(\operatorname{key}(\upsilon_\ell)^{\top}\bar h_t+\nu(\upsilon_\ell)+\phi(\ell-n)\bigr),\nonumber\\
&P_\psi(\varnothing\mid m,q,x)\propto\exp\bigl(k_\varnothing^{\top}\bar h_t+\nu_\varnothing\bigr),\label{eq:score}
\end{align}
normalized over $\Theta$. Here $\upsilon_\ell$ is the feature of entry $\ell$, $h^{\mathrm q}_t$ is the frozen transformer's output for the query
token (\cref{sec:supp-model}), $\bar h_t=\operatorname{query}(h^{\mathrm q}_t,x_t,q_t)$, and $\operatorname{key}$,
$\operatorname{query}$, $\nu$, $\phi$, $k_\varnothing$ and $\nu_\varnothing$ are learned.

\paragraph{Consistency test.}
Let $\mathcal K$ be the set of tracked objects, $\mu^k_\ell$ and $\sigma^k_\ell$ the position and standard deviation that the timeline head
predicts for object $k$ at entry $\ell$, from entry $\ell-1$ for an observed entry and from entry $n$ for a predicted one, and
$\|\cdot\|^2_{\sigma^2}$ the squared distance with each image coordinate divided by its variance. The normalized error of each object in
\cref{eq:consistent} and the bound $\kappa$ are
\begin{align}
&\zeta^k_{\ell}=\bigl\|p_t^k-\mu^k_{\ell}\bigr\|^2_{(\sigma^k_{\ell})^2},\label{eq:zeta}\\
&\kappa=\max_{\xi\in\mathcal D}\ \max_{\ell,\,k:\,v_\ell^k=1}\bigl\|p_\ell^k-\mu^k_{\ell}\bigr\|^2_{(\sigma^k_{\ell})^2},\label{eq:kappa}
\end{align}
where in \cref{eq:kappa} $p^k_\ell$ is the position recorded at step $\ell$ of $\xi$ and $\mu^k_{\ell},\sigma^k_{\ell}$ are predicted along
$\xi$, so every visible object of every demonstration frame passes the test at its expected entry. Entry $\ell$ moves the objects
\begin{equation}
\label{eq:moved}
\mathcal K_\ell=\bigl\{k\in\mathcal K:\|\mu^k_{\ell}-\mu^k_{n+1}\|^2_{(\sigma^k_{\ell})^2+(\sigma^k_{n+1})^2}>\kappa\bigr\},
\end{equation}
none for $\ell=n+1$, and it is consistent if
\begin{equation}
\label{eq:consistent-exact}
\max_{k\in\mathcal K_\ell}\zeta^k_{\ell}\le\kappa\quad\text{and}\quad\max_{k\in\mathcal K\setminus\mathcal K_\ell}\zeta^k_{n+1}\le\kappa.
\end{equation}
Each object that entry $\ell$ moves is thus tested against entry $\ell$, and every other object against entry $n+1$, where the memory expects it next and where entry $\ell$ predicts it too, up to the bound. An object that is not visible is tested at the place where it was last tracked, since the arm can hide an object without moving it.
\section{Experiments in simulation}
\label{sec:supp-sim}
\label{sec:supp-overview}

\subsection{Benchmarks, protocol and metrics}
\label{sec:supp-bench}
\label{sec:supp-protocol}
\Cref{tab:why-memory} gives, for each task of both benchmarks, what the memory must hold in an episode without a shift, why the scene does not
show it, and the decision that depends on it.%
\begin{table*}[t]
\centering\footnotesize
\definecolor{whygrp}{HTML}{E6E6E6}\definecolor{whyrow}{HTML}{F6F6F4}\definecolor{whyname}{HTML}{5A5A5A}
\definecolor{tagHbg}{HTML}{DCE5F1}\definecolor{tagHfg}{HTML}{274468}\definecolor{tagAbg}{HTML}{F3E8CC}\definecolor{tagAfg}{HTML}{684C0F}\definecolor{tagTbg}{HTML}{EADFF1}\definecolor{tagTfg}{HTML}{4F366C}
\newcommand{\whytag}[3]{{\setlength{\fboxsep}{1.3pt}\colorbox{#1}{\rule[-0.45ex]{0pt}{1.85ex}\textcolor{#2}{\scriptsize\textsf{#3}}}}}
\newcommand{\whyhidden}{\whytag{tagHbg}{tagHfg}{hidden}}\newcommand{\whyalike}{\whytag{tagAbg}{tagAfg}{look-alike}}\newcommand{\whytrace}{\whytag{tagTbg}{tagTfg}{no trace}}
\renewcommand{\arraystretch}{1.6}\setlength{\aboverulesep}{0pt}\setlength{\belowrulesep}{0pt}
\setlength{\tabcolsep}{3.5pt}
\begin{tabular}{llll}
\toprule
\textbf{Task} & \textbf{What the memory holds} & \textbf{Why the scene does not show it} & \textbf{What it decides}\\
\midrule
\rowcolor{whygrp}\multicolumn{4}{l}{\textbf{\mbox{RMBench-$\Delta$}}\quad{\color{whyname}five tasks}}\\
\textbf{Put Back Block} & Its start mat; whether pressed & \whyalike\enspace the four mats\quad \whytrace\enspace the press & Mat to return to; whether to press\\
\rowcolor{whyrow}\textbf{Rearrange Blocks} & Block it moved; whether pressed & \whyalike\enspace the blocks and mats\quad \whytrace\enspace the press & Block for the middle; press or not\\
\textbf{Swap Blocks} & Moves done; which tray was empty & \whyalike\enspace the two blocks & Next block and tray; when to press\\
\rowcolor{whyrow}\textbf{Cover Blocks} & The color under each cover & \whyhidden\enspace the covered colors & Order of lifting the covers\\
\textbf{Blocks Ranking Try} & The orders already tried & \whytrace\enspace the press that tests an order & Next swap; whether to press\\
\rowcolor{whygrp}\multicolumn{4}{l}{\textbf{\mbox{RoboMemArena-$\Delta$}}\quad{\color{whyname}eight tasks}}\\
\textbf{Task 5} & The empty drawer; drawers checked & \whyhidden\enspace insides\quad \whyalike\enspace closed drawers & Next drawer; the butter's drawer\\
\rowcolor{whyrow}\textbf{Task 11} & Cookies in the top drawer & \whyhidden\enspace the closed top drawer & Butter next, into the middle drawer\\
\textbf{Task 12} & Cookies in the middle drawer & \whyhidden\enspace inside the drawer & Pudding next, into the same drawer\\
\rowcolor{whyrow}\textbf{Task 13} & Cookies in the middle drawer & \whyhidden\enspace inside the drawer & Butter next, into the same drawer\\
\textbf{Task 14} & The drawer holding the cookies & \whyhidden\enspace the closed top drawer & Pudding into another drawer\\
\rowcolor{whyrow}\textbf{Task 21} & Butter's place in the microwave & \whyhidden\enspace inside the microwave & Where the pudding goes\\
\textbf{Task 22} & How many pours so far & \whytrace\enspace each pour & Pour again, or the cookies next\\
\rowcolor{whyrow}\textbf{Task 24} & Cookies' place in the microwave & \whyhidden\enspace inside the microwave & Where the popcorn goes\\
\bottomrule
\end{tabular}
\caption[Why each task needs memory]{\textbf{Why each task needs memory.} For each task, what the memory must hold in an episode without a shift, why the
scene does not show it, and the decision that depends on it. The tag gives the reason: \whyhidden{} out of the camera's view, \whyalike{} objects or places
that look the same, \whytrace{} an event that leaves the scene unchanged. The \mbox{RoboMemArena-$\Delta$} rows follow RoboMemArena's own description of what
each task must remember~\cite{lei2026robomemarena}, and \cref{fig:rma-shift-rows,fig:rma-shift-rows-t11,fig:rma-shift-rows-t12,fig:rma-shift-rows-t13,fig:rma-shift-rows-t14,fig:rma-shift-rows-t21,fig:rma-shift-rows-t22,fig:rma-shift-rows-t24} name the tasks.}
\label{tab:why-memory}
\end{table*}%

\paragraph{Staging a shift.}
A rendered hand makes every shift. It carries or slides the object over several steps, so the change is visible and never a teleport. The
hand's start time is drawn per episode from the seed, uniformly within a window that opens and closes at task events, such as the release of
an object, the press of a button or the robot's next grasp. In \mbox{RMBench-$\Delta$}, a backward shift starts after the robot releases the
shifted object and ends before its next press or grasp. A forward shift starts before the robot reaches the object. In
\mbox{RoboMemArena-$\Delta$}, a backward shift starts once the benchmark's check of the shifted step holds and the robot's open gripper is at
least 10\,cm away. A forward shift starts once the preceding step is complete. \Cref{tab:supp_sim_shifts} lists what another agent moves in each shift.
\begin{table*}[t]
\centering\footnotesize
\renewcommand{\arraystretch}{1.35}\setlength{\aboverulesep}{0pt}\setlength{\belowrulesep}{0pt}
\setlength{\tabcolsep}{4.6pt}
\begin{tabular}{l>{\columncolor{bwdbg}}l>{\columncolor{fwdbg}}l}
\toprule
\textbf{Task} & \bwdtag{}\enspace another agent moves, after the robot's step & \fwdtag{}\enspace another agent moves, before the robot's grasp\\
\midrule
\rowcolor{grpbg}\multicolumn{3}{l}{\textbf{\mbox{RMBench-$\Delta$}, one-step shifts}}\\
Put Back Block & the block from the center back to its mat, before the press & the block from its mat to the center\\
Rearrange Blocks & the first block from the empty mat back to where it started & the first block onto the empty mat\\
 & & (a) the first block into the empty tray\\
\multirow{-2}{*}{Swap Blocks} & \multirow{-2}{*}{the first block from the empty tray back to its own tray} & (b) the second block into the freed tray\\
Cover Blocks & the cover of the left block back to its place & a cover onto the middle block\\
Blocks Ranking Try & the block at the spare spot back to where it was & the middle block to the spare spot, after the first press\\
\midrule
\rowcolor{grpbg}\multicolumn{3}{l}{\textbf{\mbox{RoboMemArena-$\Delta$}, one-step shifts}}\\
Task 5 & the butter from the drawer back to where it stood & the butter into the empty drawer\\
Task 11 & the butter from the middle drawer back to where it stood & the cookies into the top drawer\\
Task 12 & the cookies from the middle drawer back to where they stood & the cookies into the middle drawer\\
Task 13 & the cookies from the middle drawer back to where they stood & the cookies into the middle drawer\\
Task 14 & the pudding from its drawer back to where it stood & the cookies into the top drawer\\
Task 21 & the butter from the microwave back to where it stood & the butter into the microwave\\
Task 22 & the cookies from the microwave back to where they stood & the cookies into the microwave\\
Task 24 & the cookies from the microwave back to where they stood & the cookies into the microwave\\
\bottomrule
\end{tabular}
\caption{\textbf{Task-progress shifts in simulation}, what another agent moves in each. A backward shift comes after the robot has completed the
step and undoes it, and a forward shift comes before the robot grasps the object of the step and completes it. Swap Blocks has two one-step
forward shifts, in which another agent makes the robot's first move (a) or its second move (b).}
\label{tab:supp_sim_shifts}
\end{table*}

\paragraph{Scoring.}
Every success rate starts from the task's own success check. Where needed, we adapt the check so that a step undone by another agent is neither
missed nor fails the episode on its own, and so that the policy never gets credit for a step that another agent completed. A \emph{ghost
action} is a grasp on nothing where a shifted object used to be, that is, a gripper closing after the shift starts and before the task
succeeds, within 5\,cm of where a shifted object was when the shift started and with no object within 5\,cm of the gripper. A closing after which the gripper holds
an object is a grasp, not a ghost action. TSR counts every success, while BSR and FSR count a success after a backward or a forward shift only
if the episode has no ghost action. RoboMemArena~\cite{lei2026robomemarena} counts a step as done once its check holds, even if the step
is undone later, so we also require every counted item to still be in place at the end of the episode.

\subsection{\mbox{RMBench-$\Delta$}: tasks and shifts}
\label{sec:supp-rmbench}
\begin{figure*}[p]
\centering
\includegraphics[width=\textwidth]{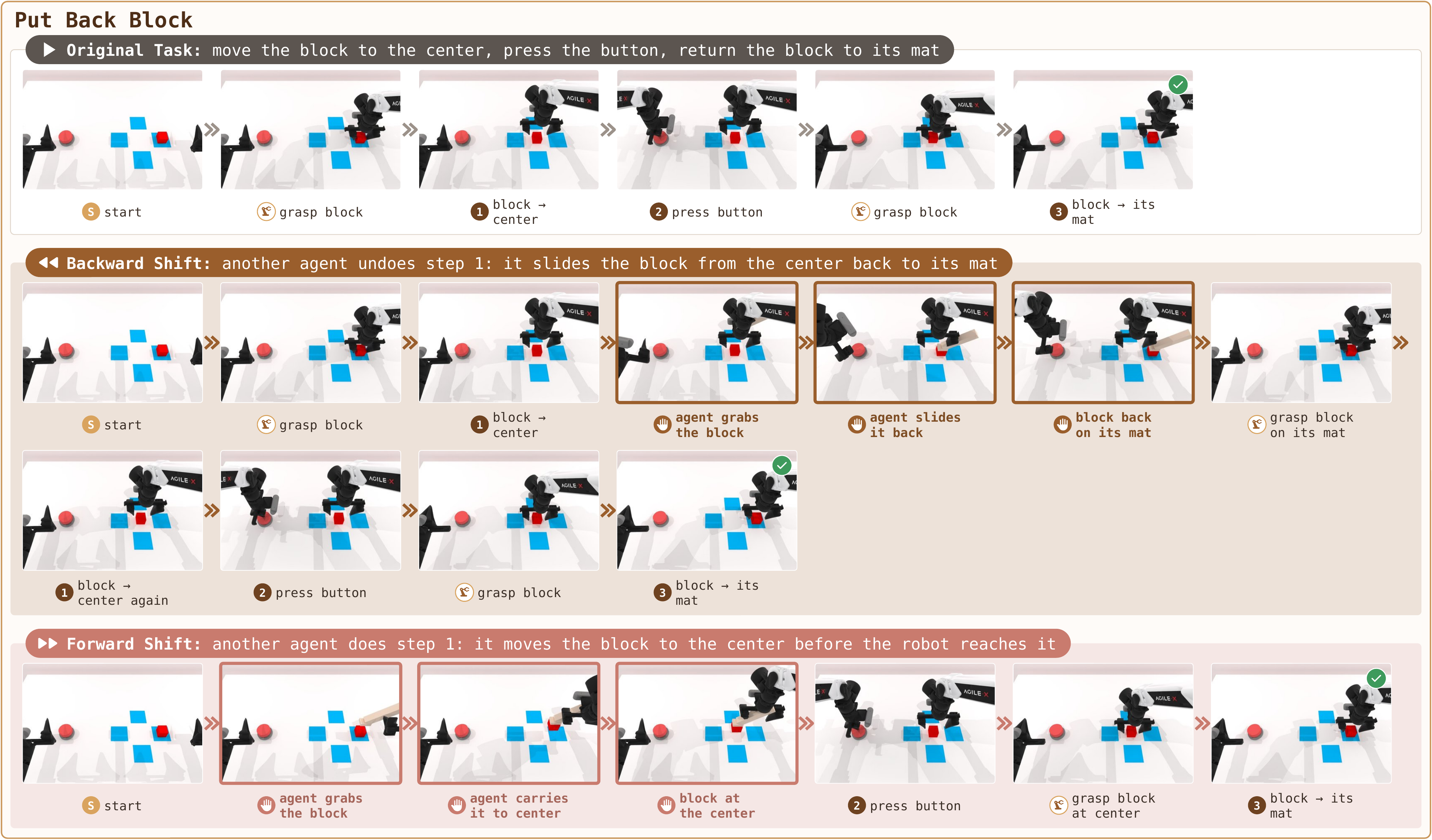}
\caption[RMBench-Delta: Put Back Block]{\textbf{Task protocol and task-progress shifts of \mbox{RMBench-$\Delta$}: Put Back Block.}\protect\shiftband{bwdband}{\protect\bwdtag: after the robot has released the block at the center and before the press, another agent slides it back to its own mat, and the robot should center it again\protect\thn{}press\protect\thn{}return it to its mat.}\protect\shiftband{fwdband}{\protect\fwdtag: another agent moves the block from its mat to the center before the robot reaches it, and the robot should leave it\protect\thn{}press\protect\thn{}return it to its mat.}The first row is a rollout without a shift (gray). Each further row is one shift, backward (brown) or forward (rose). Every row is a rollout of \methodname{} at its key frames, which are the start, every grasp, place and press, and three frames of the other agent's change, when its hand takes the object, moves it and sets it down. Frames are the policy's head camera, re-rendered at $1280{\times}960$ for display. Outlined frames show the other agent's hand. A number gives the step that a frame completes, S marks the start, an arm a grasp, a hand the other agent, and a check success.}
\label{fig:shift-rows}
\end{figure*}

\begin{figure*}[p]
\centering
\includegraphics[width=\textwidth]{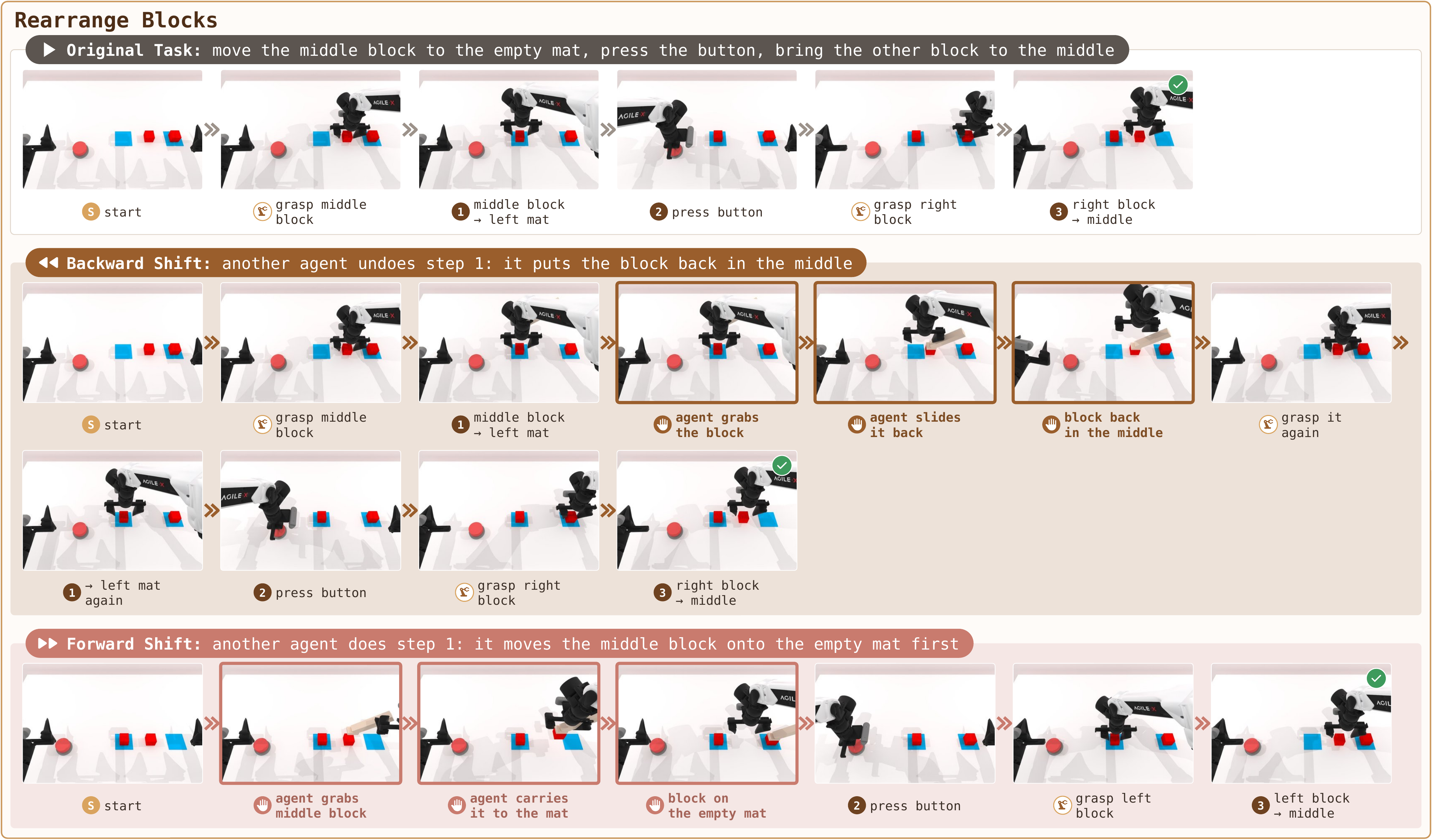}
\caption[RMBench-Delta: Rearrange Blocks]{\textbf{Task protocol and task-progress shifts of \mbox{RMBench-$\Delta$}: Rearrange Blocks.}\protect\shiftband{bwdband}{\protect\bwdtag: another agent puts the first block, just placed on the empty mat, back where it started, and the robot should move it to the empty mat again\protect\thn{}press\protect\thn{}bring the other block to the middle.}\protect\shiftband{fwdband}{\protect\fwdtag: another agent moves the first block onto the empty mat before the robot reaches it, and the robot should leave it\protect\thn{}press\protect\thn{}bring the other block to the middle.}Read as \cref{fig:shift-rows}.}
\label{fig:shift-rows-2}
\end{figure*}

\begin{figure*}[p]
\centering
\includegraphics[width=0.97\textwidth]{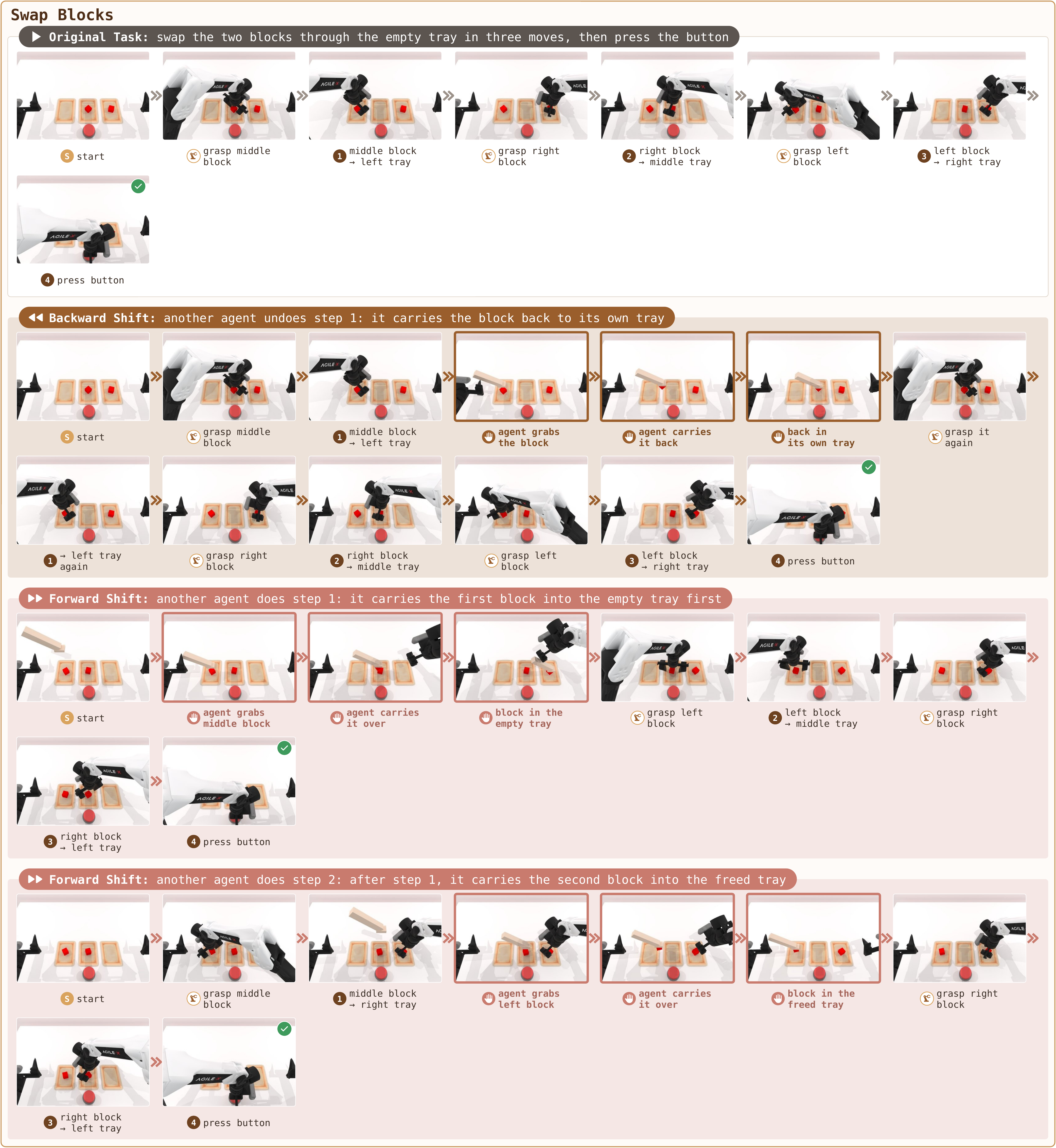}
\caption[RMBench-Delta: Swap Blocks]{\textbf{Task protocol and task-progress shifts of \mbox{RMBench-$\Delta$}: Swap Blocks.}\protect\shiftband{bwdband}{\protect\bwdtag: another agent carries the first block, just placed in the empty tray, back to its own tray, and the robot should place it in the empty tray again\protect\thn{}continue the swap.}\protect\shiftband{fwdband}{\protect\fwdtag~(a): another agent carries the first block into the empty tray before the robot grasps it.}\protect\shiftband{fwdband}{\protect\fwdtag~(b): after the robot's first move, another agent carries the second block into the freed tray. After either, the robot should skip that move\protect\thn{}continue with the next block\protect\thn{}press.}Read as \cref{fig:shift-rows}.}
\label{fig:shift-rows-3}
\end{figure*}
\begin{figure*}[p]
\centering
\includegraphics[width=\textwidth]{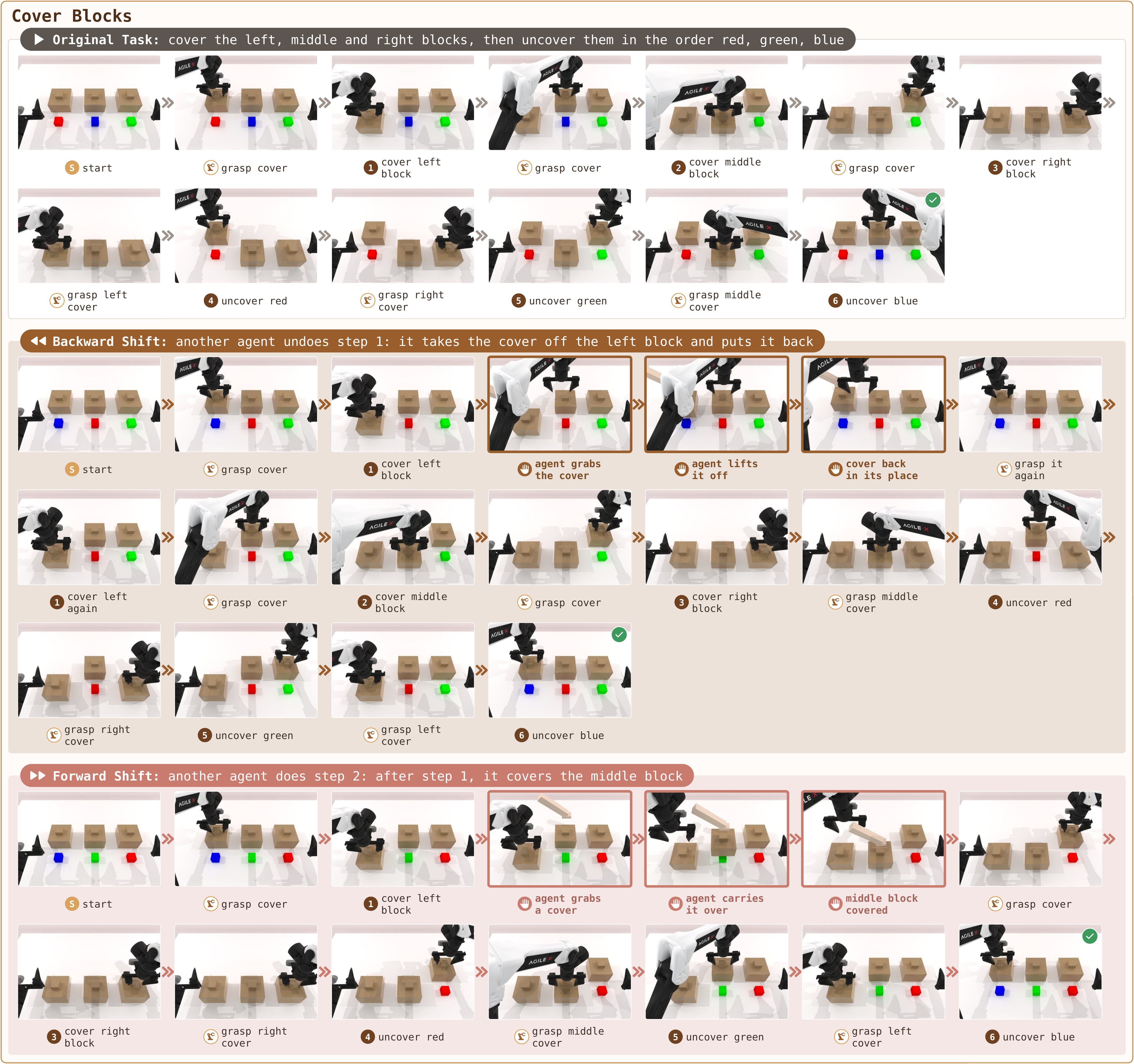}
\caption[RMBench-Delta: Cover Blocks]{\textbf{Task protocol and task-progress shifts of \mbox{RMBench-$\Delta$}: Cover Blocks.}\protect\shiftband{bwdband}{\protect\bwdtag: another agent takes the cover off the left block and returns it to the cover's place, and the robot should cover the left block again\protect\thn{}continue.}\protect\shiftband{fwdband}{\protect\fwdtag: another agent covers the middle block before the robot reaches it, and the robot should cover the right block\protect\thn{}uncover red, green, blue.}Read as \cref{fig:shift-rows}.}
\label{fig:shift-rows-4}
\end{figure*}

\begin{figure*}[p]
\centering
\includegraphics[width=\textwidth]{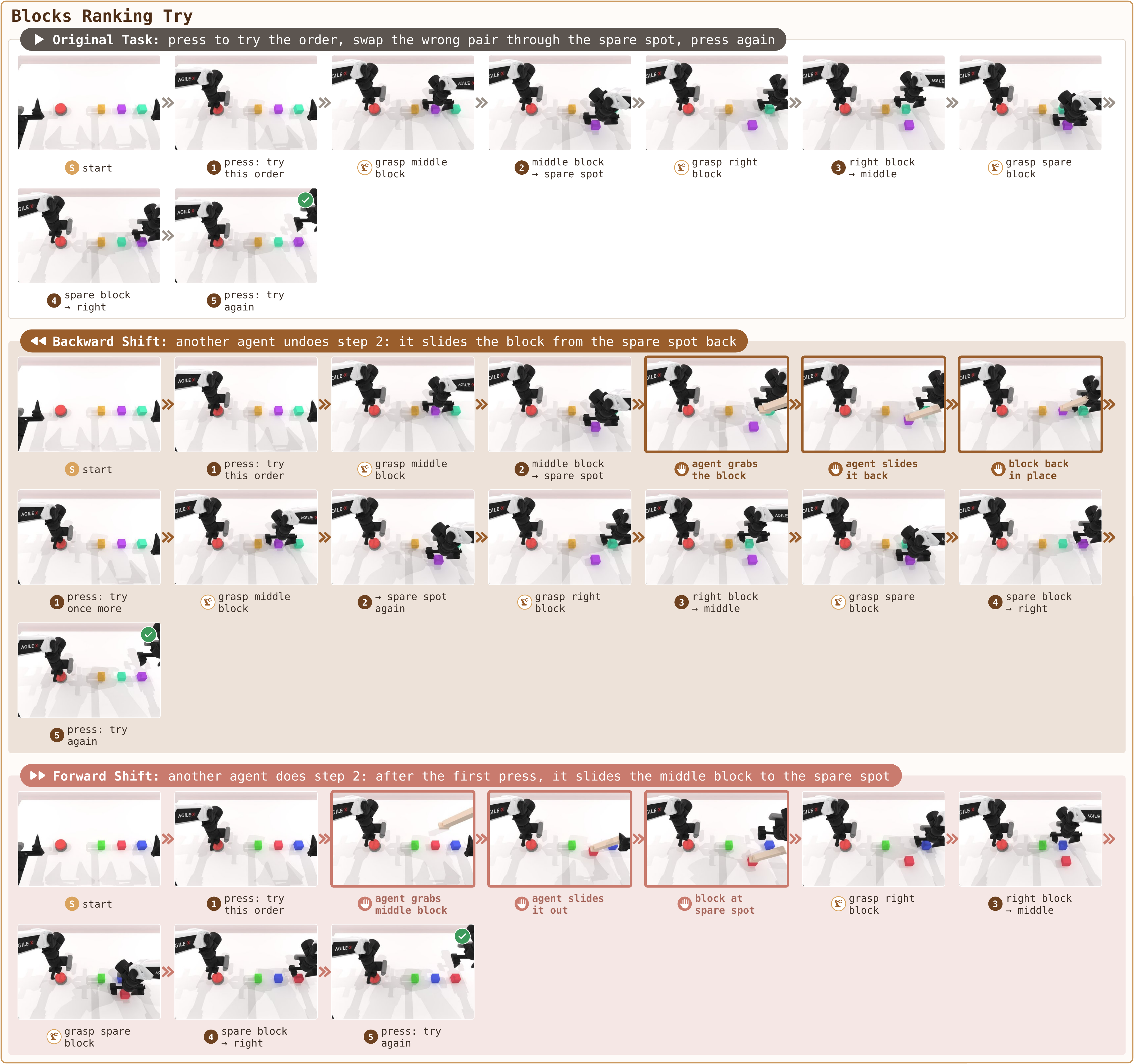}
\caption[RMBench-Delta: Blocks Ranking Try]{\textbf{Task protocol and task-progress shifts of \mbox{RMBench-$\Delta$}: Blocks Ranking Try.}\protect\shiftband{bwdband}{\protect\bwdtag: another agent slides the block that the robot has just set at the spare spot back to where it was, and the robot should move it to the spare spot again\protect\thn{}finish the swap\protect\thn{}press.}\protect\shiftband{fwdband}{\protect\fwdtag: after the first press, another agent slides the middle block to the spare spot, and the robot should leave it\protect\thn{}continue with the other block of the swap.}Read as \cref{fig:shift-rows}.}
\label{fig:shift-rows-5}
\end{figure*}

Each paragraph below describes a task without a shift. The task's figure shows a rollout without a shift and one per shift, and its caption
states each shift, what another agent does and what the robot should then do. Each episode gets one shift, staged as in \cref{sec:supp-protocol}.

\paragraph{Put Back Block.}
A block starts on one of four mats. The robot moves it to the center, presses a button, and returns it to the mat it came from, which only its
memory knows. The press leaves no trace in the scene, so only the memory records it. \Cref{fig:shift-rows} shows its shifts.

\paragraph{Rearrange Blocks.}
Two blocks and an empty mat are on the table. The robot moves the first block to the empty mat, presses a button, and brings the second block
to the middle. The two blocks look alike, so the robot must remember which one it moved. \Cref{fig:shift-rows-2} shows its shifts.

\paragraph{Swap Blocks.}
Two blocks sit in two trays next to a third, empty tray. The robot swaps them in three moves through the empty tray and then presses a button.
The two blocks look alike, so only the memory tells which moves are done. \Cref{fig:shift-rows-3} shows its shifts.

\paragraph{Cover Blocks.}
Three blocks of different colors are covered one after another and then uncovered in the order red, green, blue. Once they are covered, their
colors are only in the memory. \Cref{fig:shift-rows-4} shows its shifts.

\paragraph{Blocks Ranking Try.}
The robot presses a button to try the order of three blocks and, while the order is wrong, swaps two of them through a spare spot and presses
again. It has to remember what it has already tried. \Cref{fig:shift-rows-5} shows its shifts.

\paragraph{Task selection.}
Of the nine tasks of RMBench~\cite{chen2026rmbench}, we keep the five in which another agent can undo or complete a single step and the robot
then still needs its memory. Press Button moves no object, Observe and Pickup has a single step, Swap T moves both of its objects at once,
and Battery Try picks up both batteries at once and then only flips them in their slots.

\subsection{\mbox{RoboMemArena-$\Delta$}: tasks and shifts}
\label{sec:supp-rma}
\providecommand{\rmafigtasks}{}\renewcommand{\rmafigtasks}{5, 11, 12, 13, 14, 21, 22 and 24}
\begin{figure*}[p]
\centering
\includegraphics[width=0.91\textwidth]{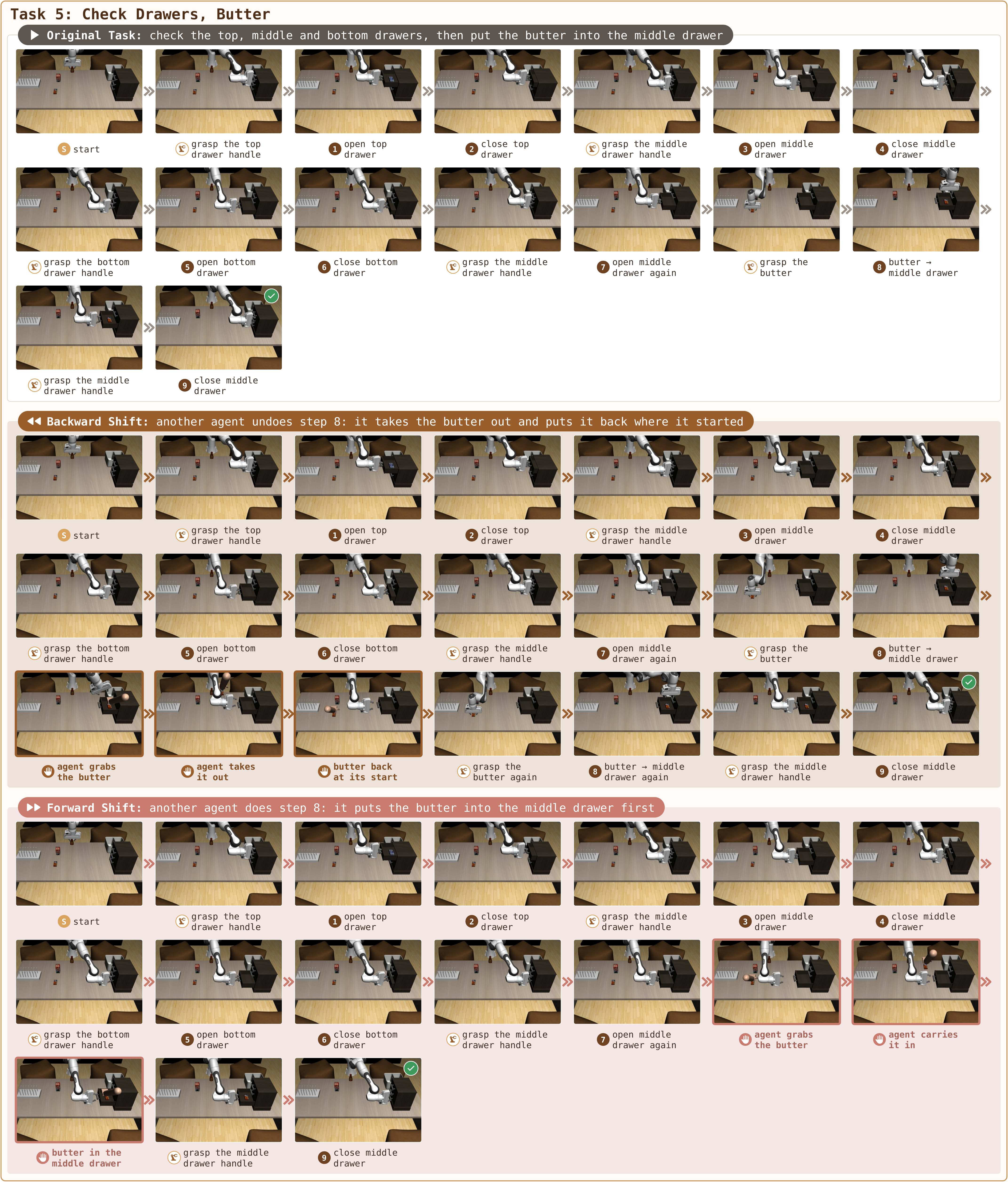}
\caption[RoboMemArena-Delta: Task 5, Check Drawers and Butter]{\textbf{Task protocol and task-progress shifts of \mbox{RoboMemArena-$\Delta$}: Task 5, Check Drawers and Butter.} Steps: open and close each drawer to check it (top\protect\thn middle\protect\thn bottom), then butter\protect\thn the empty drawer (here the middle one).\protect\shiftband{bwdband}{\protect\bwdtag: after the robot has placed the butter, another agent takes it back out of the drawer and sets it where it stood at the start, and the robot should place it again.}\protect\shiftband{fwdband}{\protect\fwdtag: another agent puts the butter into the drawer before the robot grasps it, and the robot should leave it and go on with the next step.} Read as \cref{fig:shift-rows}. Frames are the agent-view camera, re-rendered at $1440{\times}1080$ for display.}
\label{fig:rma-shift-rows}
\end{figure*}
\begin{figure*}[p]
\centering
\includegraphics[width=\textwidth]{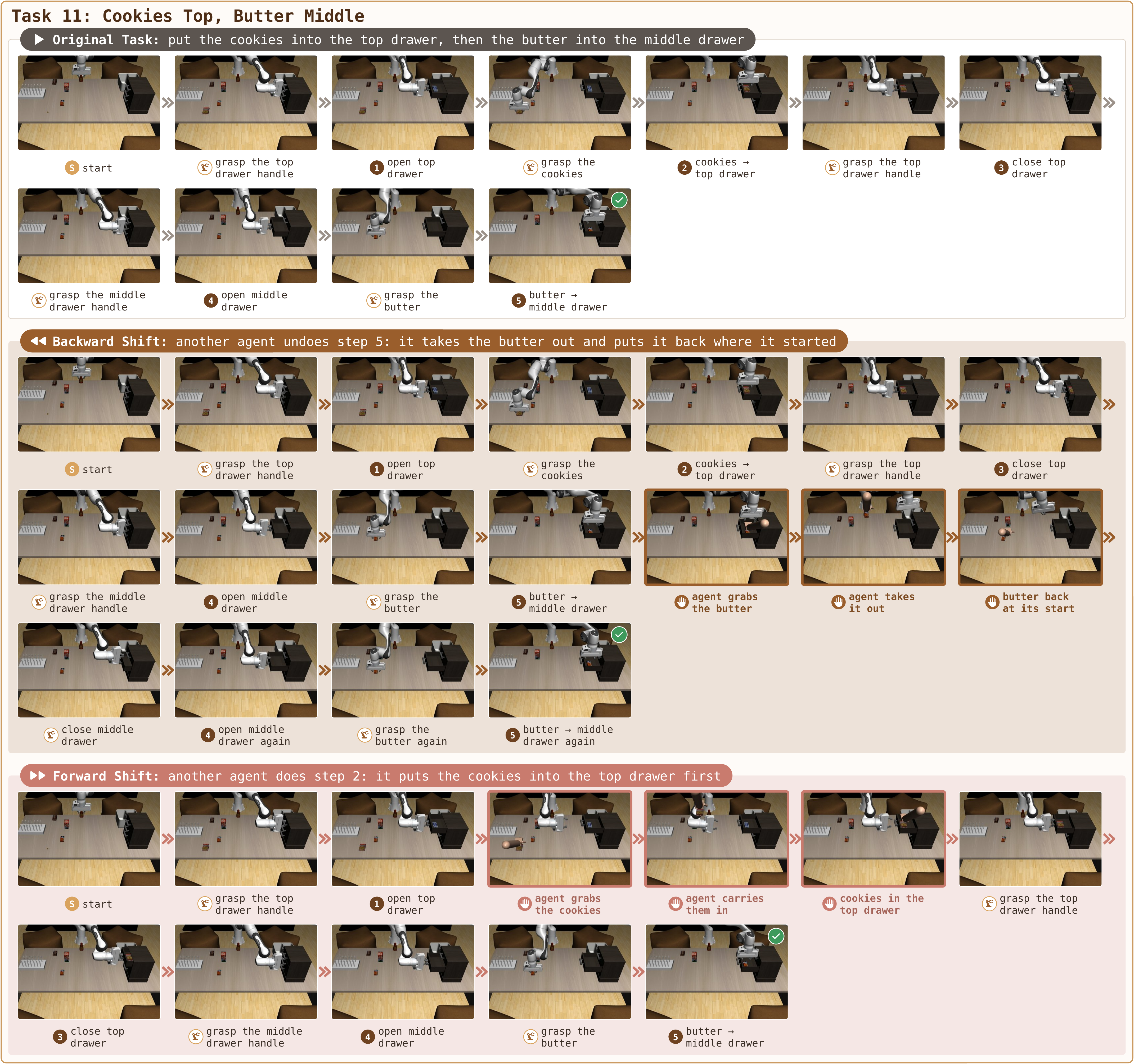}
\caption[RoboMemArena-Delta: Task 11, Cookies in the Top Drawer and Butter in the Middle Drawer]{\textbf{Task protocol and task-progress shifts of \mbox{RoboMemArena-$\Delta$}: Task 11, Cookies in the Top Drawer and Butter in the Middle Drawer.} Steps: cookies\protect\thn top drawer, then butter\protect\thn middle drawer.\protect\shiftband{bwdband}{\protect\bwdtag: after the robot has placed the butter, another agent takes it back out of the drawer and sets it where it stood at the start, and the robot should place it again.}\protect\shiftband{fwdband}{\protect\fwdtag: another agent puts the cookies into the drawer before the robot grasps them, and the robot should leave them and go on with the next step.} Read as \cref{fig:shift-rows}.}
\label{fig:rma-shift-rows-t11}
\end{figure*}

\begin{figure*}[p]
\centering
\includegraphics[width=\textwidth]{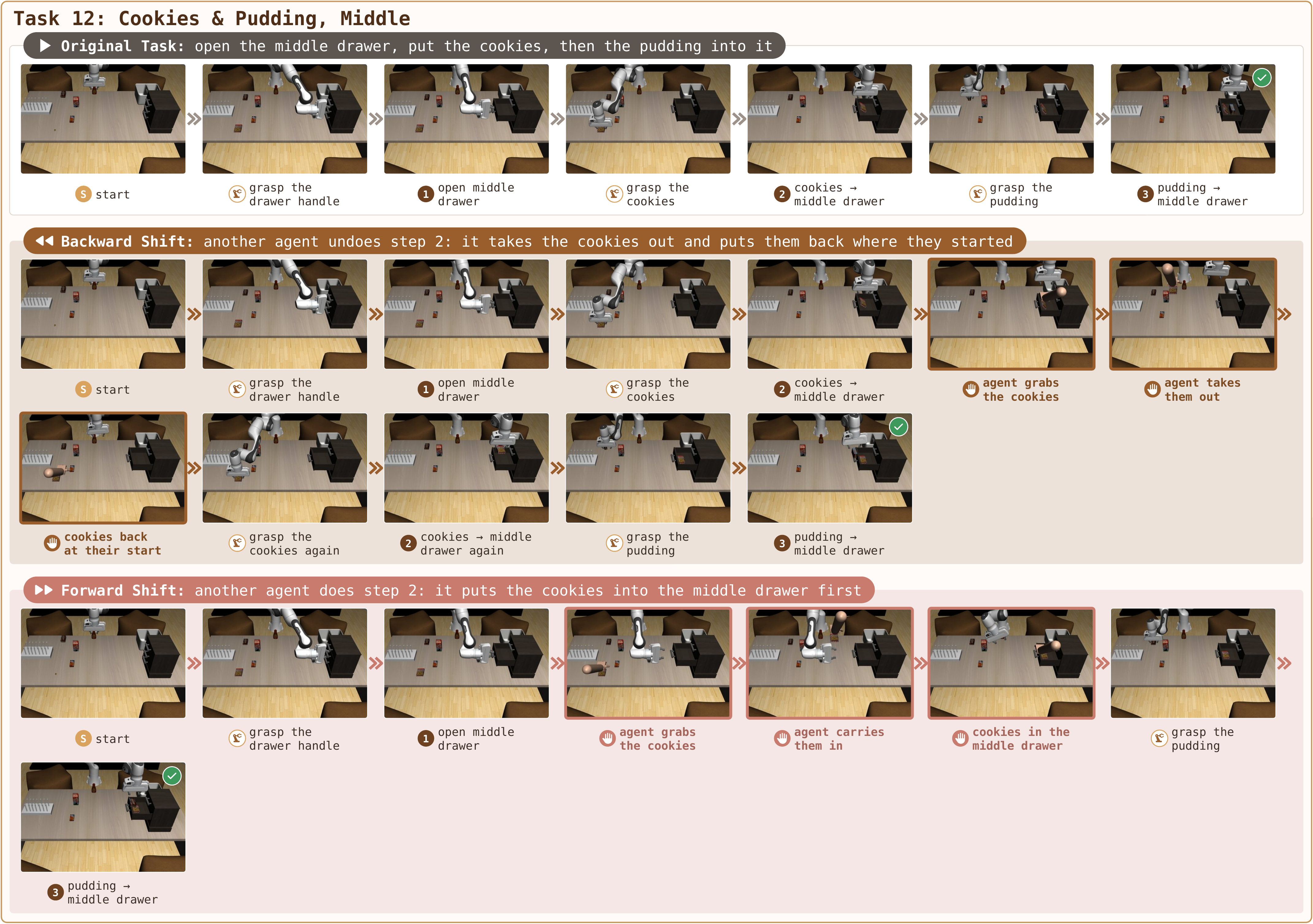}
\caption[RoboMemArena-Delta: Task 12, Cookies and Pudding]{\textbf{Task protocol and task-progress shifts of \mbox{RoboMemArena-$\Delta$}: Task 12, Cookies and Pudding.} Steps: cookies\protect\thn middle drawer, then chocolate pudding\protect\thn the same drawer.\protect\shiftband{bwdband}{\protect\bwdtag: after the robot has placed the cookies, another agent takes them back out of the drawer and sets them where they stood at the start, and the robot should place them again.}\protect\shiftband{fwdband}{\protect\fwdtag: another agent puts the cookies into the drawer before the robot grasps them, and the robot should leave them and go on with the next step.} Read as \cref{fig:shift-rows}.}
\label{fig:rma-shift-rows-t12}
\end{figure*}

\begin{figure*}[p]
\centering
\includegraphics[width=\textwidth]{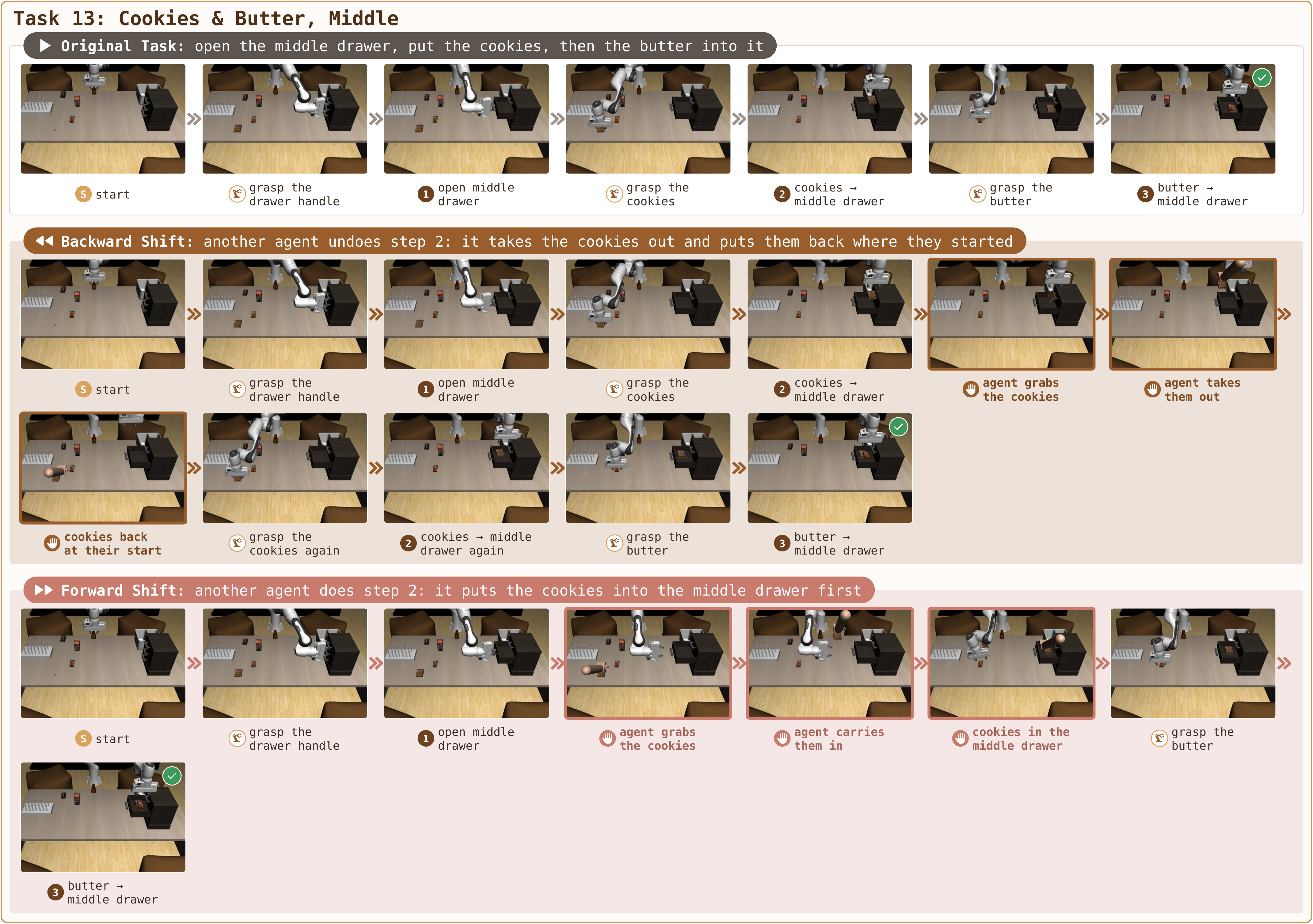}
\caption[RoboMemArena-Delta: Task 13, Cookies and Butter]{\textbf{Task protocol and task-progress shifts of \mbox{RoboMemArena-$\Delta$}: Task 13, Cookies and Butter.} Steps: cookies\protect\thn middle drawer, then butter\protect\thn the same drawer.\protect\shiftband{bwdband}{\protect\bwdtag: after the robot has placed the cookies, another agent takes them back out of the drawer and sets them where they stood at the start, and the robot should place them again.}\protect\shiftband{fwdband}{\protect\fwdtag: another agent puts the cookies into the drawer before the robot grasps them, and the robot should leave them and go on with the next step.} Read as \cref{fig:shift-rows}.}
\label{fig:rma-shift-rows-t13}
\end{figure*}

\begin{figure*}[p]
\centering
\includegraphics[width=\textwidth]{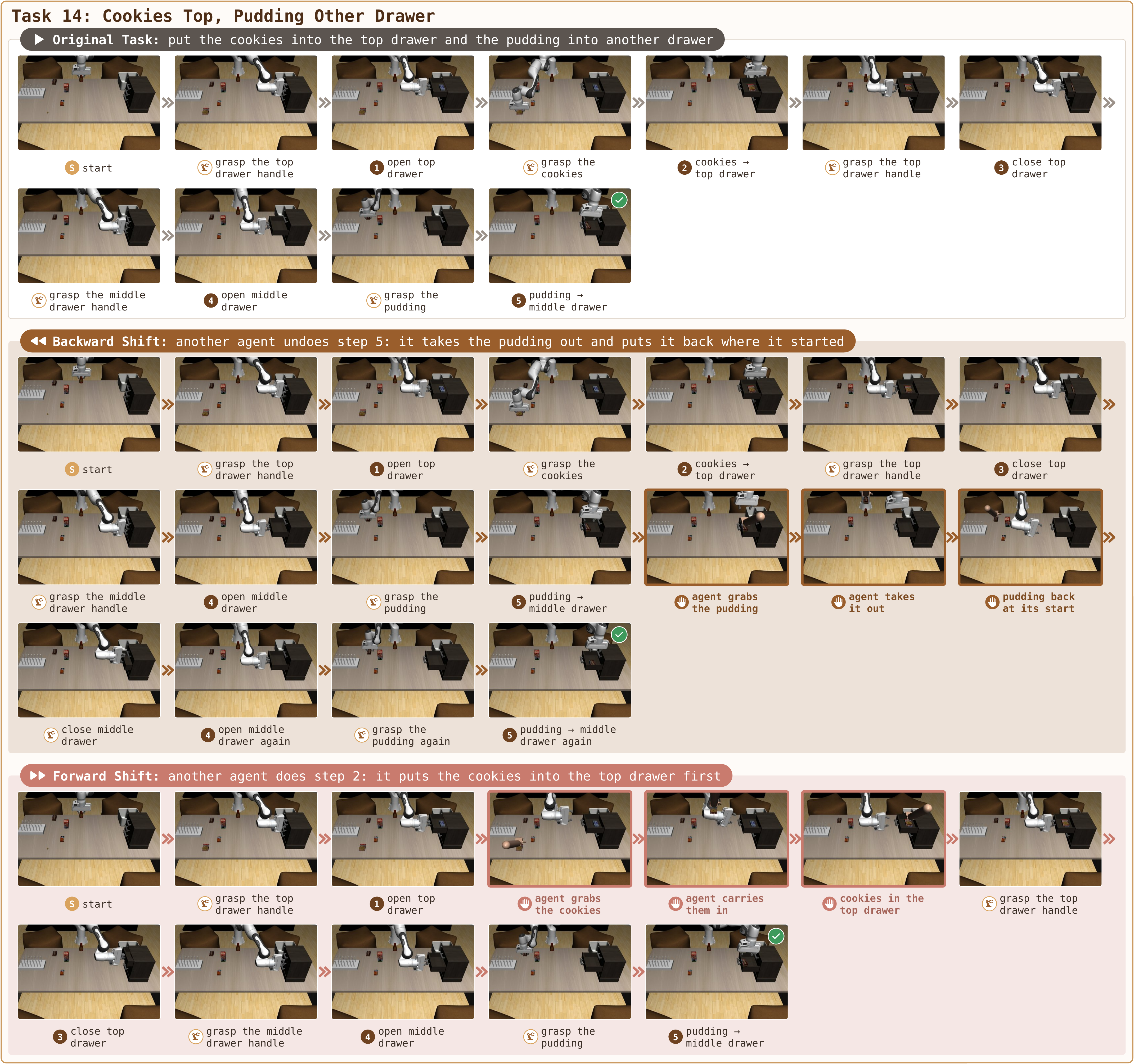}
\caption[RoboMemArena-Delta: Task 14, Cookies in the Top Drawer and Pudding in Another Drawer]{\textbf{Task protocol and task-progress shifts of \mbox{RoboMemArena-$\Delta$}: Task 14, Cookies in the Top Drawer and Pudding in Another Drawer.} Steps: cookies\protect\thn top drawer, and chocolate pudding\protect\thn another drawer.\protect\shiftband{bwdband}{\protect\bwdtag: after the robot has placed the chocolate pudding, another agent takes it back out of the drawer and sets it where it stood at the start, and the robot should place it again.}\protect\shiftband{fwdband}{\protect\fwdtag: another agent puts the cookies into the drawer before the robot grasps them, and the robot should leave them and go on with the next step.} Read as \cref{fig:shift-rows}.}
\label{fig:rma-shift-rows-t14}
\end{figure*}

\begin{figure*}[p]
\centering
\includegraphics[width=\textwidth]{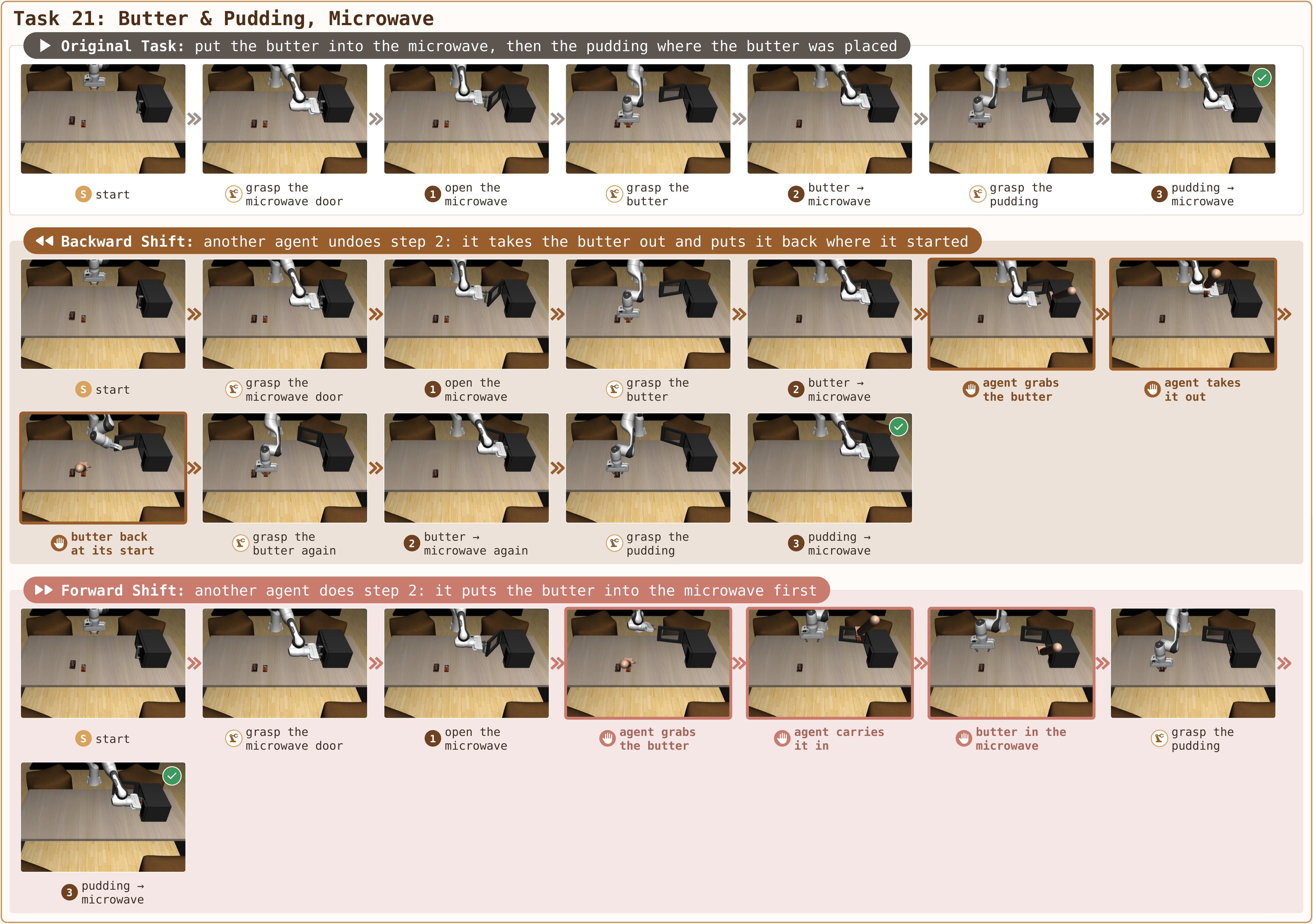}
\caption[RoboMemArena-Delta: Task 21, Butter and Pudding in the Microwave]{\textbf{Task protocol and task-progress shifts of \mbox{RoboMemArena-$\Delta$}: Task 21, Butter and Pudding in the Microwave.} Steps: butter\protect\thn microwave, then chocolate pudding\protect\thn where the butter was placed.\protect\shiftband{bwdband}{\protect\bwdtag: after the robot has placed the butter, another agent takes it back out of the microwave and sets it where it stood at the start, and the robot should place it again.}\protect\shiftband{fwdband}{\protect\fwdtag: another agent puts the butter into the microwave before the robot grasps it, and the robot should leave it and go on with the next step.} Read as \cref{fig:shift-rows}.}
\label{fig:rma-shift-rows-t21}
\end{figure*}

\begin{figure*}[p]
\centering
\includegraphics[width=\textwidth]{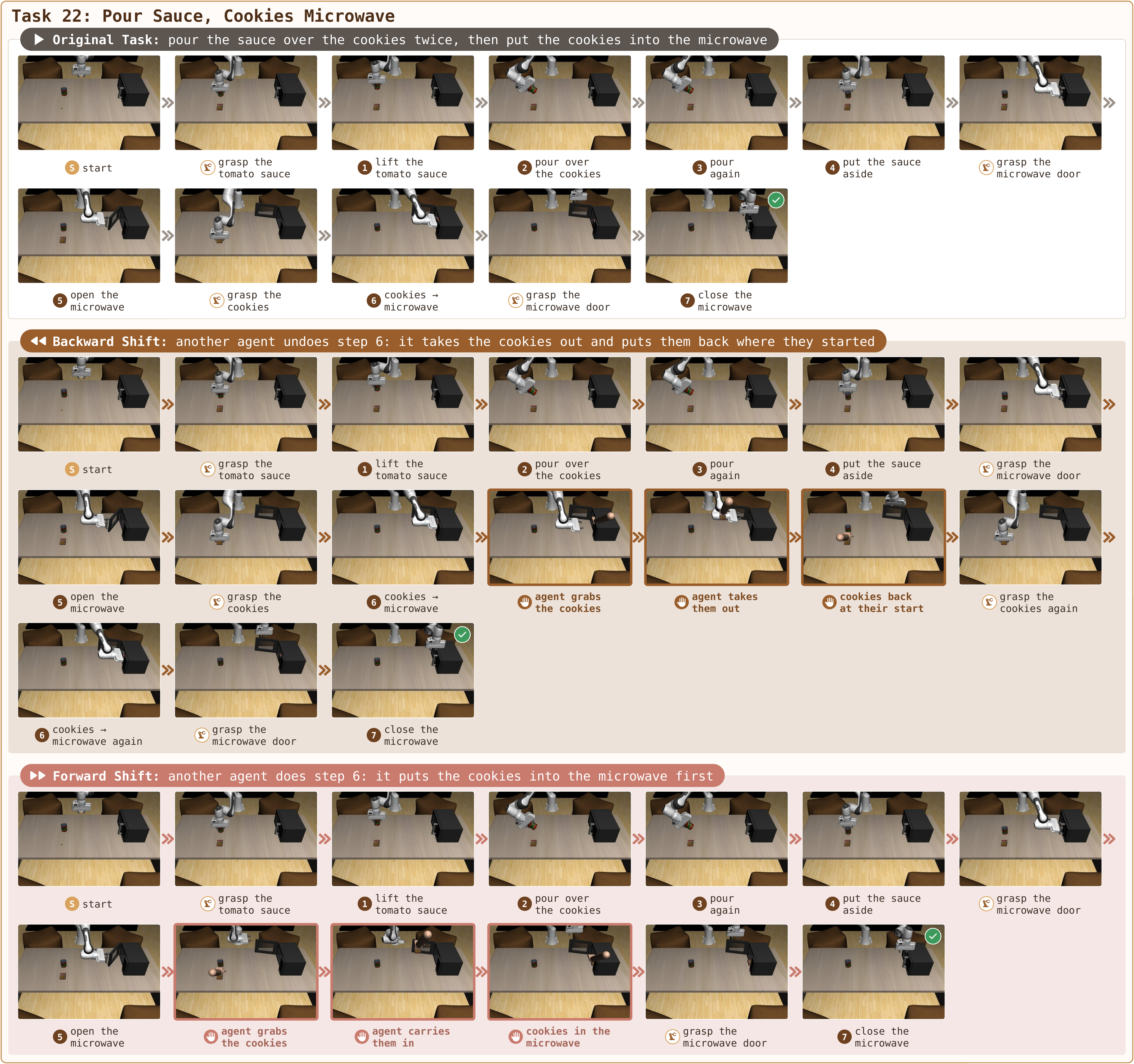}
\caption[RoboMemArena-Delta: Task 22, Pour Sauce and Cookies]{\textbf{Task protocol and task-progress shifts of \mbox{RoboMemArena-$\Delta$}: Task 22, Pour Sauce and Cookies.} Steps: pour tomato sauce over the cookies twice, then cookies\protect\thn microwave.\protect\shiftband{bwdband}{\protect\bwdtag: after the robot has placed the cookies, another agent takes them back out of the microwave and sets them where they stood at the start, and the robot should place them again.}\protect\shiftband{fwdband}{\protect\fwdtag: another agent puts the cookies into the microwave before the robot grasps them, and the robot should leave them and go on with the next step.} Read as \cref{fig:shift-rows}.}
\label{fig:rma-shift-rows-t22}
\end{figure*}

\begin{figure*}[p]
\centering
\includegraphics[width=\textwidth]{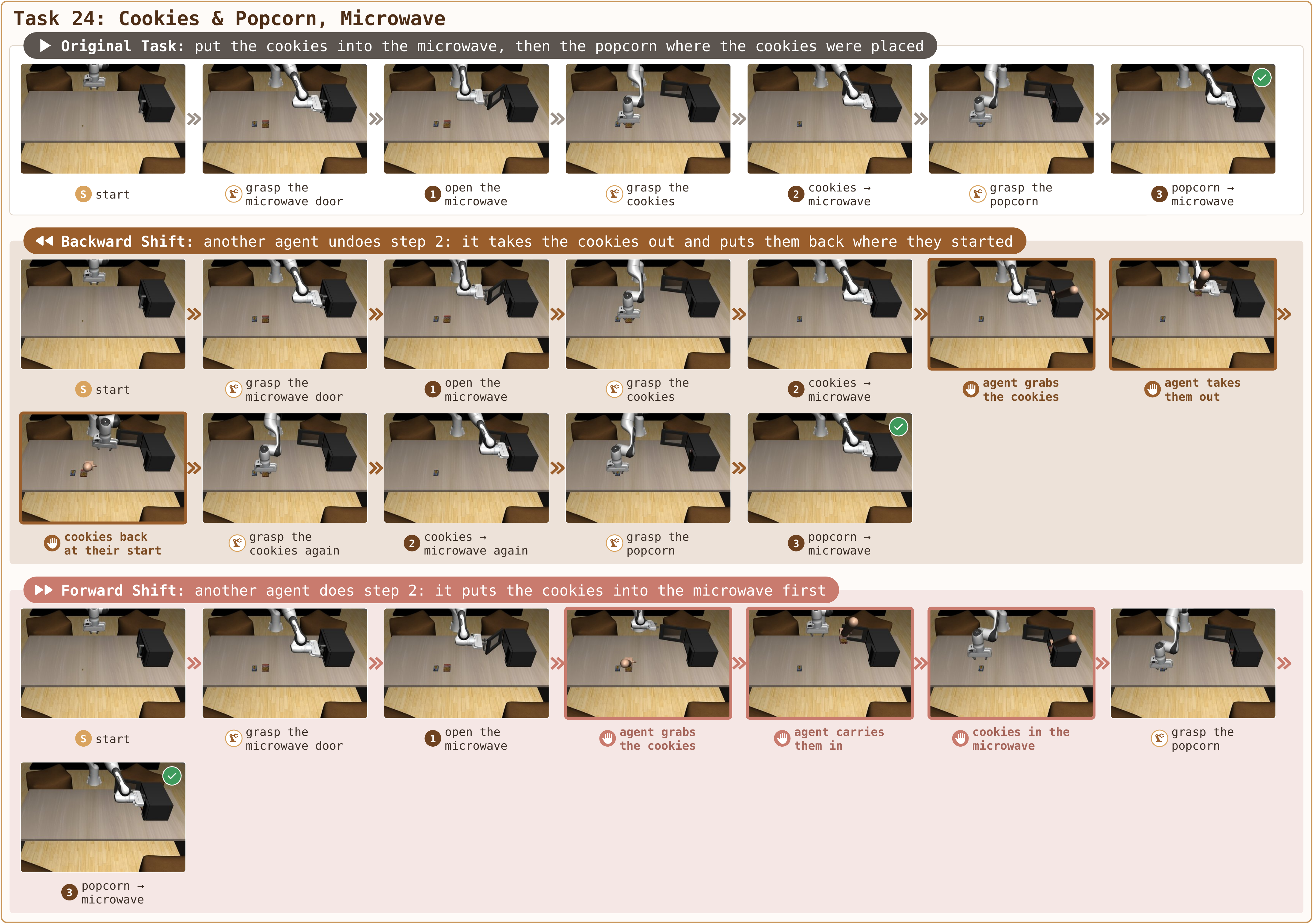}
\caption[RoboMemArena-Delta: Task 24, Cookies and Popcorn in the Microwave]{\textbf{Task protocol and task-progress shifts of \mbox{RoboMemArena-$\Delta$}: Task 24, Cookies and Popcorn in the Microwave.} Steps: cookies\protect\thn microwave, then popcorn\protect\thn where the cookies were placed.\protect\shiftband{bwdband}{\protect\bwdtag: after the robot has placed the cookies, another agent takes them back out of the microwave and sets them where they stood at the start, and the robot should place them again.}\protect\shiftband{fwdband}{\protect\fwdtag: another agent puts the cookies into the microwave before the robot grasps them, and the robot should leave them and go on with the next step.} Read as \cref{fig:shift-rows}.}
\label{fig:rma-shift-rows-t24}\label{fig:rma-shift-rows-last}
\end{figure*}

Following the same rule, eight tasks are chosen from RoboMemArena~\cite{lei2026robomemarena}. Each paragraph below describes a task without a shift. In every shift another agent moves one item, and the task's figure shows a rollout
without a shift and one per shift, with the shifts stated in its caption.

\paragraph{Task 5.}
The robot opens and closes the top, middle and bottom drawers in turn to check them, then puts the butter into the drawer that was empty. The
closed drawers look alike, so only the memory tells which one was empty. \Cref{fig:rma-shift-rows} shows its shifts.

\paragraph{Task 11.}
The robot puts the cookies into the top drawer, then the butter into the middle drawer. Once the top drawer is closed, only the memory tells
that the cookies are done and the butter comes next. \Cref{fig:rma-shift-rows-t11} shows its shifts.

\paragraph{Task 12.}
The robot puts the cookies into the middle drawer, then the chocolate pudding into the same drawer. The cookies are out of view inside the
drawer, so only the memory tells that the pudding comes next and where it goes. \Cref{fig:rma-shift-rows-t12} shows its shifts.

\paragraph{Task 13.}
The robot puts the cookies into the middle drawer, then the butter into the same drawer. The cookies are out of view inside the drawer, so
only the memory tells that the butter comes next and where it goes. \Cref{fig:rma-shift-rows-t13} shows its shifts.

\paragraph{Task 14.}
The robot puts the cookies into the top drawer and the chocolate pudding into another drawer. Once the top drawer is closed, only the memory
tells which drawer holds the cookies. \Cref{fig:rma-shift-rows-t14} shows its shifts.

\paragraph{Task 21.}
The robot puts the butter into the microwave, then the chocolate pudding where the butter was placed. The butter is out of view inside the
microwave, so only the memory tells where the pudding goes. \Cref{fig:rma-shift-rows-t21} shows its shifts.

\paragraph{Task 22.}
The robot pours tomato sauce over the cookies twice, then puts the cookies into the microwave. A pour leaves no trace in the scene, so only
the memory counts the pours. \Cref{fig:rma-shift-rows-t22} shows its shifts.

\paragraph{Task 24.}
The robot puts the cookies into the microwave, then the popcorn where the cookies were placed. The cookies are out of view inside the
microwave, so only the memory tells where the popcorn goes. \Cref{fig:rma-shift-rows-t24} shows its shifts.

\subsection{Training settings of \methodname{}}
\label{sec:supp-baselines}
\label{sec:supp-train}

\Cref{tab:supp-train} gives the training settings of both stages of \methodname{}.

\begin{table}[t]
\centering\footnotesize
\definecolor{setgrp}{HTML}{ECECEC}
\renewcommand{\arraystretch}{1.3}\setlength{\aboverulesep}{0pt}\setlength{\belowrulesep}{0pt}
\setlength{\tabcolsep}{2.5pt}
\begin{tabular}{ll}
\toprule
\rowcolor{setgrp}\multicolumn{2}{l}{\textbf{Stage 1}\quad latent world action model}\\
Data & standard demonstrations, time-warped\\
Batch & 2 whole demonstrations\\
Updates & 30{,}000\\
Optimizer & AdamW~\cite{loshchilov2017decoupled}, $(\beta_1,\beta_2)=(0.9,0.95)$\\
Learning rate & $10^{-4}$, 500 warm-up updates, cosine decay\\
Weight decay & $10^{-6}$\\
Weight average & EMA, power 0.75, decay $\le0.9999$, then frozen\\
Loss weights & $\beta=10^{-5}$, $\gamma=0.1$ in~\cref{eq:model}\\
Augmentation & random $216{\times}216$ crops, dropout 0.1\\
Proprioception & 1 input in 8 displaced by a state-head residual\\
Diffusion & 100 noise levels\\
\rowcolor{setgrp}\multicolumn{2}{l}{\textbf{Stage 2}\quad memory localizer}\\
Examples & 16 synthetic shifts per demonstration\\
Batch & 32 examples\\
Updates & 6{,}000\\
Optimizer & AdamW, $(\beta_1,\beta_2)=(0.95,0.999)$\\
Gradient norm & clipped at 10\\
Learning rate & $5\cdot10^{-4}$, constant; weight decay $10^{-6}$\\
\bottomrule
\end{tabular}
\caption{\textbf{Training settings} of \methodname{}. Stage 1 trains the latent world action model, and stage 2 trains the memory localizer on
the frozen model.}
\label{tab:supp-train}
\end{table}

\FloatBarrier
\twocolumn[{\centering
\includegraphics[width=\textwidth]{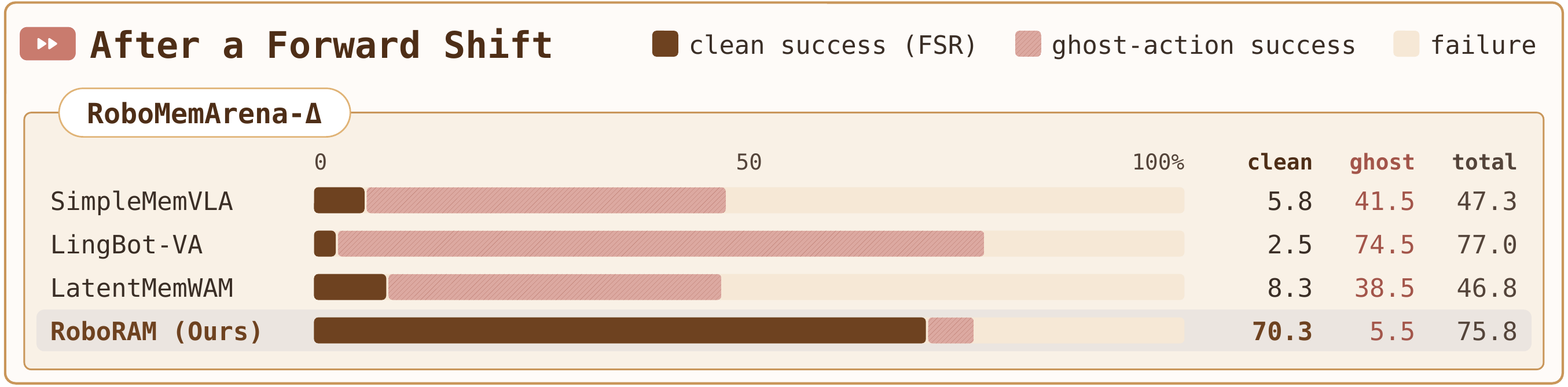}
\captionof{figure}{\textbf{Outcomes after a forward shift on \mbox{RoboMemArena-$\Delta$}.} Each bar splits a policy's forward-shift rollouts into clean
successes (FSR), successes with a ghost action and failures, in percent, averaged over the eight tasks as in~\cref{tab:main-rma}. Task success
(total) is the clean and ghost segments together. \cref{fig:ghost} shows the same split on \mbox{RMBench-$\Delta$}.}
\label{fig:forward-ghost}
\vspace{1.2em}}]

\subsection{Results without a shift}
\label{sec:supp-noshift}
\Cref{tab:rmbench_wam} adds \methodname{} and LatentMemWAM to the published results of world action models on the five tasks of
\mbox{RMBench-$\Delta$} with the robot alone. LatentMemWAM is the world action model of \methodname{} with the same weights, run without the
memory localizer, so the two show what revising the memory costs when nothing shifts. Revising the memory thus keeps the policy's success when
nothing shifts, and over 100 trials per task \methodname{} succeeds in 96.0\% of the episodes, about as often as the strongest published methods.

\begin{table*}[t]
  \centering\footnotesize
  \renewcommand{\theadfont}{\bfseries}\renewcommand{\theadgape}{}
  \renewcommand{\arraystretch}{1.18}
  \setlength{\tabcolsep}{5pt}
  \begin{tabular*}{\textwidth}{@{}l !{\extracolsep{\fill}} ccc cc ccc@{}}
    \toprule
    & \multicolumn{3}{c}{\textbf{$\boldsymbol{M(1)}$ Tasks}} & \multicolumn{2}{c}{\textbf{$\boldsymbol{M(n)}$ Tasks}} & \multicolumn{3}{c@{}}{\textbf{Average}} \\
    \cmidrule(lr){2-4} \cmidrule(lr){5-6} \cmidrule(l){7-9}
    \textbf{Method}
      & \thead{Rearrange\\Blocks} & \thead{Put Back\\Block} & \thead{Swap\\Blocks}
      & \thead{Blocks\\Ranking Try} & \thead{Cover\\Blocks}
      & $\boldsymbol{M(1)}$ & $\boldsymbol{M(n)}$ & \textbf{Total} \\
    \midrule
    Fast-WAM~\cite{yuan2026fast} & 0 & 0 & 0 & 26 & 0 & 0.0 & 13.0 & 5.2 \\
    HiMem-WAM~\cite{sun2026himem} & 33 & 32 & 38 & 24 & 19 & 34.3 & 21.5 & 29.2 \\
    RM-WAM~\cite{miao2026rm} & 65 & 70 & 62 & 75 & 64 & 65.7 & 69.5 & 67.2 \\
    DiM-WAM~\cite{wang2026dim} & 99 & 98 & 96 & 87 & 56 & 97.7 & 71.5 & 87.2 \\
    MaP-WAM~\cite{zhao2026memory} & 66 & \textbf{100} & 97 & 94 & \textbf{100} & 87.7 & \textbf{97.0} & 91.4 \\
    LingBot-VA~\cite{li2026causal} & \textbf{100} & \textbf{100} & \textbf{99} & \textbf{100} & 79 & \textbf{99.7} & 89.5 & 95.6 \\
    \midrule
    LatentMemWAM (Ours) & \textbf{100} & \textbf{100} & 90 & 90 & 82 & 96.7 & 86.0 & 92.4 \\
    \textbf{\methodname{} (Ours)} & \textbf{100} & \textbf{100} & 97 & 94 & 89 & 99.0 & 91.5 & \textbf{96.0} \\
    \bottomrule
  \end{tabular*}
  \caption{\textbf{Success rates (\%) of world action models on five RMBench tasks without any task-progress shifts}, as published, and for our two models with
  RMBench's protocol.}
  \label{tab:rmbench_wam}
\end{table*}

\subsection{Failure mode analysis}
\label{sec:supp-failures}
\label{sec:supp-clock}
\label{sec:supp-failtask}

\paragraph{Task success after a forward shift.}
After a forward shift, another agent has completed one of the robot's steps, so the scene already shows the result of that step. The success
checks of both benchmarks do not record who completed the step or how often. A policy that ignores the shift can
repeat the step, grasping at the place the object has left or lifting the object and setting it down where it already lies, and still pass the
check. Task success after a forward shift can therefore be high for a policy that does not react, and FSR, which does not count a success with a ghost
action, separates the two. In Put Back Block, for example, after another agent moves the block to the center, a policy that ignores the shift
closes its gripper on the empty mat, presses the button and returns the block from the center to its mat. The final scene is the goal, so the
episode counts as a success.

\Cref{fig:forward-ghost} splits each policy's forward-shift rollouts on \mbox{RoboMemArena-$\Delta$} into clean successes, successes with a
ghost action and failures, as \cref{fig:ghost} does on \mbox{RMBench-$\Delta$}. Ghost actions make up 74.5 of LingBot-VA's 77.0\% task
success, 41.5 of SimpleMemVLA's 47.3\% and 38.5 of LatentMemWAM's 46.8\%, but only 5.5 of \methodname{}'s 75.8\%, so nearly all of its
successes after a forward shift are clean. On \mbox{RMBench-$\Delta$}, its lower task success after a forward shift comes from one task, Blocks
Ranking Try (48\%). On the other four tasks it reaches 94.0\%, and 92.0\% of its rollouts there succeed cleanly, against 5.5\% for SimpleMemVLA.

\begin{figure*}[tp]
\centering
\includegraphics[width=\textwidth]{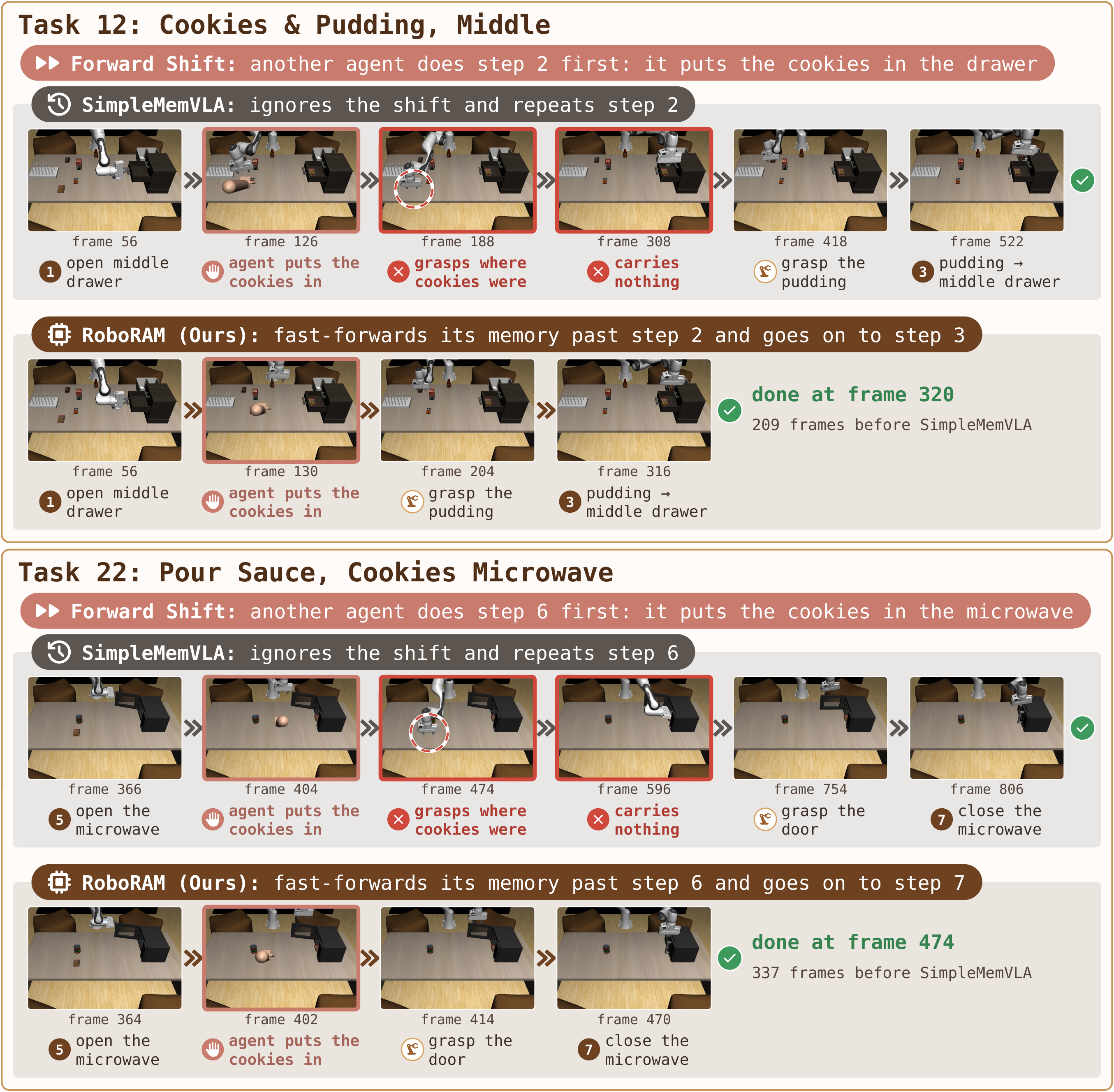}
\caption{\textbf{Repeating a step after a forward shift (SimpleMemVLA) and skipping it (\methodname{}).} Two \mbox{RoboMemArena-$\Delta$} tasks,
the same seed for both policies. Rose frames: another agent does the robot's next step. Red frames: SimpleMemVLA closes its gripper where the
cookies were (dashed ring), a ghost action, and carries nothing to the container. Frame numbers count the robot's steps. Frames are re-rendered at $1440{\times}1080$ for display.}
\label{fig:ghost-vs-reactive}
\end{figure*}

\paragraph{What a ghost action looks like.}
\Cref{fig:ghost-vs-reactive} shows a forward shift in two \mbox{RoboMemArena-$\Delta$} tasks, each rolled out on one seed with SimpleMemVLA and with
\methodname{}. After another agent puts the cookies into the drawer or the microwave, SimpleMemVLA goes back to the place the cookies left, closes
its gripper on nothing and carries nothing to the container before it goes on. Both rollouts end in the goal scene, but SimpleMemVLA's contains a
ghost action and does not count for FSR. \methodname{} fast-forwards its memory past the step that the other agent did and goes on to the next one,
so it finishes 209 and 337 steps earlier.

\begin{figure*}[tp]
\centering
\includegraphics[width=\textwidth]{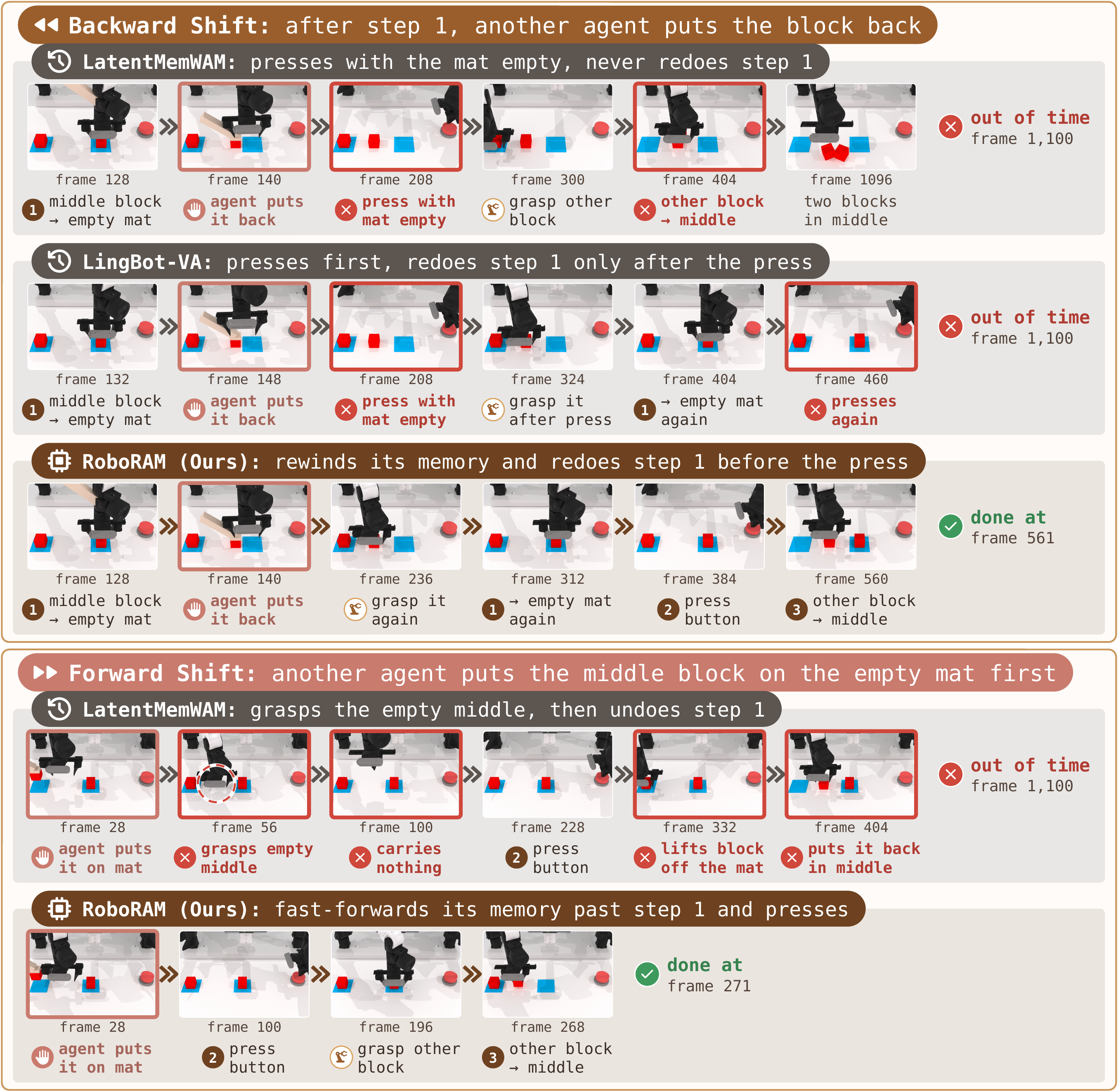}
\caption{\textbf{Failure modes on Rearrange Blocks.} The robot moves the middle block onto the empty mat (step~1), presses the button (step~2) and
brings the other block into the middle (step~3). Top: one scene after a backward shift. Bottom: one scene after a forward shift. Rose frames: another agent shifts the task progress. Red frames: an action that follows the memory against the scene, such as pressing with the mat empty, closing the gripper where the middle block was (dashed ring) or taking the placed block off the mat again. Frame numbers count the robot's steps. The rows are filmed from a second camera for better visualization.}
\label{fig:failure-task}
\end{figure*}
\paragraph{Exemplar failure modes.}
\Cref{fig:failure-task} shows rollouts on Rearrange Blocks, where the robot moves the first of two identical blocks onto an empty mat, presses a
button and brings the second block to the middle. Only the memory tells the two blocks apart. After a backward shift, which returns the first block
to its start before the press, LatentMemWAM, LingBot-VA and Long-Context DP act on their outdated memory in all 50 rollouts of~\cref{tab:main}. They press with the block still at its start and run out of time. \methodname{} rewinds its memory, places the block again before it
presses, and succeeds in all 50. After a forward shift, which places the first block on the mat before the robot reaches it, the three baselines
close the gripper on nothing at the block's start place in all 50 rollouts, and 13, 23 and 23 of them still pass the task's check with this ghost
action. \methodname{} fast-forwards its memory and succeeds without a ghost action in 48 of 50 rollouts.
\clearpage
\makeatletter\global\@colroom\textheight\global\vsize\textheight\makeatother
\section{Real-robot experiments}
\label{sec:supp-real}
\label{sec:supp_real}
\subsection{Setup}
\label{sec:supp_real_setup}
\begin{figure}[ht]
\centering
\includegraphics[width=\columnwidth]{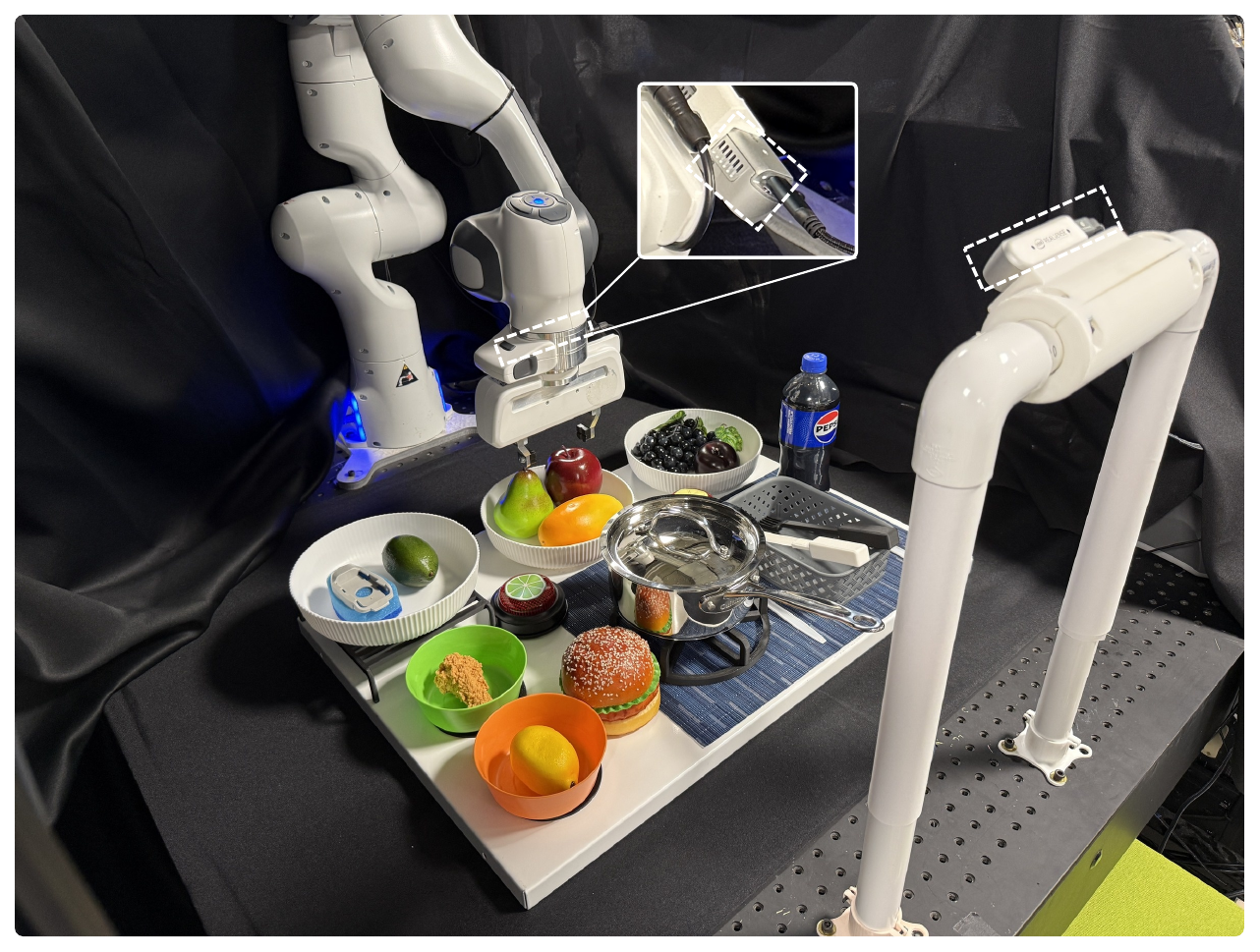}
\caption{\textbf{Real-robot setup.} The white dashed boxes mark the third-view camera, on top of the arch, and the wrist camera, which the wrist
flange hides in this view. The inset shows the wrist camera from the robot's other side.}
\label{fig:real-setup}
\end{figure}

A Franka Emika Panda arm with a parallel gripper is observed by two Intel RealSense D435i cameras, one facing the workspace and one mounted on the
wrist (\cref{fig:real-setup,tab:real-setup}). Every method outputs end-effector pose targets and a binary gripper command on a 10\,Hz grid, and all
methods reach the robot through the same interface. For Put Back Block and Swap Blocks, we recorded 80 standard demonstrations by teleoperation with Phantom Omni. For other tasks, we recorded 100 standard demonstrations in the same way. All of
them start from the same home pose of the arm, and none contains a task-progress shift.

\begin{table}[!b]
\centering\footnotesize
\definecolor{setgrp}{HTML}{ECECEC}
\renewcommand{\arraystretch}{1.35}\setlength{\aboverulesep}{0pt}\setlength{\belowrulesep}{0pt}
\setlength{\tabcolsep}{5pt}
\begin{tabular}{ll}
\toprule
\rowcolor{setgrp}\textbf{Setting} & \textbf{Value}\\
Robot & Franka Emika Panda, 7 joints\\
Gripper & Franka Hand, two fingers, opening 0 to 8\,cm\\
Third-view camera & Intel RealSense D435i on a fixed arch\\
Wrist camera & Intel RealSense D435i on the hand\\
Camera stream & color only, $640{\times}480$ at 30\,Hz\\
Control & ROS~2, 10\,Hz\\
Teleoperation & Phantom Omni haptic device, 10\,Hz\\
Images & $256{\times}256$, random $216{\times}216$ crops\\
Proprioception $q_t$ & position (3), rotation (6), gripper opening (1)\\
\bottomrule
\end{tabular}
\caption{\textbf{Real-robot setup} of all six tasks.}
\label{tab:real-setup}
\end{table}

\paragraph{Actions.}
An action $a_t$ of \methodname{} is the absolute end-effector target of the next step in the robot base frame: the position (3), the orientation
as the first two rows of its rotation matrix (6), and a gripper command in $[0,1]$, with 1 for open. In the demonstrations, $a_t$ is the measured
pose and gripper opening of step $t+1$, so on the robot the previous action $a_{t-1}$ in the token is the pose measured after the executed step.
The policy predicts 16 actions and executes the first 8, one every 0.1\,s, before it decides again. Each orientation is made orthonormal by
Gram--Schmidt and sent with the position as a Cartesian pose target. A gripper command below 0.5 closes the hand and one at or above 0.5 opens it,
with at most one gripper command per second.

\subsection{Task and Shifts}
\label{sec:supp_real_tasks}
\label{sec:supp_real_shifts}
\paragraph{Tasks.}
\Cref{tab:real-why-memory} gives, for each task, whether it needs memory and, if it does, what the memory must hold, why the scene does not show
it and the decision that depends on it.%
\begin{table*}[t]
\centering\footnotesize
\definecolor{whygrp}{HTML}{E6E6E6}\definecolor{whyrow}{HTML}{F6F6F4}\definecolor{whyname}{HTML}{5A5A5A}
\definecolor{tagHbg}{HTML}{DCE5F1}\definecolor{tagHfg}{HTML}{274468}\definecolor{tagAbg}{HTML}{F3E8CC}\definecolor{tagAfg}{HTML}{684C0F}\definecolor{tagTbg}{HTML}{EADFF1}\definecolor{tagTfg}{HTML}{4F366C}
\definecolor{tagVbg}{HTML}{E1EEE2}\definecolor{tagVfg}{HTML}{2F5A36}
\newcommand{\whytag}[3]{{\setlength{\fboxsep}{1.3pt}\colorbox{#1}{\rule[-0.45ex]{0pt}{1.85ex}\textcolor{#2}{\scriptsize\textsf{#3}}}}}
\newcommand{\whyhidden}{\whytag{tagHbg}{tagHfg}{hidden}}\newcommand{\whyalike}{\whytag{tagAbg}{tagAfg}{look-alike}}\newcommand{\whytrace}{\whytag{tagTbg}{tagTfg}{no trace}}
\newcommand{\whyview}{\whytag{tagVbg}{tagVfg}{in view}}
\renewcommand{\arraystretch}{1.6}\setlength{\aboverulesep}{0pt}\setlength{\belowrulesep}{0pt}
\setlength{\tabcolsep}{3.2pt}
\begin{tabular}{lclll}
\toprule
\textbf{Task} & \textbf{Memory} & \textbf{What the memory holds} & \textbf{Why the scene does not show it} & \textbf{What it decides}\\
\midrule
\textbf{Put Back Block} & yes & Its start mat; whether pressed & \whyalike\enspace the four mats\quad \whytrace\enspace the press & Mat to return to; whether to press\\
\rowcolor{whyrow}\textbf{Swap Blocks} & yes & Moves and presses done & \whytrace\enspace the presses\quad \whyalike\enspace done and new & Next move or press; when to stop\\
\textbf{Serve Bowls} & no & None, a multi-stage task & \whyview\enspace both bowls, all the time & --\\
\rowcolor{whyrow}\textbf{Clean Dirty Plate} & yes & The lime's plate; whether brushed & \whyalike\enspace the two plates\quad \whytrace\enspace the brushing & Plate to brush and rack\\
\textbf{Make Citrus Tea} & yes & The fruits in the pot & \whyhidden\enspace the pot under its lid & Buttons to press\\
\rowcolor{whyrow}\textbf{Swap Utensil} & yes & Utensil that was in the bowl & \whyalike\enspace the scene after the chicken step & Utensil for the bowl\\
\bottomrule
\end{tabular}
\caption[Why each real-robot task needs memory]{\textbf{Why each real-robot task needs memory.} For each task, whether it needs memory (Memory) and, if it does,
what the memory must hold in a trial without a shift, why the scene does not show it, and the decision that depends on it. The tag gives the reason:
\whyhidden{} out of the camera's view, \whyalike{} objects, places or scenes that look the same (on Swap Blocks a finished task and a new one), \whytrace{} an event that leaves the scene unchanged, and
\whyview{} the scene shows everything the task depends on. Serve Bowls is a multi-stage task, not a memory task.}
\label{tab:real-why-memory}
\end{table*}%

\paragraph{Put Back Block.}
A blue block stands on one of four outer mats arranged around a center mat, next to a button. The robot moves the block to the
center mat, presses the button, and puts the block back on the mat it came from. Once the block has left, the four outer mats look
alike, and a press leaves no visible trace, so only the memory holds where the block came from and whether the button has been
pressed. A trial succeeds if the button is pressed while the block is on the center mat and the block ends on its start mat.

\paragraph{Swap Blocks.}
A pink and a blue block stand on two neighboring mats of a row of three, next to a button, and the mat at one end is empty. The
robot moves the block on the middle mat onto the empty mat and presses the button, and then moves the other block onto the middle
mat and presses the button again. The presses leave no visible trace, and the final scene equals the other initial layout, with
the empty mat at the opposite end, so only the memory tells a finished task from a new one. A trial succeeds if both blocks have
moved one mat toward the mat that was empty and the button was pressed after each move.

\paragraph{Serve Bowls.}
A green and an orange bowl stand on two mats beside a place mat with a fork. The robot places the green bowl on the place mat on one
side of the fork and then the orange bowl on the other side. Serve Bowls is a multi-stage task, not a memory task. Both bowls stay in view
throughout, so the scene always shows how far the task has gone, and the task checks that a policy follows the progress that the scene shows. A trial succeeds if both bowls stand upright in
their places on the place mat.

\paragraph{Clean Dirty Plate.}
A lime lies on one of two identical plates on a place mat, next to a basket, a dish rack and a sponge in its holder. The robot puts
the lime into the basket, brushes the plate that held it with the sponge and returns the sponge to its holder, and then moves the
same plate to the rack. Once the lime is in the basket, the two plates look alike, and brushing leaves no visible trace, so only the
memory holds which plate to brush and rack and whether it has been brushed. A trial succeeds if the lime is in the basket, the
plate that held it has been brushed and stands in the rack, the sponge is back in its holder, and the other plate is still on the
place mat.

\paragraph{Make Citrus Tea.}
A lemon, a lime or both lie on the table next to a pot, its lid and two buttons, one marked with a lemon and one with a lime. The
robot puts each fruit into the pot, the lemon first when both are present, covers the pot with the lid, and presses the button of
every fruit in the pot. Once the lid is on, the contents of the pot cannot be seen, so only the memory holds which buttons to press.
A trial succeeds if every fruit is in the pot, the lid is on, and after the lid went on the robot pressed the button of each fruit
in the pot and no other button.

\paragraph{Swap Utensil.}
A white bowl in the middle of a place mat holds a fork or a knife, and the other utensil lies on its own side of the mat, the knife
on the left and the fork on the right. The robot takes the utensil out of the bowl and puts it on its own side, moves a toy chicken
from a metal bowl into the white bowl, and then puts the utensil that started on the mat into the white bowl. After the chicken step
both utensils lie on the mat, so only the memory tells which of them goes into the bowl. A trial succeeds if the utensil that started
in the bowl lies on its side of the mat and the chicken and the other utensil are in the white bowl.

\paragraph{Shifts.}
In every shifted trial a person makes one shift by hand while the robot keeps running (\cref{tab:supp_real_shifts}). In a backward
shift, the person waits until the robot has completed a step and returns the object of that step to where it was before the step.
In a forward shift, the person carries out a step before the robot has grasped its object and leaves the object where the step would
leave it. The person then withdraws, and since no signal marks the shift, the policies see it only in the camera images. After a
backward shift, a successful policy carries out the undone step again and then continues, and after a forward shift it goes on with
the next step without returning to the completed one. As in simulation, a success after a shift must not contain a ghost action,
which we read from the videos as the gripper closing at, or reaching down to, a place that a shifted object has left, such as the
empty start mat of the block after a forward shift on Put Back Block.

\begin{table*}[t]
\centering\footnotesize
\renewcommand{\arraystretch}{1.35}\setlength{\aboverulesep}{0pt}\setlength{\belowrulesep}{0pt}
\setlength{\tabcolsep}{6pt}
\begin{tabular}{l>{\columncolor{bwdbg}}l>{\columncolor{fwdbg}}l}
\toprule
\textbf{Task} & \bwdtag{}\enspace the person moves, after the robot's step & \fwdtag{}\enspace the person moves, before the robot's grasp\\
\midrule
\rowcolor{grpbg}\multicolumn{3}{l}{\textbf{One-step shifts}}\\
Put Back Block & the block back to its start mat, before the press & the block from its start mat to the center mat\\
Swap Blocks & the moved block back to the middle mat & the middle block onto the empty mat\\
Serve Bowls & the green bowl back to its start mat, before the orange bowl & the green bowl to its place on the place mat\\
 & & (a) the lime from its plate into the basket\\
\multirow{-2}{*}{Clean Dirty Plate} & \multirow{-2}{*}{the lime back on its plate, before the brushing} & (b) the sponge, brushing the plate, then back to its holder\\
Make Citrus Tea & a fruit from the pot back to its place, before the lid goes on & a fruit from the table into the pot\\
Swap Utensil & the utensil back into the white bowl & the utensil from the white bowl to its side of the mat\\
\midrule
\rowcolor{grpbg}\multicolumn{3}{l}{\textbf{Multi-step shifts}}\\
Clean Dirty Plate & the lime back on the plate after the brushing & the lime into the basket, and the plate brushed\\
Make Citrus Tea & both fruits from the pot back to their places & both fruits from the table into the pot\\
\bottomrule
\end{tabular}
\caption{\textbf{Task-progress shifts on the real robot}, what the person moves in each. A backward shift comes after the robot has completed the
step and undoes it, and a forward shift comes before the robot grasps the object of the step and completes it. Clean Dirty Plate has two one-step forward shifts, in which the person does the lime step (a) or the brushing (b).
The multi-step shifts undo or complete two consecutive steps at once. On Clean Dirty Plate, the lime put back on the brushed plate undoes both the lime step and the brushing,
and the lime put into the basket together with the brushing leaves only the plate to be racked.}
\label{tab:supp_real_shifts}
\end{table*}

On Make Citrus Tea, a fruit that the person puts into the pot still counts as being in the pot, so the robot must press its button
although it never held that fruit. On Put Back Block and Swap Utensil, and in the forward shift (a) of Clean Dirty Plate, the person moves the object before
the robot has touched it, so only the frames before the shift show where the block came from, which plate held the lime and which
utensil started in the bowl.

\begin{figure*}[p]
\centering
\includegraphics[width=\textwidth]{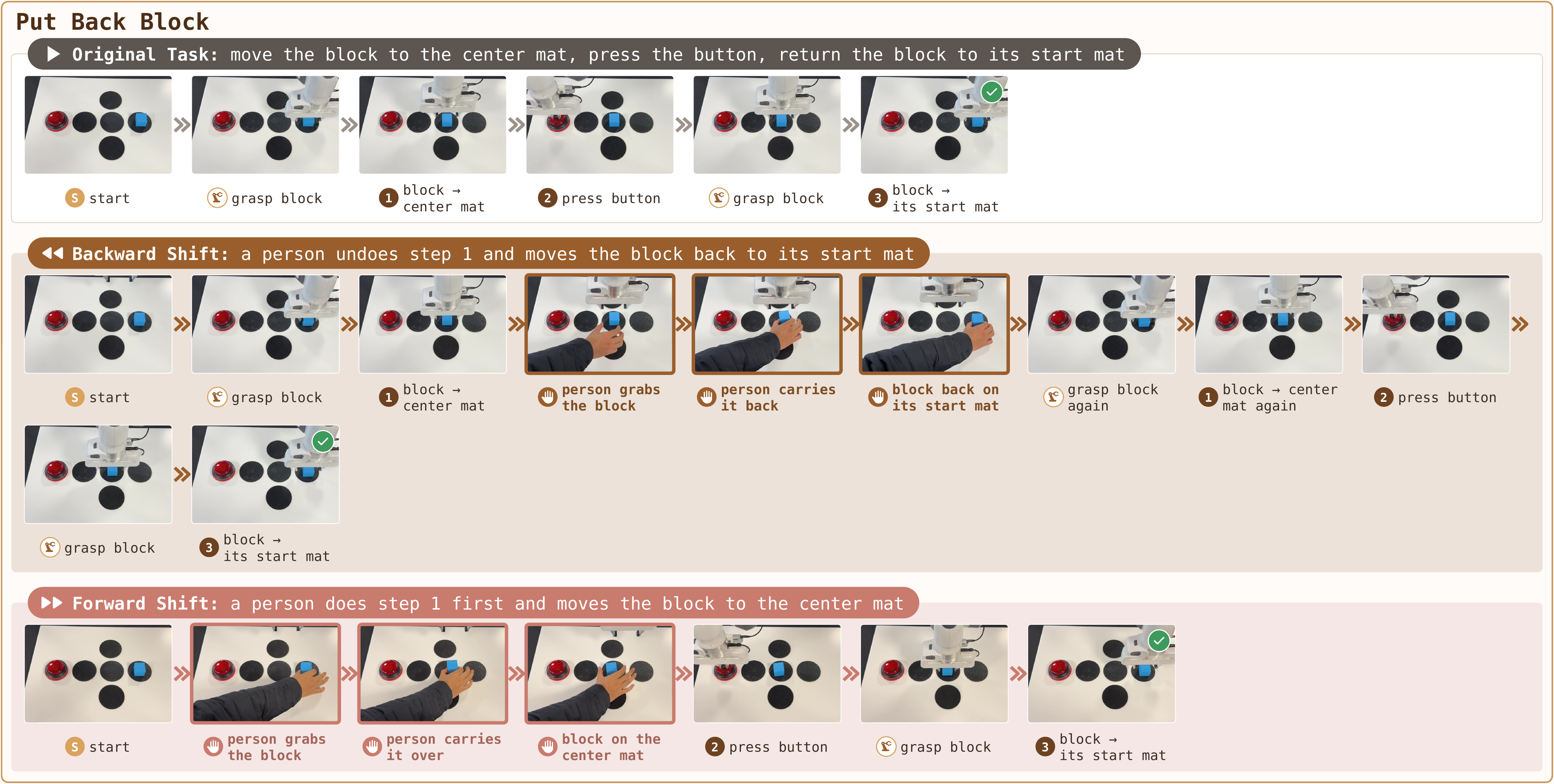}
\caption[Real robot: Put Back Block]{\textbf{Task-progress shifts on the real robot: Put Back Block.}\protect\shiftband{bwdband}{\protect\bwdtag: after the robot has moved the block to the center mat and before it presses the button, a person moves the block back to its start mat, and the robot should move it to the center mat again\protect\thn{}press the button\protect\thn{}return the block to its start mat.}\protect\shiftband{fwdband}{\protect\fwdtag: a person moves the block to the center mat before the robot grasps it, and the robot should leave it\protect\thn{}press the button\protect\thn{}return the block to the start mat that only the frames before the shift show.}The first row is a trial without a shift (gray). Each further row is a trial with one shift, backward (brown) or forward (rose). Every row is a trial of \methodname{} at its key frames, which are the start, every grasp, every place, brush or press that completes a step, and three frames of the person's change, when the hand takes the object, moves it and sets it down (four in a multi-step shift). Outlined frames show the person's hand. A number gives the step that a frame completes, S marks the start, an arm a grasp, a hand the person, and a check success.}
\label{fig:real-shift-rows}
\end{figure*}

\begin{figure*}[p]
\centering
\includegraphics[width=\textwidth]{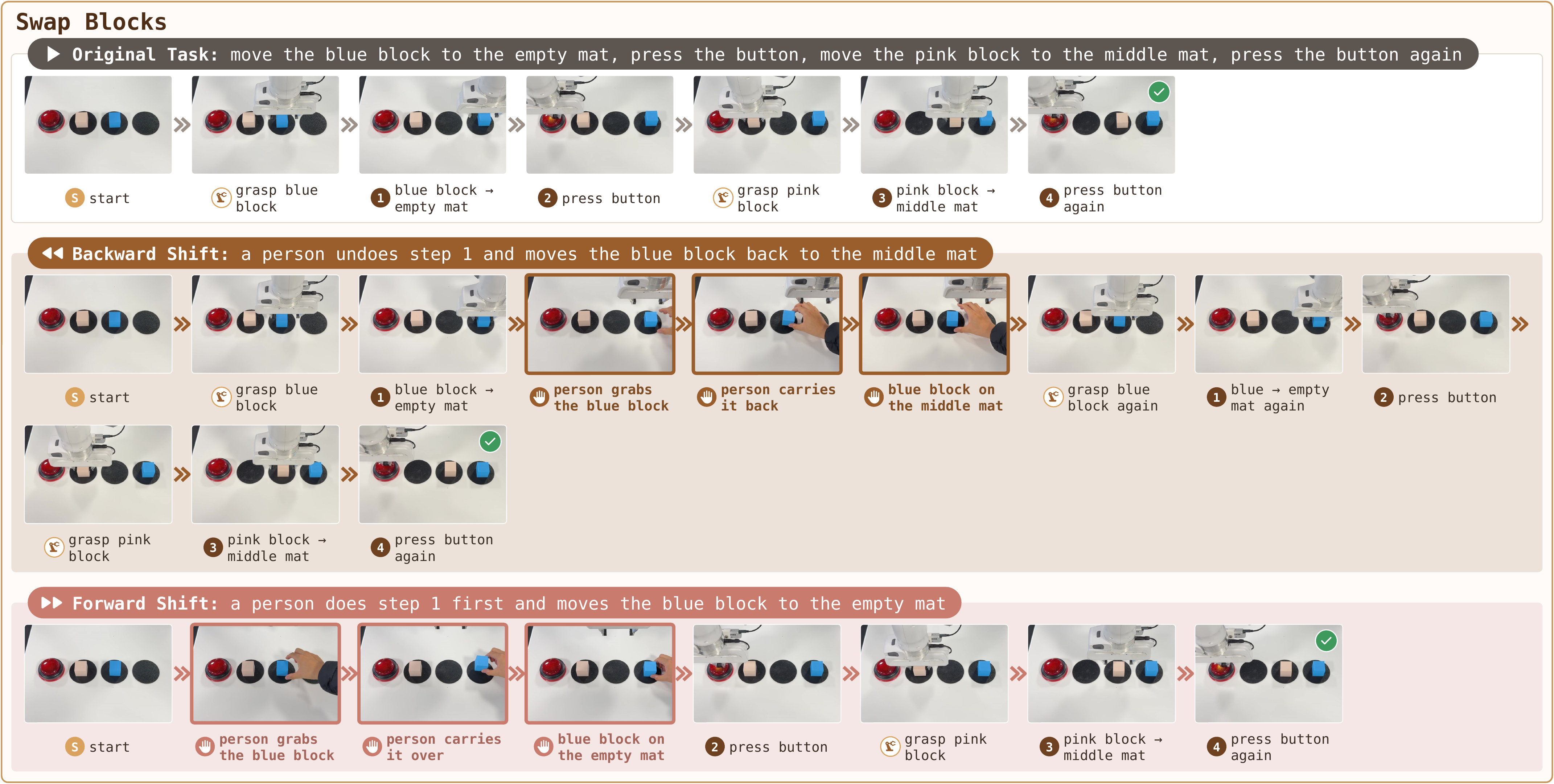}
\caption[Real robot: Swap Blocks]{\textbf{Task-progress shifts on the real robot: Swap Blocks.}\protect\shiftband{bwdband}{\protect\bwdtag: after the robot has moved the blue block onto the empty mat and before it presses the button, a person moves the blue block back to the middle mat, and the robot should move it onto the empty mat again\protect\thn{}press the button\protect\thn{}move the pink block to the middle mat\protect\thn{}press the button again.}\protect\shiftband{fwdband}{\protect\fwdtag: a person moves the blue block onto the empty mat before the robot grasps it, and the robot should leave it\protect\thn{}press the button\protect\thn{}move the pink block to the middle mat\protect\thn{}press the button again.}Read as \cref{fig:real-shift-rows}.}
\label{fig:real-shift-rows-2}
\end{figure*}

\begin{figure*}[p]
\centering
\includegraphics[width=\textwidth]{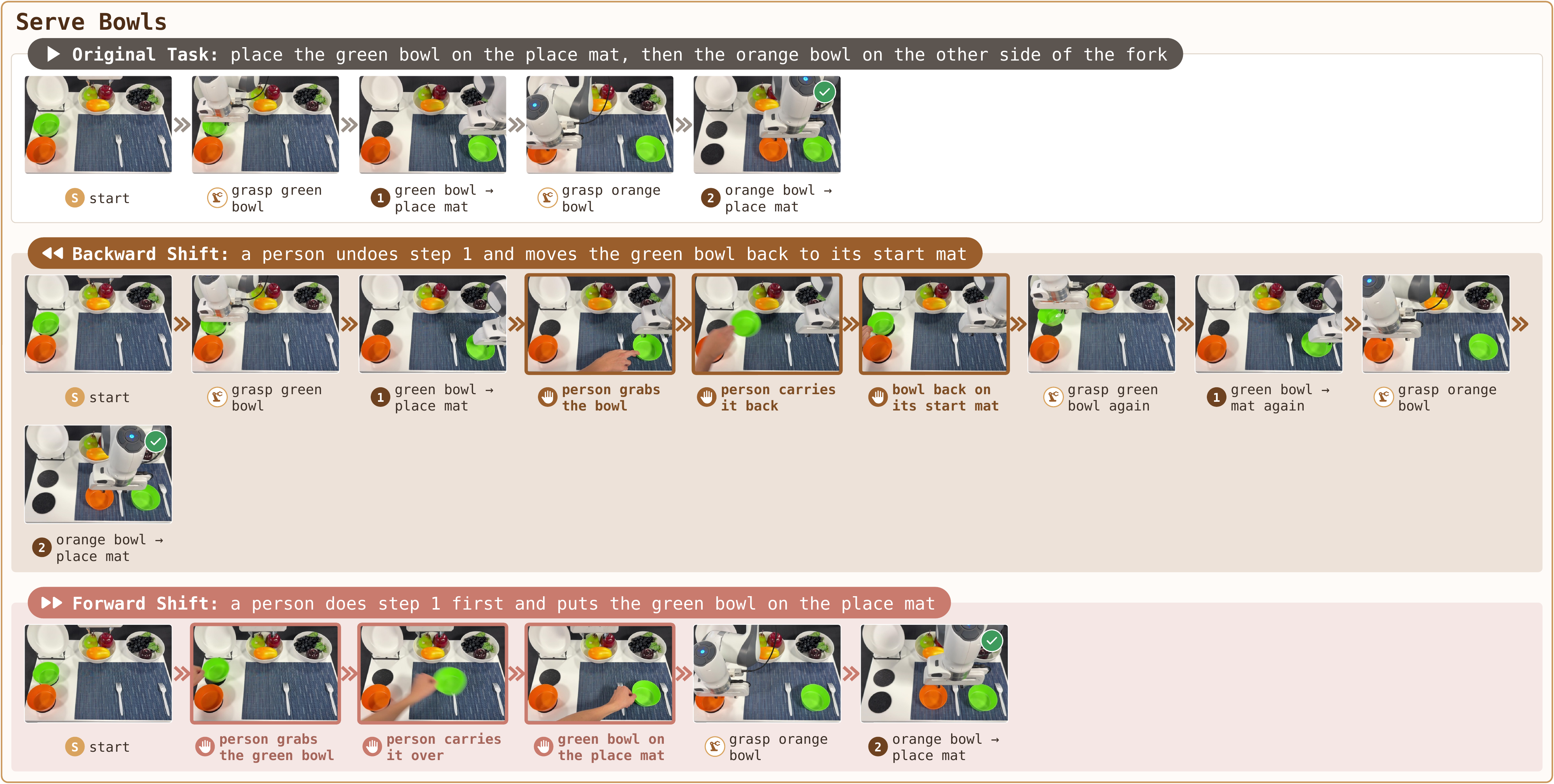}
\caption[Real robot: Serve Bowls]{\textbf{Task-progress shifts on the real robot: Serve Bowls.} Serve Bowls is a multi-stage task, not a memory task, as both bowls stay in view.\protect\shiftband{bwdband}{\protect\bwdtag: after the robot has placed the green bowl and before it takes the orange bowl, a person moves the green bowl back to its start mat, and the robot should place it again\protect\thn{}place the orange bowl.}\protect\shiftband{fwdband}{\protect\fwdtag: a person puts the green bowl on the place mat before the robot grasps it, and the robot should leave it\protect\thn{}place the orange bowl.}Read as \cref{fig:real-shift-rows}.}
\label{fig:real-shift-rows-3}
\end{figure*}

\begin{figure*}[p]
\centering
\includegraphics[width=\textwidth]{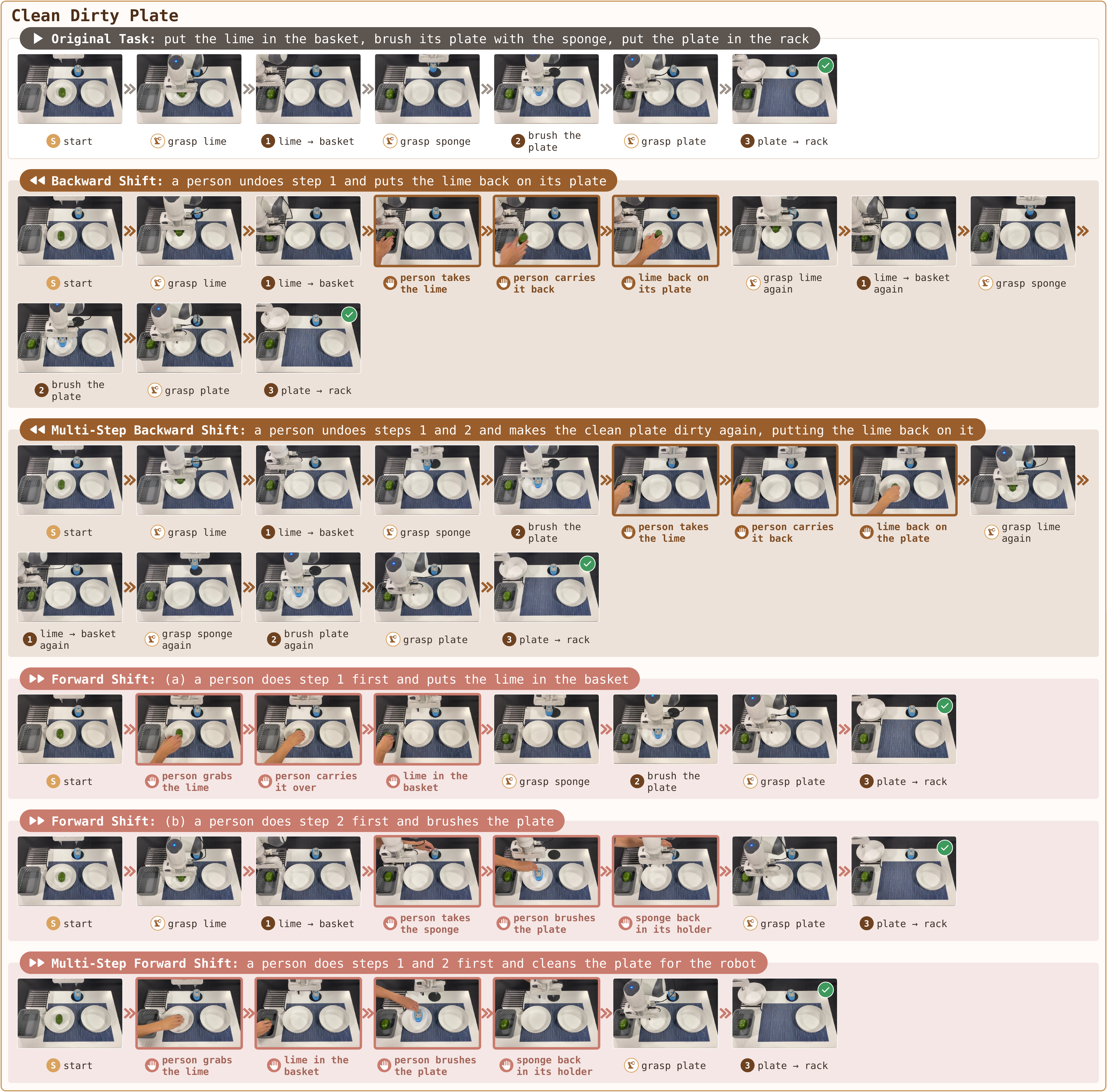}
\caption[Real robot: Clean Dirty Plate]{\textbf{Task-progress shifts on the real robot: Clean Dirty Plate.}\protect\shiftband{bwdband}{\protect\bwdtag: after the robot has put the lime in the basket and before the brushing, a person puts the lime back on its plate, and the robot should put it in the basket again\protect\thn{}brush the plate\protect\thn{}rack it. Multi-step: after the brushing, the person makes the clean plate dirty again by putting the lime back on it, which undoes the lime step and the brushing, and the robot should do both again\protect\thn{}rack the plate.}\protect\shiftband{fwdband}{\protect\fwdtag, two one-step shifts (\cref{tab:supp_real_shifts}): (a) a person puts the lime in the basket before the robot grasps it, and the robot should leave it\protect\thn{}brush the plate\protect\thn{}rack it. (b) After the lime step, the person brushes the plate and returns the sponge, and the robot should leave the sponge\protect\thn{}rack the plate. Multi-step: the person cleans the plate for the robot, putting the lime in the basket and brushing the plate, and the robot should only rack the plate.}Read as \cref{fig:real-shift-rows}.}
\label{fig:real-shift-rows-4}
\end{figure*}

\begin{figure*}[p]
\centering
\includegraphics[width=\textwidth]{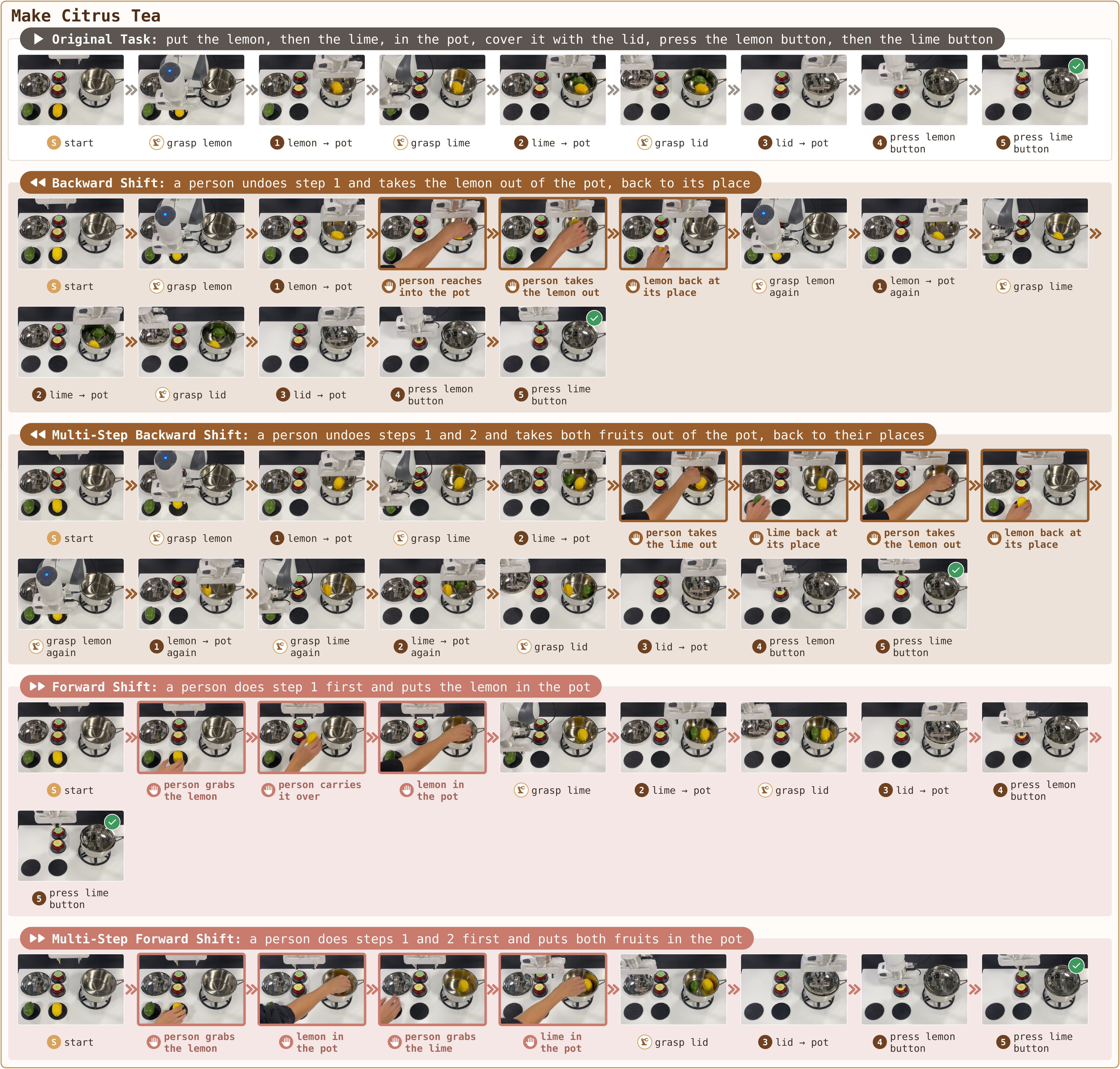}
\caption[Real robot: Make Citrus Tea]{\textbf{Task-progress shifts on the real robot: Make Citrus Tea.}\protect\shiftband{bwdband}{\protect\bwdtag: after the robot has put the lemon in the pot and before it takes the lime, a person takes the lemon out and sets it back at its place, and the robot should put it in again\protect\thn{}put in the lime\protect\thn{}cover the pot\protect\thn{}press the lemon button and the lime button. Multi-step: after both fruits are in the pot, the person sets both back at their places, and the robot should put both in again\protect\thn{}cover the pot\protect\thn{}press both buttons.}\protect\shiftband{fwdband}{\protect\fwdtag: a person puts the lemon in the pot before the robot grasps it, and the robot should leave it\protect\thn{}put in the lime\protect\thn{}cover the pot\protect\thn{}press both buttons, the lemon's too. Multi-step: the person puts both fruits in the pot, and the robot should cover the pot\protect\thn{}press both buttons.}Read as \cref{fig:real-shift-rows}.}
\label{fig:real-shift-rows-5}
\end{figure*}

\begin{figure*}[p]
\centering
\includegraphics[width=\textwidth]{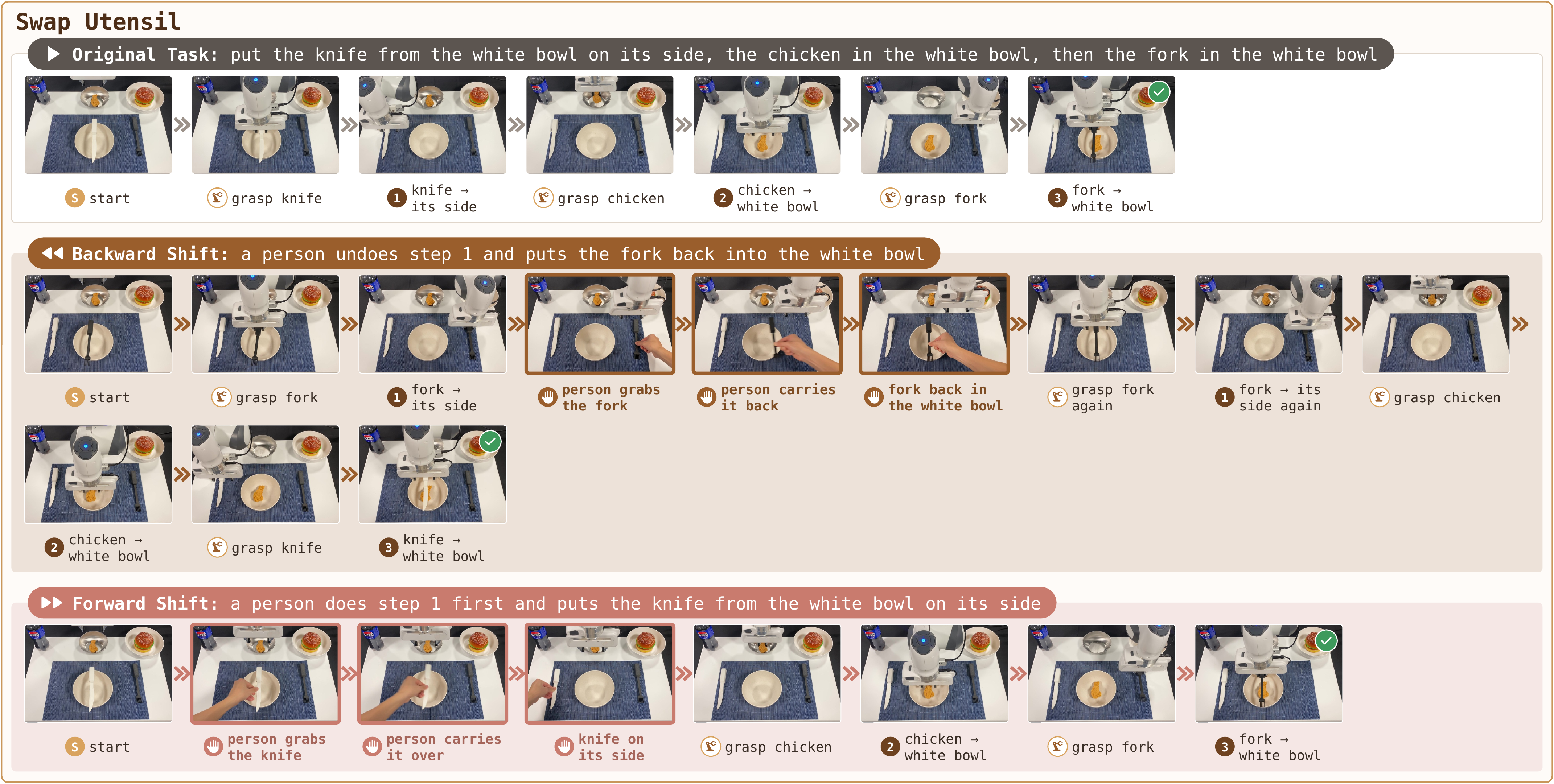}
\caption[Real robot: Swap Utensil]{\textbf{Task-progress shifts on the real robot: Swap Utensil.}\protect\shiftband{bwdband}{\protect\bwdtag, in a trial in which the fork starts in the white bowl: after the robot has put the fork on its side and before it takes the chicken, a person puts the fork back into the white bowl, and the robot should put it on its side again\protect\thn{}move the chicken into the white bowl\protect\thn{}put the knife into the white bowl.}\protect\shiftband{fwdband}{\protect\fwdtag: a person puts the knife from the white bowl on its side before the robot grasps it, and the robot should leave it\protect\thn{}move the chicken into the white bowl\protect\thn{}put the fork into the white bowl.}Read as \cref{fig:real-shift-rows}.}
\label{fig:real-shift-rows-6}
\end{figure*}

\Cref{fig:real-shift-rows,fig:real-shift-rows-2,fig:real-shift-rows-3,fig:real-shift-rows-4,fig:real-shift-rows-5,fig:real-shift-rows-6} show, for
every task, a trial of \methodname{} without a shift and one with each shift, at its key frames.

\clearpage
\makeatletter\global\@colroom\textheight\global\vsize\textheight\makeatother
\subsection{Failure mode analysis}
\label{sec:supp-real-failures}
\label{sec:supp_real_failures}
\begin{figure*}[tp]
\centering
\includegraphics[width=\textwidth]{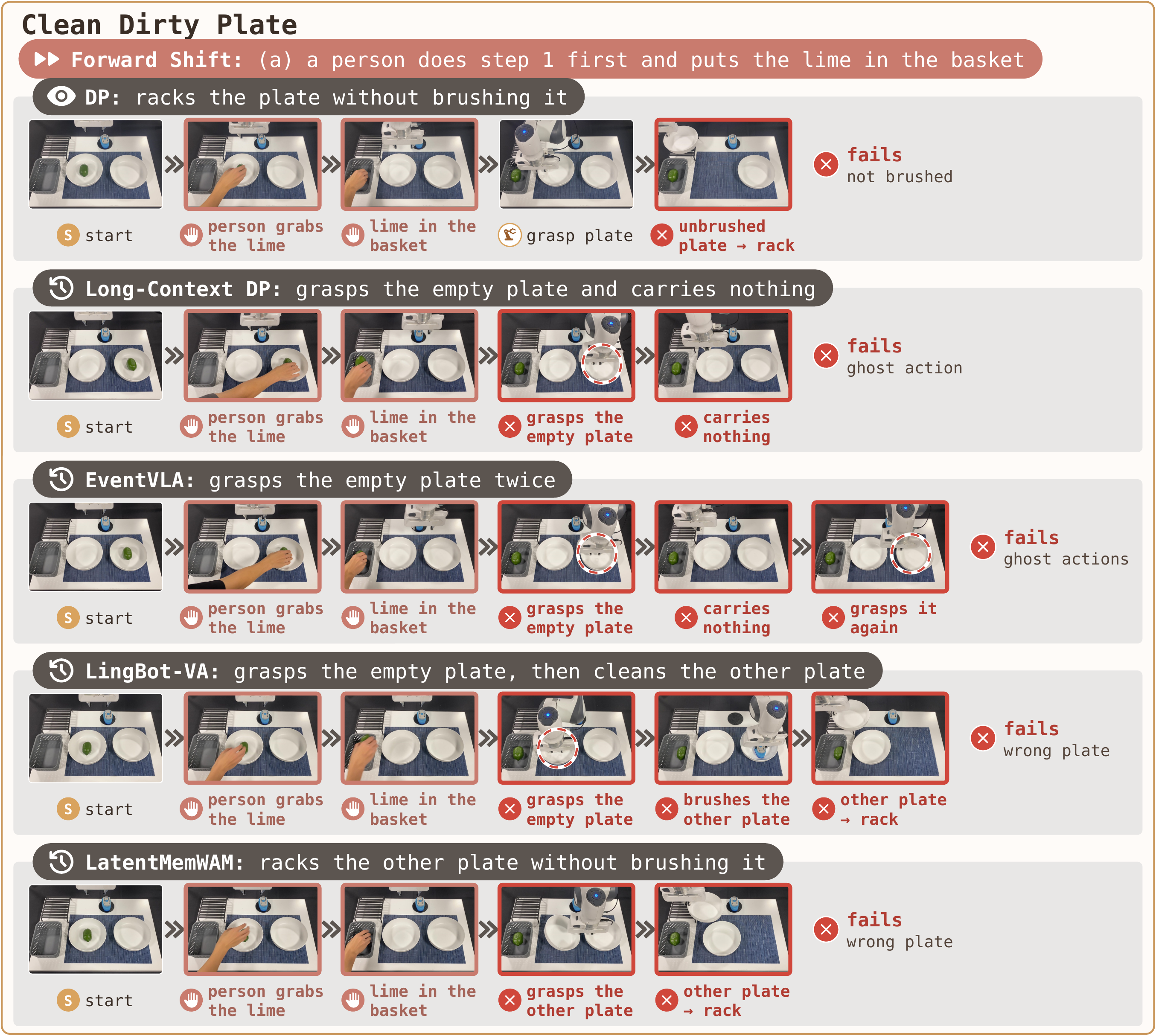}
\caption{\textbf{Baseline failures after a forward shift.} One trial of each baseline on Clean Dirty Plate, where the robot puts the lime in the basket
(step~1), brushes its plate with the sponge (step~2) and puts that plate in the rack (step~3). In the one-step forward shift (a), a person puts the lime
in the basket before the robot reaches it. Rose frames: the person shifts the task progress. Red frames: a wrong action, such as closing the gripper on
the empty plate where the lime was (dashed ring), racking a plate that was not brushed or cleaning the plate that never held the lime.}
\label{fig:real-failures}
\end{figure*}

How each method fails follows from how it keeps track of the task. DP~\cite{chi2025diffusion} keeps no memory and loses track wherever the scene
hides the progress. The baselines with memory advance it only with the robot's own actions, so after a backward shift they continue as if the undone
step were still done, and after a forward shift they redo the person's step, closing the gripper where the moved object was. \methodname{} instead
revises its memory to the step that the scene shows. For every trial we note the first failure that decided it and whether a success after a shift
contains a ghost action. As in simulation, many baseline successes after a forward shift do, so task success overstates how often baselines react to it.

\Cref{fig:real-failures} shows one failed trial of each baseline after the forward shift on Clean Dirty Plate in which the person puts the lime in the
basket first. Long-Context DP, EventVLA and LingBot-VA still go for the lime and close their gripper on the empty plate. DP cannot tell whether the
plate has been brushed, since brushing leaves no trace, and racks it unbrushed. LatentMemWAM and LingBot-VA rack the plate that never held the lime,
although their memory holds the frames that show which plate held it. In the same shift, \methodname{} fast-forwards its memory past the lime step and
brushes and racks the plate that held the lime (\cref{fig:real-shift-rows-4}).

\textbf{DP}~\cite{chi2025diffusion} completes only Serve Bowls, whose scene always shows the progress, and some trials of Swap Utensil without a
shift. On the other tasks it presses the button again and again on Put Back Block, moves a block back on Swap Blocks and brushes or racks the plate
that never held the lime on Clean Dirty Plate. It succeeds after a shift only on Swap Utensil, and on Serve Bowls it closes its gripper on the empty
start mat of the green bowl after a forward shift.

\textbf{Long-Context DP}~\cite{torne2025learning} conditions on a sparse history of past frames, so without a shift it completes memory tasks that DP
fails, such as Clean Dirty Plate, Swap Utensil and Put Back Block, but it never succeeds after a shift without a ghost action. After the forward shift
on Swap Utensil it closes its gripper in the empty white bowl, where DP goes straight on to the chicken, which suggests that its history holds it to
the step before the shift. Without a shift it fails most on Swap Blocks and Make Citrus Tea.

\textbf{EventVLA}~\cite{yang2026eventvla} completes the most unperturbed trials of the baselines other than LatentMemWAM but rarely reacts to a shift.
All its successes after forward shifts on Make Citrus Tea, Put Back Block and Clean Dirty Plate contain a ghost action, and it redoes an undone step
only on Make Citrus Tea. Without a shift it mostly misses grasps on Put Back Block and presses a wrong button on Make Citrus Tea.

\textbf{LingBot-VA}~\cite{li2026causal} holds the whole episode in its key-value cache. After a backward shift it presses the button with the block
back on its start mat or puts the lid on before the fruit is back in the pot, and it succeeds only on Make Citrus Tea. It never succeeds after a
forward shift without a ghost action. Without a shift it grasps the object of its first step again on Swap Blocks and Swap Utensil and presses wrong
or repeated buttons on Make Citrus Tea.

\textbf{LatentMemWAM} is the world action model of \methodname{} with the same weights, run without the memory localizer, so its memory only appends
what the robot observes. It is on par with \methodname{} without a shift but almost always fails after one, as the other memory baselines do or by
skipping ahead with the wrong object (\cref{fig:real-failures}), so the gains of \methodname{} come from revising the memory.

\textbf{\methodname{}} recovers from shifts far more often than any baseline. It rewinds its memory shortly after the person puts the object back and
redoes the step, and it fast-forwards its memory shortly after the person's move and goes on. Every baseline fails all multi-step shifts, while
\methodname{} succeeds after both backward and forward ones on Make Citrus Tea and even on Clean Dirty Plate, where the brushing leaves
no trace and is seen only while the person performs it.
\label{sec:supp_real_setup}


\end{document}